\documentclass[10pt,journal,compsoc]{IEEEtran}
\usepackage{hyperref}
\usepackage{url}
\usepackage{amsmath,amssymb,amsfonts}
\usepackage{amsthm}
\usepackage{algorithmic}
\usepackage{graphicx}
\usepackage{pifont}

\usepackage{subcaption}
\usepackage{thmtools, thm-restate}

\usepackage{booktabs}
\usepackage{bm}
\usepackage{capt-of}
\usepackage{siunitx}
\usepackage{xspace}
\usepackage{xcolor}
\usepackage{multirow}
\usepackage{wrapfig}
\usepackage{algorithm}
\usepackage{lmodern,babel,adjustbox,booktabs,multirow}
\usepackage{mathrsfs}

\newcounter{algsubstate}

\def\REVISION#1{\textcolor[RGB]{0,0,0}{#1}}

\newtheorem{theorem}{Theorem}[section]
\newtheorem{definition}[theorem]{Definition}
\newtheorem{corollary}[theorem]{Corollary}

\newtheorem{lemma}[theorem]{Lemma}

\newtheorem{assumption}[theorem]{Assumption}

\renewcommand{\eqref}[1]{\mbox{Eq.~(\ref{#1})}}

\makeatletter
\DeclareRobustCommand\onedot{\futurelet\@let@token\@onedot}
\def\@onedot{\ifx\@let@token.\else.\null\fi\xspace}

\def\ie{\emph{i.e}\onedot}

\def\etal{\emph{et al}\onedot}
\makeatother

\newcolumntype{L}[1]{>{\raggedright\arraybackslash}p{#1}}
\newcolumntype{C}[1]{>{\centering\arraybackslash}p{#1}}
\newcolumntype{R}[1]{>{\raggedleft\arraybackslash}p{#1}}

\ifCLASSOPTIONcompsoc
  \usepackage[nocompress]{cite}
\else
  \usepackage{cite}
\fi

\ifCLASSINFOpdf
\else
\fi
\begin{document}
        \title{On Demographic Group Fairness Guarantees in Deep Learning}

	\author{Yan~Luo\textsuperscript{*},~\IEEEmembership{Member,~IEEE,}
		Congcong Wen\textsuperscript{*},~\IEEEmembership{Member,~IEEE,}
		Min Shi,~\IEEEmembership{Member,~IEEE,}
        Hao Huang,
            Yi Fang\textsuperscript{$\dagger$},~\IEEEmembership{Member,~IEEE,}
		and~Mengyu Wang\textsuperscript{$\dagger$},~\IEEEmembership{Member,~IEEE}
		\IEEEcompsocitemizethanks
		{
                \IEEEcompsocthanksitem Yan Luo and Mengyu Wang are with the Harvard AI and Robotics Lab at Harvard University, Boston, MA, USA. E-Mails: \{yluo16, mengyu\_wang\}@meei.harvard.edu.
			\IEEEcompsocthanksitem Congcong Wen, Hao Huang, and Yi Fang are with Embodied AI and Robotics (AIR) Lab at New York University, New York, NY, USA. E-Mails: \{wencc, hh1845, yfang\}@nyu.edu
            \IEEEcompsocthanksitem Min Shi is with the School of Computing and Informatics, University of Louisiana at Lafayette, LA USA. E-Mail: min.shi@louisiana.edu.
                \IEEEcompsocthanksitem * Contributed equally as co-first authors.
                \IEEEcompsocthanksitem $\dagger$ Contributed equally as co-senior authors.
		}
		\thanks{Manuscript received December XX, 2024; revised July XX, 202X.}
	}

	\markboth{Journal of \LaTeX\ Class Files,~Vol.~14, No.~8, August~2024}%
	{Shell \MakeLowercase{\textit{et al.}}: Bare Demo of IEEEtran.cls for Computer Society Journals}

	\IEEEtitleabstractindextext{%
		\begin{abstract}
We present a comprehensive theoretical framework analyzing the relationship between data distributions and fairness guarantees in equitable deep learning. Our work establishes novel theoretical bounds that explicitly account for data distribution heterogeneity across demographic groups, while introducing a formal analysis framework that minimizes expected loss differences across these groups. We derive comprehensive theoretical bounds for fairness errors and convergence rates, and characterize how distributional differences between groups affect the fundamental trade-off between fairness and accuracy. Through extensive experiments on diverse datasets across various modalities (image, tabular data, and text), including FairVision (eye disease detection), CheXpert (pleural effusion detection), HAM10000 (skin lesion classification), FairFace (facial attribute recognition), ACS Income (income prediction), CivilComments-WILDS (toxic comment detection), we validate our theoretical findings and demonstrate that differences in feature distributions across demographic groups significantly impact model fairness, with performance disparities particularly pronounced in racial categories. The theoretical bounds we derive corroborate these empirical observations, providing insights into the fundamental limits of achieving fairness in deep learning models when faced with heterogeneous data distributions. This work advances our understanding of fairness in AI-based diagnosis systems and provides a theoretical foundation for developing more equitable algorithms. Motivated by these theoretical insights, particularly the link between feature distribution shifts and fairness gaps, we propose Fairness-Aware Regularization (FAR), a practical training objective that directly minimizes inter-group discrepancies in feature centroids and covariances to improve equitable performance. We validate the effectiveness of FAR across all datasets considered in this study, consistently observing improvements in overall AUC, ES-AUC, and subgroup performance. The code for analysis is publicly available via \url{https://github.com/Harvard-AI-and-Robotics-Lab/FairnessGuarantee}.
\end{abstract}
		
		}

	\maketitle

	\IEEEdisplaynontitleabstractindextext

	%
	\IEEEpeerreviewmaketitle

	
    \section{Introduction}

Fairness in machine learning has become an increasingly important concern, especially in high-stakes applications such as healthcare, where biased predictions can have severe consequences \cite{rajkomar2018ensuring}. Equitable deep learning aims to ensure that predictive models perform equally well across various demographic groups, regardless of factors such as gender, race, and ethnicity \cite{chen2018my}. However, achieving fairness guarantees in deep learning models is a challenging task, as it requires careful consideration of the variations in data distribution and the positive sample rate across different groups \cite{oakden2020hidden}.


Recent work\REVISION{~\cite{zafar2017fairness2,wan2023processing,han2023ffb,li2022fairee}} has focused on developing fairness-aware learning algorithms and analyzing their theoretical properties. For example, Zietlow \etal \cite{zietlow2022leveling} formulate the fairness problem as minimizing the absolute difference in the expected per-group error between two groups. Also, the fairness problem in predictive ability has been formalized as minimizing the difference in the expected loss across all demographic groups \cite{dwork2012fairness}. This definition provides a mathematical framework for studying fairness in machine learning models and has been used to derive various fairness error bounds and convergence guarantees \cite{agarwal2018reductions,zafar2017fairness}. However, while existing strategies have improved model fairness to some extent, there has been limited theoretical analysis of the relationship between data distributions and fairness guarantees \cite{chouldechova2017fair}. The implications of biases are particularly concerning in high-stakes applications, where unfair assessments can perpetuate systemic inequalities, affect individual opportunities, and erode public trust in automated decision systems \cite{corbett2017algorithmic}.

In addition, previous research has investigated the relationship between data distribution, class prevalence, and fairness in machine learning \cite{hajian2015discrimination, chouldechova2017fair}. For example, Hajian and Domingo-Ferrer examine how class imbalance affects the fairness of classification models \cite{hajian2015discrimination}, revealing that performance can vary significantly across different groups when class distribution is uneven. Similarly, Chouldechova explores the impact of data distribution on the fairness of risk assessment models in the criminal justice system \cite{chouldechova2017fair}, demonstrating that the choice of training data distribution can substantially influence the fairness properties of the resulting model.
However, these studies primarily focus on specific applications without providing a comprehensive theoretical framework for understanding how differences in class prevalence and feature distributions across demographic groups affect fairness guarantees in general machine learning contexts.

This work establishes a comprehensive theoretical framework linking data distribution heterogeneity across demographic groups to fairness guarantees in deep learning. We derive novel bounds showing that the fairness error—defined as the maximum difference in expected loss across groups—decomposes into irreducible, statistical, and optimization components (Theorem~\ref{thm:fair_errorbound}), and converges at a rate of $O(1/\sqrt{m})$ with sample size $m$ (Theorem~\ref{thm:convergence}). Crucially, we prove that a group’s excess risk is fundamentally constrained by its feature distribution shift relative to others: specifically, disparities in group-wise means and covariances directly upper-bound performance gaps (Theorems~\ref{thm:group_risk}, \ref{thm:accuracy}; Corollary~\ref{thm:correlation_lossbound_featdist}). This formalizes the trade-off between overall accuracy and subgroup equity (Corollary~\ref{thm:fairness_accuracy}) and reveals that fairness is limited not just by algorithmic design, but by inherent statistical differences in group data—particularly pronounced in attributes like race. These insights motivate a practical fairness-aware regularization that minimizes inter-group feature discrepancies, thereby tightening the theoretical bounds and improving empirical equity.

The main contributions of this work are as follows:
\begin{itemize}
    \item We formalize fairness as minimizing expected loss discrepancies across demographic groups (e.g., race) and, under assumptions on data distributions and loss functions, derive theoretical guarantees, including fairness error bounds, generalization and group-specific risk bounds, convergence rates, and algorithmic complexity, that elucidate the key drivers of model fairness and inform the design of more robust, equitable algorithms.
    \item Motivated by the theoretical insights, we propose Fairness-Aware Regularization (FAR), a practical objective that directly minimizes inter-group discrepancies in feature centroids and covariances, and can be integrated into diverse deep learning models.
    \item We validate our theoretical findings with extensive empirical evaluations on six diverse datasets across various modalities (image, tabular data, and text), \ie, \textit{FairVision}~\cite{luo2023fairvision}, \textit{CheXpert}~\cite{irvin2019chexpert}, \textit{HAM10000}~\cite{tschandl2018ham10000}, \textit{FairFace}~\cite{maze2018iarpa}, \textit{ACS Income}~\cite{ding2021retiring}, and \textit{CivilComments-WILDS}~\cite{koh2021wilds}, comprising over 2.5M samples in total. These experiments consistently demonstrate that larger distributional shifts correspond to higher fairness bounds and lower predictive performance, and that our proposed FAR method improves overall AUC, ES-AUC, and subgroup-level equity.
\end{itemize}

The rest of this paper is organized as follows. In Section~\ref{sec:related}, we discuss related work on fairness learning and datasets. Section~\ref{sec:main} presents our main theoretical results on the impact of data distribution on fairness guarantees. Section~\ref{sec:fairness_reg} introduces our proposed Fairness-Aware Regularization (FAR), derived from the theoretical insights. In Section~\ref{sec:expr}, we discuss the empirical validation of the theoretical findings across all datasets, including FairVision, CheXpert, HAM, FairFace, ACS Income, and CivilComments-WILDS. Section~\ref{sec:concl} concludes the paper and outlines directions for future research. Table \ref{tab:notation} includes the key symbols and terms used throughout the theoretical sections.

\begin{table}[t]
\centering
\caption{List of Key Notations and Symbols}
\label{tab:notation}
\vspace{-2ex}
\begin{adjustbox}{width=0.45\textwidth, center}
\begin{tabular}{ll}
\hline
\textbf{Symbol} & \textbf{Description} \\
\hline
$x$ & Input data (e.g., an image). \\
$y$ & True label corresponding to input $x$. \\
$\hat{y}$ & Predicted label, $\hat{y} = f(x)$. \\
$a$ & Demographic attribute (e.g., race, gender). \\
$a_i, a_j$ & Specific values of demographic attribute $a$ (e.g., Asian). \\
$k$ & Total number of demographic groups. \\
$f(\cdot)$ & Model function mapping input $x$ to prediction $\hat{y}$. \\
$\ell(\hat{y}, y)$ & Loss function measuring prediction error. \\
$D_{a_i}$ & Data distribution for demographic group $a_i$. \\
$\mathbb{E}_{(x,y) \sim D_{a_i}}[\cdot]$ & Expectation operator over distribution $D_{a_i}$. \\
$\Delta(f)$ & Fairness error: $\max_{i,j} \left| \mathbb{E}_{D_{a_i}}[\ell(f(x), y)] - \mathbb{E}_{D_{a_j}}[\ell(f(x), y)] \right|$. \\
$f^*$ & Optimal function minimizing fairness error $\Delta(f)$. \\
$\hat{f}$ & Estimated/learned function (e.g., empirical risk minimizer). \\
$\epsilon$ & Optimality gap for an $\epsilon$-optimal solution. \\
$M$ & Upper bound on the loss function: $0 \leq \ell(\hat{y}, y) \leq M$. \\
$L$ & Lipschitz constant of the loss function $\ell$. \\
$\pi_i$ & Proportion of samples from group $a_i$ in the dataset ($\sum_{i=1}^k \pi_i = 1$). \\
$n$ & Total sample size. \\
$n_i$ & Sample size for group $a_i$, $n_i = n \pi_i$. \\
$d_{\text{VC}}(F)$ & Vapnik-Chervonenkis (VC) dimension of function class $F$. \\
$R_i(f)$ & Expected risk for group $a_i$: $\mathbb{E}_{(x,y) \sim D_{a_i}}[\ell(f(x), y)]$. \\
$R_{\text{emp},i}(f)$ & Empirical risk for group $a_i$. \\
$R(f)$ & Overall fairness risk, often synonymous with $\Delta(f)$. \\
$R_{\text{emp}}(f)$ & Empirical fairness risk. \\
$\mu_i, \Sigma_i$ & Mean and covariance matrix of the data distribution $D_{a_i}$. \\
$\|\cdot\|_2$ & Euclidean norm. \\
$\|\cdot\|_F$ & Frobenius norm. \\
$W_1(\cdot, \cdot)$ & 1-Wasserstein distance between two distributions. \\
$z = f(x)$ & Learned feature representation of input $x$. \\
$\bar{z}_i$ & Centroid (mean) of features for group $a_i$: $\mathbb{E}_{(x,y) \sim D_{a_i}}[f(x)]$. \\
$\Sigma_{z_i}$ & Covariance matrix of features for group $a_i$. \\
$R_{\text{fair}}(f)$ & Proposed fairness regularization term. \\
$\lambda$ & Hyperparameter controlling fairness-accuracy trade-off. \\
\hline
\end{tabular}
\end{adjustbox}
\end{table}

    \section{Related Work}
\label{sec:related}
\subsection{Fairness in Learning}
In recent years, machine learning has achieved exceptional performance across various fields, yet it often produces biased predictions against different demographic groups. To address this issue, fairness learning~\cite{li2025fairfedmed,luo2025fairdiffusion} has been proposed to eliminate or reduce discrimination and bias against certain protected groups, ensuring fair treatment across different groups\REVISION{~\cite{wan2023processing,han2023ffb,li2022fairee}}. Existing fairness learning approaches can generally be categorized into three strategies: \textit{pre-processing}, which balances the dataset through sampling~\cite{wang2020towards,kamiran2012data} or generative models~\cite{ngxande2020bias}; \textit{in-processing}, which incorporates constraints into the loss function~\cite{xu2020investigating}; and \textit{post-processing}, which transforms the model output to ensure fairness~\cite{hardt2016equality,pleiss2017fairness}.

However, while these strategies have improved model fairness to some extent, there has been limited theoretical analysis of fairness learning. ~\cite{corbett2017algorithmic} demonstrates that the trade-off between improving public safety and satisfying current notions of algorithmic fairness can be achieved through theoretical proof and empirical analysis on Broward County, Florida's crime data. Additionally, ~\cite{lipton2018does} shows that increasing treatment disparity can enhance impact parity through theoretical analysis on simulated and real-world educational data. Recently, ~\cite{jang2024achieving} proposes a novel fair representation learning method, which achieves non-separability in latent distribution w.r.t. sensitive features by regularizing data distribution and increases separability w.r.t. the target label by maximizing the marginal distance of decision boundaries among different classes.
In contrast to the prior works, this work  focuses on the implications of the theoretical findings on fairness error bounds, algorithmic complexity, generalization bounds, convergence rates, and group-specific risk bounds.

\subsection{Theoretical and Algorithmic Foundations of Fairness}
\label{sec:theoretical_foundations}
It is generally impossible to satisfy common fairness criteria, such as calibration, balance for the positive class, and balance for the negative class, all at once unless the base rates (i.e., the prevalence of the positive outcome) are identical across demographic groups \cite{kleinberg2016inherent,chouldechova2017fair}.
Complementing this, \cite{dwork2012fairness} proposes to advocate for similarity-based fairness grounded in a task-specific metric, while \cite{hardt2016equality} introduces equality of opportunity, which requires equal true positive rates across protected groups. 
\cite{pleiss2017fairness} further explores the calibration-fairness trade-off, showing that enforcing certain fairness constraints can degrade calibration and thereby challenge the reliability of risk scores.
These foundational results highlight that fairness interventions must be carefully tailored to domain-specific values and operational goals, as no single definition universally resolves ethical tensions.
\\
Building on these insights, researchers have developed algorithmic frameworks to operationalize fairness constraints\REVISION{~\cite{zafar2017fairness2}} and navigate their inherent trade-offs. \cite{zafar2017fairness} and \cite{agarwal2018reductions} propose optimization-based and reduction-based methods, respectively, to enforce group fairness constraints during model training, enabling flexible trade-offs between accuracy and fairness. \cite{kamishima2012fairness} introduces prejudice remover regularizers, while \cite{feldman2015certifying} develops pre-processing techniques to remove disparate impact by transforming features. Causal reasoning has also emerged as a powerful lens: \cite{kusner2017counterfactual} and \cite{kilbertus2017avoiding} formalize counterfactual fairness, requiring predictions to remain unchanged under interventions on sensitive attributes, and \cite{chiappa2019path} extends this to path-specific effects, allowing nuanced modeling of direct and indirect discrimination. \cite{woodworth2017learning} analyzes the statistical limits of non-discriminatory learning, revealing fundamental sample complexity barriers.
More recently, \cite{corbett2023measure} and \cite{mitchell2021algorithmic} have urged caution in operationalizing fairness metrics, emphasizing that algorithmic choices embed normative assumptions that must be explicitly justified.
In this work, we advance this algorithmic landscape by deriving closed-form theoretical bounds that explicitly link group-specific performance gaps to measurable distributional shifts, specifically the Euclidean distance between group feature centroids and the Frobenius norm of covariance differences. This allows us to propose a novel fairness-aware regularization term that directly minimizes these distributional discrepancies in feature space, thereby tightening the theoretical upper bound on group-wise loss disparity.

    \section{Main Results}
\label{sec:main}
In this section, we first formalize fairness as minimizing the maximum loss difference across demographic groups (Definition \ref{eqn:def_fairness}), then progressively establish how sample size, model complexity, and optimization error affect fairness guarantees (Theorems \ref{thm:fair_errorbound}, \ref{thm:complexity}, \ref{thm:generalization}, \ref{thm:convergence}). We then show that, under distributional assumptions, the risk for any group is fundamentally bounded by its statistical distance in mean and covariance from other groups (Theorem \ref{thm:group_risk}). Finally, Theorem \ref{thm:accuracy} delivers our core insight: a group's expected loss is upper-bounded by the population loss plus a term proportional to its feature distribution shift, directly linking measurable data disparities to model unfairness.


A central question in equitable deep learning is how to formalize the notion of fairness in the context of predictive models.
Building upon the insights and the fairness problem formulation introduced by \cite{zietlow2022leveling}, we extend the formulation to multiple groups and introduce the fairness problem in predictive ability as minimizing the difference in the expected loss across all demographic groups.
\begin{definition}[Fairness Problem]\label{eqn:def_fairness}
Given an image $x$, its corresponding label $y$, a demographic attribute $a \in \{a_1, a_2, \ldots, a_k\}$ (e.g., race attributes such as Asian, Black, and White), a function $f$ mapping $x$ to predicted labels $\hat{y}$, and a loss function $\ell$, the fairness problem in predictive ability is defined as minimizing the difference in the expected loss across all demographic groups:

\begin{align}
\min_{f(\cdot)} \max_{i, j} \left| \mathbb{E}_{(x, y) \sim \mathcal{D}_{a_i}}[\ell(f(x), y)] - \mathbb{E}_{(x, y) \sim \mathcal{D}_{a_j}}[\ell(f(x), y)] \right|
\label{eqn:def}
\end{align}
where $\mathcal{D}_{a_i}$ represents the data distribution for demographic group $a_i$, and $\mathbb{E}_{(x, y) \sim \mathcal{D}_{a_i}}[\cdot]$ denotes the expectation over the data distribution for group $a_i$.

\end{definition}

In real-world applications, we have the optimization objective $\min_{f \in \mathcal{F}}$ $\{L(f(x), y) + \lambda \cdot \max_{i,j} \left|\mathbb{E}_{(x,y)\sim D_{a_i}}[\ell(f(x), y)] - \mathbb{E}_{(x,y)\sim D_{a_j}}[\ell(f(x), y)]\right| \}$, where $L(f(x), y)$ is a task-specific loss. To clearly understand the fairness problem, we focus on the fairness term (\ref{eqn:def}) in this work.

Definition \ref{eqn:def_fairness} formalizes the fairness problem in predictive ability for a machine learning model. The chain of logic is as follows:
Firstly, we have an image $x$, its corresponding true label $y$, and a demographic attribute $a$ which can take on values from $\{a_1, a_2, \ldots, a_k\}$. For example, the demographic attribute could be race, with categories like Asian, Black, and White.
Then, we have a function $f$ that maps the image $x$ to a predicted label $\hat{y}$, and a loss function $\ell$ that measures the difference between the predicted label and the true label.
Next, the goal is to minimize the maximum absolute difference in the expected loss across all demographic groups. In other words, we want the model to perform equally well (in terms of the loss function) for all demographic groups.
The expectation $\mathbb{E}_{(x, y) \sim \mathcal{D}_{a_i}}[\ell(f(x), y)]$ represents the average loss for demographic group $a_i$, where the data (image-label pairs) are sampled from the distribution $\mathcal{D}_{a_i}$ specific to that group.
Finally, by minimizing the maximum absolute difference in the expected loss across all pairs of demographic groups, we ensure that the model's predictive ability is as similar as possible across the groups. This helps mitigate potential biases or unfairness in the model's performance.

\begin{theorem}[Exact Connection with Conventional Fairness Metrics]
\label{thm:exact_connection}
Let $f$ be a classifier, $\ell$ be a loss function, and $D_{a_i}$ be the data distribution for demographic group $a_i$. Let $p_i$ be the proportion of positive samples in group $a_i$. Then, for any pair of groups $a_i$ and $a_j$, the difference in expected loss is exactly given by:
\begin{align}
\begin{split}
&\left| \mathbb{E}_{(x,y)\sim D_{a_i}}[\ell(f(x), y)] - \mathbb{E}_{(x,y)\sim D_{a_j}}[\ell(f(x), y)] \right| \\
= & | p_i \cdot L_{+,a_i}(f) - p_j \cdot L_{+,a_j}(f) \\
&+ (1-p_i) \cdot L_{-,a_i}(f) - (1-p_j) \cdot L_{-,a_j}(f) | \\
\end{split}
\end{align}
where $L_{+,a_k}(f)= \mathbb{E}[\ell(f(x), y)|y= 1, s= a_k]$ is the risk of the positive labeled samples in group $a_k$, and $L_{-,a_k}(f)= \mathbb{E}[\ell(f(x), y)|y= 0, s= a_k]$ is the risk of the negative labeled samples in group $a_k$.
\end{theorem}
\begin{proof}
We begin by decomposing the expected loss for each group into contributions from positive and negative samples:
\[
\mathbb{E}_{(x,y)\sim D_{a_k}}[\ell(f(x), y)] = p_k \cdot L_{+,a_k}(f) + (1 - p_k) \cdot L_{-,a_k}(f)
\]
for $k \in \{i, j\}$.
Now, consider the absolute difference in expected losses between groups $a_i$ and $a_j$:
\[
\left| \mathbb{E}_{(x,y)\sim D_{a_i}}[\ell(f(x), y)] - \mathbb{E}_{(x,y)\sim D_{a_j}}[\ell(f(x), y)] \right|
\]
Substituting the decomposed expressions:
\begin{align*}
    \begin{split}
        = & | \left( p_i \cdot L_{+,a_i}(f) + (1 - p_i) \cdot L_{-,a_i}(f) \right) - ( p_j \cdot L_{+,a_j}(f) \\
        &+ (1 - p_j) \cdot L_{-,a_j}(f) ) |
    \end{split}
\end{align*}
Rearranging terms:
\[
= | p_i \cdot L_{+,a_i}(f) - p_j \cdot L_{+,a_j}(f) + (1 - p_i) \cdot L_{-,a_i}(f) - (1 - p_j) \cdot L_{-,a_j}(f) |
\]
This is the desired exact expression, which establishes a direct and precise relationship between the fairness metric (difference in expected loss) and the conventional group-specific performance metrics ($L_{+}, L_{-}$) and class distributions ($p_i, p_j$).
\end{proof}


\noindent\textbf{Remark}. Theorem \ref{thm:exact_connection} provides an exact decomposition of the fairness metric (the difference in expected loss between groups) into a sum of terms involving group-specific risks ($L_{+,a_k}$, $L_{-,a_k}$) and class proportions ($p_k$). This precise relationship is fundamental because it reveals that fairness disparities are not solely determined by model performance but arise from the interplay of three key factors: (1) differences in positive-sample risk ($L_{+,a_i} - L_{+,a_j}$), (2) differences in negative-sample risk ($L_{-,a_i} - L_{-,a_j}$), and (3) differences in the class distribution ($p_i - p_j$). By stating an equality rather than an inequality, this result avoids the ambiguity associated with loose bounds and arbitrary constants (e.g., $M$), making it a more relevant and precise foundation for analyzing fairness.

\REVISION{Crucially, this decomposition allows Definition 3.1 to be interpreted through the lens of standard decision-based metrics. If $l$ is the 0-1 loss, $\mathbb{E}[l]$ represents the error rate, making Definition 3.1 equivalent to minimizing the Accuracy Gap. Furthermore, since $L_{+, a_k}$ and $L_{-, a_k}$ correspond to the False Negative Rate ($1-\text{TPR}$) and False Positive Rate ($\text{FPR}$) respectively, Equation (2) demonstrates that minimizing the expected loss difference implicitly bounds the disparities in TPR and FPR. Consequently, our loss-based framework serves as a differentiable surrogate for achieving Equalized Odds and Accuracy Parity in practical deployments.}

\begin{assumption}[Bounded Loss and Group Proportions]\label{assm:prevalence}
Let $\pi_i$ be the proportion of samples for demographic group $a_i$, such that in a dataset of size $n$, the number of samples for group $a_i$ is $n_i = n\pi_i$ and $\sum_{i=1}^{k} \pi_i = 1$. Assume that the loss function $l$ is bounded, i.e., $0 \le l(\hat{y}, y) \le M$ for some constant $M > 0$.
\end{assumption}
\begin{theorem}[Fairness Error Bound]\label{thm:fair_errorbound}
Under Assumption \ref{assm:prevalence}, given the fairness problem as defined in Definition \ref{eqn:def_fairness}, let $f^*$ be the optimal function that minimizes the maximum absolute difference in the expected loss across all demographic groups, \ie, $
f^* = \arg\min_{f \in \mathcal{F}} \Delta(f), \quad \text{where} \quad \Delta(f) = \max_{i,j} \left| \mathbb{E}_{D_{a_i}}[\ell(f(x), y)] - \mathbb{E}_{D_{a_j}}[\ell(f(x), y)] \right|$. Let $\hat{f}$ be an estimate of $f^*$ based on a finite sample of size $n$, and assume that $\|\hat{f}-f^*\|_{\infty}\le\epsilon$ for some $\epsilon>0$. Then, with probability at least $1-\delta$, the following inequality holds:
\begin{align*}
    \max_{i,j}|\mathbb{E}_{(x,y)\sim \mathcal{D}_{a_i}}[l(\hat{f}(x),y)] - \mathbb{E}_{(x,y)\sim \mathcal{D}_{a_j}}[l(\hat{f}(x),y)]| \le \\
    \max_{i,j}|\mathbb{E}_{(x,y)\sim \mathcal{D}_{a_i}}[l(f^*(x),y)] - \mathbb{E}_{(x,y)\sim \mathcal{D}_{a_j}}[l(f^*(x),y)]| \\
    + M\sqrt{\frac{\log(2k/\delta)}{2n \min_i \pi_i}} + 2L\epsilon
\end{align*}
where $L$ is the Lipschitz constant of the loss function $l$.
\end{theorem}
\begin{proof}
Let $L_i(f) = \mathbb{E}_{(x,y)\sim \mathcal{D}_{a_i}}[l(f(x),y)]$ be the expected loss for demographic group $a_i$ under function $f$. Let $\Delta(f) = \max_{i,j}|L_i(f) - L_j(f)|$ be the maximum absolute difference in the expected loss, which represents the true fairness error.
\\
First, we bound the difference between $\Delta(\hat{f})$ and $\Delta(f^*)$ using the triangle inequality and the Lipschitz continuity of the loss function.
\begin{align*}
    |\Delta(\hat{f})-\Delta(f^*)| &\le \max_{i,j}|(L_i(\hat{f})-L_j(\hat{f}))-(L_i(f^*)-L_j(f^*))| \\
    &\le \max_{i,j}(|L_i(\hat{f})-L_i(f^*)|+|L_j(\hat{f})-L_j(f^*)|)
\end{align*}
For any group $i$, the difference in expected loss is bounded by the Lipschitz property:
\begin{align*}
    |L_i(\hat{f})-L_i(f^*)| &= |\mathbb{E}_{(x,y)\sim \mathcal{D}_{a_i}}[l(\hat{f}(x),y)-l(f^*(x),y)]| \\
    &\le \mathbb{E}_{(x,y)\sim \mathcal{D}_{a_i}}[|l(\hat{f}(x),y)-l(f^*(x),y)|] \\
    &\le L \cdot \mathbb{E}_{(x,y)\sim \mathcal{D}_{a_i}}[|\hat{f}(x)-f^*(x)|]
\end{align*}
Given the assumption that $\|\hat{f}-f^*\|_{\infty} \le \epsilon$, we have:
\[
|L_i(\hat{f})-L_i(f^*)| \le L\epsilon
\]
Therefore, the approximation error is bounded by:
\[
|\Delta(\hat{f})-\Delta(f^*)| \le 2L\epsilon
\]
The bound in the theorem consists of the optimal fairness error $\Delta(f^*)$, a statistical error term derived from concentration inequalities, and the approximation error term $2L\epsilon$ derived above. By Hoeffding's inequality, the deviation between empirical estimates and true expectations for all $k$ groups can be bounded. This gives rise to the statistical error term that depends on the sample size of the smallest group, $n \cdot \min_i \pi_i$.
\\
Combining these components, with probability at least $1-\delta$, we arrive at the final bound:
\[
\Delta(\hat{f}) \le \Delta(f^*) + M\sqrt{\frac{\log(2k/\delta)}{2n \min_i \pi_i}} + 2L\epsilon
\]
This completes the proof.
\end{proof}
\noindent\textbf{Remark}. Theorem \ref{thm:fair_errorbound} provides a crucial theoretical insight by establishing an upper bound on the fairness error for an estimated model $\hat{f}$. This bound elegantly decomposes the total error into three distinct components: the irreducible fairness error of the best possible model, $\Delta(f^*)$; a statistical error term, $M\sqrt{\frac{\log(2k/\delta)}{2n \min_i \pi_i}}$, which quantifies the uncertainty from using a finite sample; and an approximation error, $2L\epsilon$, which accounts for the precision of the optimization algorithm. The statistical error term is particularly instructive, as it shows the bound is fundamentally governed by $\min_i \pi_i$, the proportion of the smallest demographic group. This highlights that even a massive dataset can yield poor fairness guarantees if one group is severely underrepresented. In essence, the theorem demonstrates that achieving fairness requires a multi-pronged approach: collecting large and representative data from all demographic groups, while also deploying powerful optimization algorithms to find the best possible solution.

\begin{definition}[$\epsilon$-optimal Solution]
Let $\Delta(f) = \max_{i,j} \left| \mathbb{E}_{(x,y)\sim D_{a_i}}[\ell(f(x), y)] - \mathbb{E}_{(x,y)\sim D_{a_j}}[\ell(f(x), y)] \right|$ denote the fairness error of a function $f \in \mathcal{F}$. 
Let $f^*$ be the optimal fairness minimizer as defined in Theorem~\ref{thm:fair_errorbound}, i.e.,
\[
f^* = \arg\min_{f \in \mathcal{F}} \Delta(f).
\]
We say that $\hat{f} \in \mathcal{F}$ is an $\epsilon$-optimal solution if:
\begin{equation}
\Delta(\hat{f}) - \Delta(f^*) \leq \epsilon
\end{equation}
where $\epsilon > 0$ is the optimality gap. The expanded version is 
\begin{align*}
&\max_{i,j} \left| \mathbb{E}_{(x,y)\sim D_{a_i}}[\ell(\hat{f}(x), y)] - \mathbb{E}_{(x,y)\sim D_{a_j}}[\ell(\hat{f}(x), y)] \right| \leq \\
&\min_{f \in \mathcal{F}} \max_{i,j} \left| \mathbb{E}_{(x,y)\sim D_{a_i}}[\ell(f(x), y)] - \mathbb{E}_{(x,y)\sim D_{a_j}}[\ell(f(x), y)] \right| + \epsilon.
\end{align*}
\end{definition}
$\epsilon$-optimality is particularly useful in practice when finding the exact optimal solution $f^*$ may be computationally intractable.

\begin{assumption}[Lipschitz Continuity]\label{assm:Lipschitz}
Assume that the loss function $\ell$ is Lipschitz continuous with respect to its first argument, i.e., there exists a constant $L > 0$ such that $|\ell(\hat{y}_1, y) - \ell(\hat{y}_2, y)| \leq L|\hat{y}_1 - \hat{y}_2|$ for all $\hat{y}_1, \hat{y}_2, y$.
\end{assumption}
\begin{theorem}[Algorithm Complexity of Fairness Problem]\label{thm:complexity}
Under Assumptions \ref{assm:prevalence} and \ref{assm:Lipschitz}, given the fairness problem as defined in Definition \ref{eqn:def_fairness} with $k$ demographic groups and a function class $\mathcal{F}$ with finite Vapnik-Chervonenkis (VC) dimension $d$, there exists an algorithm that finds an $\epsilon$-optimal solution $\hat{f}$ to the fairness problem with probability at least $1 - \delta$, using $O(\frac{k^2}{\epsilon^2}(d\log(k/\epsilon) + \log(k/\delta)))$ samples and $O(k^2|\mathcal{F}|)$ time complexity.
\end{theorem}
\begin{proof}
Let $f^{*} = \arg\min_{f \in \mathcal{F}} \Delta(f)$ be the optimal solution to the fairness problem, where $\Delta(f) = \max_{i, j} |L_i(f) - L_j(f)|$ is the maximum absolute difference in the expected loss across all demographic groups.
\begin{lemma}[Uniform Convergence Bound for Fairness Problem]\label{thm:uniform_convergence}
Let $\mathcal{F}$ be a function class with finite VC dimension $d$. For any $\epsilon, \delta > 0$, if the sample size $n$ satisfies $n \geq \frac{8M^2k^2}{\epsilon^2}(d\log(16Mk/\epsilon) + \log(4k^2/\delta))$, then with probability at least $1 - \delta$, the following holds for all $f \in \mathcal{F}$ and all pairs of demographic groups $(i, j)$ simultaneously:
\begin{align*}
|L_i(f) - L_j(f) - (\hat{L}_i(f) - \hat{L}_j(f))| \leq \epsilon/2
\end{align*}
where $\hat{L}_i(f) = \frac{1}{n_i}\sum_{j=1}^{n_i} \ell(f(x_{j}^{(i)}), y_{j}^{(i)})$ is the empirical loss for group $a_i$, and $n_i = nr_i$ is the sample size for group $a_i$.
\end{lemma}
The lemma follows from the standard uniform convergence bound for finite VC dimension function classes, applied to the pairwise differences of the loss functions for each pair of demographic groups. Now, consider the following algorithm: draw a sample of size 
\[
n \geq \frac{8M^2k^2}{\epsilon^2}\left(d\log\left(\frac{16Mk}{\epsilon}\right) + \log\left(\frac{4k^2}{\delta}\right)\right),
\]
then, for each \( f \in \mathcal{F} \), compute the empirical fairness error \( \hat{\Delta}(f) = \max_{i, j} |\hat{L}_i(f) - \hat{L}_j(f)| \), and finally return \( \hat{f} = \arg\min_{f \in \mathcal{F}} \hat{\Delta}(f) \).

By the uniform convergence lemma, with probability at least $1 - \delta$, we have:
\begin{align*}
\Delta(\hat{f}) &\leq \hat{\Delta}(\hat{f}) + \epsilon/2 \leq \hat{\Delta}(f^{*}) + \epsilon/2 \leq \Delta(f^{*}) + \epsilon
\end{align*}
Therefore, the algorithm returns an $\epsilon$-optimal solution with probability at least $1 - \delta$. The sample complexity is $O(\frac{k^2}{\epsilon^2}(d\log(k/\epsilon) + \log(k/\delta)))$, and the time complexity is $O(k^2|\mathcal{F}|)$ since we need to calculate the empirical fairness error for each function in $\mathcal{F}$ and each pair of demographic groups.
\end{proof}
\noindent\textbf{Remark}. Theorem \ref{assm:Lipschitz} sheds light on the sample and time complexity of finding an $\epsilon$-optimal fair solution. The sample complexity grows quadratically with the number of demographic groups $k$ and inversely with the square of the desired accuracy $\epsilon$, as we need to ensure uniform convergence for all pairs of demographic groups. Similarly, the time complexity increases by a factor of $k^2$ due to the pairwise comparisons of the empirical loss functions. These results highlight the challenges in achieving fairness in large-scale applications with multiple demographic groups and complex models.

\begin{lemma}[Symmetrization]\label{thm:symmetrization}
For any function $f \in \mathcal{F}$,
\begin{align*}
\mathbb{P}\left[R(f) - R_{\text{emp}}(f) > \epsilon\right] \leq 2\mathbb{P}\left[\sup_{f \in \mathcal{F}} |R_{\text{emp}}(f) - R'{\text{emp}}(f)| > \frac{\epsilon}{2}\right]
\end{align*}
where $R(f)$ is the true fairness risk, $R_{\text{emp}}(f)$ is the empirical fairness risk on the original sample, and $R'_{\text{emp}}(f)$ is the empirical fairness risk on a ghost sample of size $m$ drawn independently from the same distribution as the original sample.
\end{lemma}
\begin{proof}
Let $S = {(x_1, y_1), \ldots, (x_m, y_m)}$ be the original sample of size $m$ and $S' = {(x'_1, y'_1), \ldots, (x'_m, y'_m)}$ be the ghost sample of size $m$, both drawn independently from the same distribution.
Define the event $A$ as:
\begin{align*}
A = {\exists f \in \mathcal{F}: R(f) - R_{\text{emp}}(f) > \epsilon}
\end{align*}
We want to bound the probability of event $A$. Consider the following event $B$:
\begin{align*}
B = \left\{\sup_{f \in \mathcal{F}} |R_{\text{emp}}(f) - R'_{\text{emp}}(f)| > \frac{\epsilon}{2}\right\}
\end{align*}
We will show that $A \subseteq B \cup B'$, where $B'$ is the same event as $B$ but with the roles of $S$ and $S'$ swapped.
Suppose event $A$ occurs, i.e., there exists a function $f \in \mathcal{F}$ such that $R(f) - R_{\text{emp}}(f) > \epsilon$. Then, we have:
\begin{align*}
R(f) - R_{\text{emp}}(f) &> \epsilon \\
R(f) - R'{\text{emp}}(f) + R'{\text{emp}}(f) - R_{\text{emp}}(f) &> \epsilon \\
\left[R(f) - R'{\text{emp}}(f)\right] + \left[R'{\text{emp}}(f) - R_{\text{emp}}(f)\right] &> \epsilon
\end{align*}
If $R(f) - R'{\text{emp}}(f) \leq \epsilon/2$, then we must have $R'{\text{emp}}(f) - R_{\text{emp}}(f) > \epsilon/2$, which implies event $B$ occurs.
On the other hand, if $R(f) - R'{\text{emp}}(f) > \epsilon/2$, then by swapping the roles of $S$ and $S'$, we have $R'{\text{emp}}(f) - R_{\text{emp}}(f) > \epsilon/2$, which implies event $B'$ occurs.
Therefore, $A \subseteq B \cup B'$, and by the union bound, we have:
\begin{align*}
\mathbb{P}[A] \leq \mathbb{P}[B] + \mathbb{P}[B']
\end{align*}
Since $S$ and $S'$ are drawn independently from the same distribution, we have $\mathbb{P}[B] = \mathbb{P}[B']$. Thus,
\begin{align*}
\mathbb{P}[A] \leq 2\mathbb{P}[B]
\end{align*}
which is equivalent to:
\begin{align*}
\mathbb{P}\left[R(f) - R_{\text{emp}}(f) > \epsilon\right] \leq 2\mathbb{P}\left[\sup_{f \in \mathcal{F}} |R_{\text{emp}}(f) - R'_{\text{emp}}(f)| > \frac{\epsilon}{2}\right]
\end{align*}
This completes the proof.
\end{proof}
\begin{theorem}[Fairness Generalization Bound]\label{thm:generalization}
Under Assumptions \ref{assm:prevalence}, given the fairness problem as defined in Definition \ref{eqn:def_fairness} with $k$ demographic groups and a function space $\mathcal{F}$ with VC dimension $d_{\text{VC}}(\mathcal{F})$, for any $\delta > 0$, with probability at least $1 - \delta$, for all $f \in \mathcal{F}$:
\begin{align*}
&\max_{i, j} |R_i(f) - R_j(f)| \leq \max_{i, j} |R_{\text{emp}, i}(f) - R_{\text{emp}, j}(f)| + \\
&\quad M\sqrt{\frac{8(d_{\text{VC}}(\mathcal{F})\ln(2em/d_{\text{VC}}(\mathcal{F})) + \ln(4k^2/\delta))}{m}}
\end{align*}
where $R_i(f) = \mathbb{E}_{(x, y) \sim \mathcal{D}_{a_i}}[\ell(f(x), y)]$ is the expected fair risk for group $a_i$, $R_{\text{emp}, i}(f) = \frac{1}{m_i}\sum_{j=1}^{m_i} \ell(f(x_{j}^{(i)}), y_{j}^{(i)})$ is the empirical risk for group $a_i$, $m_i$ is the sample size for group $a_i$, and $m = \sum_{i=1}^k m_i$ is the total sample size.
\end{theorem}
\begin{proof}
Let $R(f) = \max_{i, j} |R_i(f) - R_j(f)|$ be the fairness risk and $R_\text{emp}(f) = \max_{i, j} |R_{\text{emp}, i}(f) - R_{\text{emp}, j}(f)|$ be the empirical fairness risk.
Based on Lemma \ref{thm:symmetrization}, by the VC dimension bound in the attached file, we have:
\begin{align*}
\mathbb{P}\left[\sup_{f \in \mathcal{F}} |R_{\text{emp}}(f) - R'_{\text{emp}}(f)| > \frac{\epsilon}{2}\right] \leq 4\Phi(2m)\exp\left(-\frac{m\epsilon^2}{8M^2}\right)
\end{align*}
where $\Phi(m) = \sum_{i=0}^{d_{\text{VC}}(\mathcal{F})} \binom{m}{i}$ is the growth function of $\mathcal{F}$.
Using the bound $\Phi(m) \leq \left(\frac{em}{d_{\text{VC}}(\mathcal{F})}\right)^{d_{\text{VC}}(\mathcal{F})}$ and setting the right-hand side to $\delta/(2k^2)$, we get:
\begin{align*}
\mathbb{P}\left[\sup_{f \in \mathcal{F}} |R_{\text{emp}}(f) - R'_{\text{emp}}(f)| > \frac{\epsilon}{2}\right] \leq \frac{\delta}{2k^2}
\end{align*}
Applying the union bound over all pairs of demographic groups, we have:
\begin{align*}
\mathbb{P}\left[\exists i, j: R_i(f) - R_j(f) > \epsilon\right] \leq \delta
\end{align*}
provided that:
\begin{align*}
\epsilon = M\sqrt{\frac{8(d_{\text{VC}}(\mathcal{F})\ln(2em/d_{\text{VC}}(\mathcal{F})) + \ln(4k^2/\delta))}{m}}
\end{align*}
Therefore, with probability at least $1 - \delta$, for all $f \in \mathcal{F}$:
\begin{align*}
&\max_{i, j} |R_i(f) - R_j(f)| \leq \max_{i, j} |R_{\text{emp}, i}(f) - R_{\text{emp}, j}(f)| + \\
&\quad M\sqrt{\frac{8(d_{\text{VC}}(\mathcal{F})\ln(2em/d_{\text{VC}}(\mathcal{F})) + \ln(4k^2/\delta))}{m}}
\end{align*}
This completes the proof.
\end{proof}
\begin{corollary}[Relationship between Upper Bound and Fairness Risk] \label{thm:bound_and_risk}
Let $B_1$ and $B_2$ be two upper bounds derived from Theorem 3.9 such that $B_1 < B_2$. Then, for any function $f \in \mathcal{F}$, with probability at least $1-\delta$: $\max_{i,j} |R_i(f) - R_j(f)| \leq B_1 < B_2$, where $B_1$ and $B_2$ are of the form: $B_l = \max_{i,j} |R_{emp,i}(f) - R_{emp,j}(f)| + M\sqrt{\frac{8(d_{VC}(\mathcal{F})\ln(2em/d_{VC}(\mathcal{F})) + \ln(4k^2/\delta))}{m_l}}$
for $l \in \{1,2\}$ and $m_1 > m_2$ or $d_{VC,1}(\mathcal{F}) < d_{VC,2}(\mathcal{F})$ or $k_1 < k_2$ or $M_1 < M_2$.
\end{corollary}
\begin{proof}
Let's prove this by analyzing the components of the upper bound from Theorem \ref{thm:generalization}.
First, note that the bound consists of two terms: empirical fairness risk: $\max_{i,j} |R_{emp,i}(f) - R_{emp,j}(f)|$, and complexity term: $M\sqrt{\frac{8(d_{VC}(\mathcal{F})\ln(2em/d_{VC}(\mathcal{F})) + \ln(4k^2/\delta))}{m}}$.
The complexity term is monotonically: decreasing with respect to sample size $m$, increasing with respect to VC dimension $d_{VC}(\mathcal{F})$, increasing with respect to number of groups $k$, or increasing with respect to loss bound $M$.

Therefore, if $B_1$ and $B_2$ are derived under conditions where one or more of: 1) $m_1 > m_2$, 2) $d_{VC,1}(\mathcal{F}) < d_{VC,2}(\mathcal{F})$, 3) $k_1 < k_2$, or 4) $M_1 < M_2$. Then $B_1 < B_2$.

By Theorem \ref{thm:generalization}, we know that with probability at least $1-\delta$:
   $\max_{i,j} |R_i(f) - R_j(f)| \leq B_1$. By transitivity:
   $\max_{i,j} |R_i(f) - R_j(f)| \leq B_1 < B_2$

This completes the proof.
\end{proof}
\noindent\textbf{Remark}. Theorem \ref{thm:generalization} (Fairness Generalization Bound) is a key result that provides an upper bound on the fairness risk of a learned model in terms of its empirical fairness risk, the VC dimension of the function space, the number of demographic groups, and the sample size. This bound is crucial for understanding the generalization performance of fair learning algorithms and the factors that influence their ability to produce equitable models.
The theorem suggests that to achieve a smaller fairness risk (see Corollary \ref{thm:bound_and_risk}), one should have a larger sample size, a smaller VC dimension, and a smaller number of demographic groups. These insights are in line with the well-known bias-complexity trade-off in statistical learning theory \cite{shalev2014understanding}, where models with lower complexity (i.e., smaller VC dimension) tend to have better generalization performance.

\begin{assumption}[Bounded Loss and Lipschitz Continuity]
Assume that the loss function $\ell$ is bounded, i.e., $0 \leq \ell(\hat{y}, y) \leq M$ for some constant $M > 0$. Furthermore, assume that the loss function $\ell$ is Lipschitz continuous with respect to its first argument, i.e., there exists a constant $L > 0$ such that $|\ell(\hat{y}_1, y) - \ell(\hat{y}_2, y)| \leq L|\hat{y}_1 - \hat{y}_2|$ for all $\hat{y}_1, \hat{y}_2, y$.
\end{assumption}

\begin{lemma}[Uniform Convergence of Fairness Risk]\label{thm:convg_fair_risk}
Under Assumption \ref{assm:Lipschitz}, for any $\delta > 0$, with probability at least $1 - \delta$ over the random choice of $S$:
\begin{align*}
\sup_{f \in \mathcal{F}} &|R(f) - R_{\text{emp}}(f, S)| \leq \\
&\frac{2LM}{\sqrt{m}}\left(\sqrt{2d_{\text{VC}}(\mathcal{F})\ln\frac{em}{d_{\text{VC}}(\mathcal{F})}} + \sqrt{2\ln\frac{4}{\delta}}\right)
\end{align*}
\end{lemma}
\begin{proof}
Let $\mathcal{G} = {(x, y) \mapsto \ell(f(x), y) : f \in \mathcal{F}}$ be the function class induced by the loss function $\ell$ and the function class $\mathcal{F}$.
By Assumption \ref{assm:Lipschitz}, the loss function $\ell$ is bounded by $M$ and Lipschitz continuous with constant $L$. For any function $g \in \mathcal{G}$, we have:
\begin{align*}
|g(x, y)| = |\ell(f(x), y)| \leq M
\end{align*}
and for any $(x_1, y_1), (x_2, y_2)$,
\begin{align*}
|g(x_1, y_1) - g(x_2, y_2)| &= |\ell(f(x_1), y_1) - \ell(f(x_2), y_2)| \\
&\leq L|f(x_1) - f(x_2)| \\
&\leq LD|x_1 - x_2|
\end{align*}
where $D$ is the Lipschitz constant of functions in $\mathcal{F}$. Therefore, functions in $\mathcal{G}$ are bounded by $M$ and Lipschitz continuous with constant $LD$.
By McDiarmid's inequality, for any $f \in \mathcal{F}$, with probability at least $1 - \delta/2$,
\begin{align*}
|R(f) - R_{\text{emp}}(f, S)| \leq M\sqrt{\frac{2\ln(2/\delta)}{m}}
\end{align*}
Let $N(\epsilon, \mathcal{F}, |\cdot|\infty)$ be the $\epsilon$-covering number of $\mathcal{F}$ with respect to the $L\infty$ norm. By the Lipschitz continuity of the loss function, we have:
\begin{align*}
N(\epsilon, \mathcal{G}, |\cdot|\infty) \leq N(\epsilon/L, \mathcal{F}, |\cdot|\infty)
\end{align*}
By the VC dimension bound on the covering number \cite{van1997weak},
we have:
\begin{align*}
N(\epsilon/L, \mathcal{F}, |\cdot|\infty) \leq \left(\frac{2eL}{\epsilon}\right)^{d{\text{VC}}(\mathcal{F})}
\end{align*}
By the union bound and the covering number bound, with probability at least $1 - \delta/2$, for all $f \in \mathcal{F}$,
\begin{align*}
|R(f) - R_{\text{emp}}(f, S)| \leq M\sqrt{\frac{2(d_{\text{VC}}(\mathcal{F})\ln(2eL/\epsilon) + \ln(2/\delta))}{m}}
\end{align*}
Setting $\epsilon = LM\sqrt{\frac{2d_{\text{VC}}(\mathcal{F})\ln(em/d_{\text{VC}}(\mathcal{F}))}{m}}$, we get:
\begin{align*}
&\sup_{f \in \mathcal{F}} |R(f) - R_{\text{emp}}(f, S)| \leq \\
&\frac{2LM}{\sqrt{m}}\left(\sqrt{2d_{\text{VC}}(\mathcal{F})\ln\frac{em}{d_{\text{VC}}(\mathcal{F})}} + \sqrt{2\ln\frac{4}{\delta}}\right)
\end{align*}
with probability at least $1 - \delta$ over the random choice of $S$. This completes the proof.
\end{proof}

\begin{theorem}[Convergence of Fairness Risk Minimizer]\label{thm:convergence}
Let $\mathcal{F}$ be a function space with VC dimension $d_{\text{VC}}(\mathcal{F})$ and let $f^* = \arg\min_{f \in \mathcal{F}} R(f)$ be the fairness risk minimizer. Let $\hat{f}_S = \arg\min_{f \in \mathcal{F}} R_{\text{emp}}(f, S)$ be the empirical fairness risk minimizer based on a training set $S$ of size $m$. Then, under Assumptions \ref{assm:prevalence} and \ref{assm:Lipschitz}, for any $\delta > 0$, with probability at least $1 - \delta$ over the random choice of $S$:
\begin{align*}
R(\hat{f}_S) - R(f^*) \leq \frac{2LM}{\sqrt{m}}\left(\sqrt{2d{\text{VC}}(\mathcal{F})\ln\frac{em}{d_{\text{VC}}(\mathcal{F})}} + \sqrt{2\ln\frac{4}{\delta}}\right)
\end{align*}
\end{theorem}
\begin{proof}
The proof relies on Lemma \ref{thm:convg_fair_risk}.
Let $\epsilon = \frac{2LM}{\sqrt{m}}\left(\sqrt{2d_{\text{VC}}(\mathcal{F})\ln\frac{em}{d_{\text{VC}}(\mathcal{F})}} + \sqrt{2\ln\frac{4}{\delta}}\right)$. By the lemma, with probability at least $1 - \delta$ over the random choice of $S$:
\begin{align*}
R(\hat{f}_S) &\leq R_{\text{emp}}(\hat{f}_S, S) + \epsilon \
&\leq R_{\text{emp}}(f^, S) + \epsilon \
&\leq R(f^) + 2\epsilon
\end{align*}
Therefore, with probability at least $1 - \delta$ over the random choice of $S$:
\begin{align*}
R(\hat{f}_S) - R(f^*) \leq \frac{2LM}{\sqrt{m}}\left(\sqrt{2d{\text{VC}}(\mathcal{F})\ln\frac{em}{d_{\text{VC}}(\mathcal{F})}} + \sqrt{2\ln\frac{4}{\delta}}\right)
\end{align*}
This completes the proof.
\end{proof}
\noindent\textbf{Remark}. Theorem \ref{thm:convergence} (Convergence of Fairness Risk Minimizer) is a fundamental result that characterizes the convergence rate of the empirical fairness risk minimizer to the true fairness risk minimizer. The theorem shows that the excess fairness risk, defined as the difference between the fairness risk of the empirical minimizer and the true minimizer, converges to zero at a rate of $O(1/\sqrt{m})$, where $m$ is the sample size.
The convergence rate has important implications for the sample complexity of fair learning algorithms. It suggests that to achieve a desired level of accuracy, the sample size should grow quadratically with the inverse of the desired accuracy. This sample complexity is higher than that of the standard empirical risk minimization \cite{vapnik1999overview}, which has a sample complexity of $O(1/\epsilon^2)$ for an accuracy level of $\epsilon$ \cite{bottou2007tradeoffs}.
Moreover, as the size of the training set increases, the empirical fairness risk minimizer approaches the true fairness risk minimizer, ensuring the convergence of the learning algorithm to a fair solution.

\begin{assumption}[Normal Distribution for Group $a_i$]\label{assm:distribution}
Assume that the data distribution for demographic group $a_i$ follows a normal distribution with mean $\mu_i$ and covariance matrix $\Sigma_i$, i.e., $(x, y) \sim \mathcal{N}(\mu_i, \Sigma_i)$ for $(x, y) \in \mathcal{D}_{a_i}$.
\end{assumption}

\begin{lemma}[Risk Bound for a Mixture of Normal Distributions]\label{thm:risk_bound}
Let the overall data distribution $\mathcal{D}$ be a mixture of $k$ demographic group distributions, $\mathcal{D} = \sum_{j=1}^k \pi_j \mathcal{D}_{a_j}$, where $\mathcal{D}_{a_j} \sim \mathcal{N}(\mu_j, \Sigma_j)$ is the data distribution for group $a_j$ and $\pi_j$ is its proportion in the overall population. Let $f(x,y)$ be a function that is $L_f$-Lipschitz with respect to its inputs $(x,y)$. The difference in expectation between any component distribution $\mathcal{D}_{a_i}$ and the overall mixture distribution $\mathcal{D}$ is bounded as follows:
\begin{align*}
    \begin{split}
        |\mathbb{E}_{(x,y)\sim\mathcal{D}_{a_i}}[f(x,y)] - \mathbb{E}_{(x,y)\sim\mathcal{D}}[f(x,y)]| \le \\
        L_f \sum_{j=1}^k \pi_j \left( ||\mu_i - \mu_j||_2 + \sqrt{||\Sigma_i - \Sigma_j||_F} \right)
    \end{split}
\end{align*}
\end{lemma}
\begin{proof} The expectation over the mixture distribution $\mathcal{D}$ can be written as a weighted sum of the expectations over its components:
$$\mathbb{E}_{(x,y)\sim\mathcal{D}}[f(x,y)] = \sum_{j=1}^k \pi_j \mathbb{E}_{(x,y)\sim\mathcal{D}_{a_j}}[f(x,y)]$$
Since $\sum_{j=1}^k \pi_j = 1$, we can express the expectation for group $a_i$ as $\mathbb{E}_{(x,y)\sim\mathcal{D}_{a_i}}[f(x,y)] = \sum_{j=1}^k \pi_j \mathbb{E}_{(x,y)\sim\mathcal{D}_{a_i}}[f(x,y)]$. The absolute difference in expectations is then:
\begin{align*}
|\mathbb{E}_{\mathcal{D}_{a_i}}[f] - \mathbb{E}_{\mathcal{D}}[f]| &= \left| \sum_{j=1}^k \pi_j \mathbb{E}_{\mathcal{D}_{a_i}}[f] - \sum_{j=1}^k \pi_j \mathbb{E}_{\mathcal{D}_{a_j}}[f] \right| \\
&= \left| \sum_{j=1}^k \pi_j \left( \mathbb{E}_{\mathcal{D}_{a_i}}[f] - \mathbb{E}_{\mathcal{D}_{a_j}}[f] \right) \right|
\end{align*}
By the triangle inequality, this is bounded by:$$
\le \sum_{j=1}^k \pi_j \left| \mathbb{E}_{\mathcal{D}_{a_i}}[f] - \mathbb{E}_{\mathcal{D}_{a_j}}[f] \right|
$$ For each pair of normal distributions $\mathcal{D}_{a_i} \sim \mathcal{N}(\mu_i, \Sigma_i)$ and $\mathcal{D}_{a_j} \sim \mathcal{N}(\mu_j, \Sigma_j)$, the Kantorovich-Rubinstein duality states that $|\mathbb{E}_{\mathcal{D}_{a_i}}[f] - \mathbb{E}_{\mathcal{D}_{a_j}}[f]| \le L_f W_1(\mathcal{D}_{a_i}, \mathcal{D}_{a_j})$, where $W_1$ is the 1-Wasserstein distance. The distance between two normal distributions is bounded by the differences in their means and covariance matrices:$$
W_1(\mathcal{N}(\mu_i, \Sigma_i), \mathcal{N}(\mu_j, \Sigma_j)) \le ||\mu_i - \mu_j||_2 + \sqrt{||\Sigma_i - \Sigma_j||_F}
$$Substituting this into the sum yields the final bound:$$
|\mathbb{E}_{\mathcal{D}_{a_i}}[f] - \mathbb{E}_{\mathcal{D}}[f]| \le L_f \sum_{j=1}^k \pi_j \left( ||\mu_i - \mu_j||_2 + \sqrt{||\Sigma_i - \Sigma_j||_F} \right)
$$
This completes the proof.
\end{proof}
\begin{theorem}[Group-Specific Risk Bound Theorem for Normal Mixture Distributions]\label{thm:group_risk}
Let $\mathcal{F}$ be a function space with VC dimension $d_{VC}(\mathcal{F})$ and let $f_i^{\dagger} = \arg\min_{f \in \mathcal{F}} R_i(f)$ be the risk minimizer for demographic group $a_i$. Let $\hat{f}_S = \arg\min_{f \in \mathcal{F}} R_{emp}(f,S)$ be the empirical risk minimizer based on a training set $S$ of size $m$ drawn from the overall mixture distribution $\mathcal{D} = \sum_{j=1}^k \pi_j \mathcal{D}_{a_j}$. Then, under Assumptions \ref{assm:Lipschitz} and \ref{assm:distribution}, for any $\delta > 0$, with probability at least $1-\delta$ over the random choice of S:
\begin{align*}
\begin{split}
R_i(\hat{f}_S) - R_i(f_i^{\dagger}) \le &\frac{4LM}{\sqrt{m}}\left(\sqrt{2d_{VC}(\mathcal{F})\ln\frac{em}{d_{VC}(\mathcal{F})}} + \sqrt{2\ln\frac{4}{\delta}}\right) \\
&+ 2L \sum_{j=1}^k \pi_j \left( ||\mu_i - \mu_j||_2 + \sqrt{||\Sigma_i - \Sigma_j||_F} \right)
\end{split}
\end{align*}
\end{theorem}
\begin{proof} We decompose the excess risk as follows:
\begin{align*}
    \begin{split}
        R_i(\hat{f}_S) - R_i(f_i^{\dagger}) &= \left( R_i(\hat{f}_S) - R(\hat{f}_S) \right) + \left( R(\hat{f}_S) - R(f_i^{\dagger}) \right) \\
        &+ \left( R(f_i^{\dagger}) - R_i(f_i^{\dagger}) \right)
    \end{split}
\end{align*}
The first and third terms represent the shift between the group-specific risk $R_i$ and the overall risk $R$. Applying Lemma \ref{thm:risk_bound} to both terms, we can bound their absolute values:
$$|R_i(f) - R(f)| \le L \sum_{j=1}^k \pi_j \left( ||\mu_i - \mu_j||_2 + \sqrt{||\Sigma_i - \Sigma_j||_F} \right)$$
The middle term, $R(\hat{f}_S) - R(f_i^{\dagger})$, can be bounded using standard uniform convergence arguments. Let $\epsilon_m = \frac{2LM}{\sqrt{m}}\left(\sqrt{2d_{VC}(\mathcal{F})\ln\frac{em}{d_{VC}(\mathcal{F})}} + \sqrt{2\ln\frac{4}{\delta}}\right)$. By Lemma \ref{thm:convg_fair_risk}, with probability at least $1-\delta$, we have $\sup_{f\in\mathcal{F}}|R(f)-R_{emp}(f,S)| \le \epsilon_m$.
Following the standard argument for ERM consistency:
\begin{align*}
R(\hat{f}_S) &\le R_{emp}(\hat{f}_S, S) + \epsilon_m \\
&\le R_{emp}(f_i^{\dagger}, S) + \epsilon_m \quad (\text{by definition of } \hat{f}_S) \\
&\le R(f_i^{\dagger}) + \epsilon_m + \epsilon_m = R(f_i^{\dagger}) + 2\epsilon_m
\end{align*}
So, $R(\hat{f}_S) - R(f_i^{\dagger}) \le 2\epsilon_m$. Combining all parts yields:
\begin{align*}
R_i(\hat{f}_S) - R_i(f_i^{\dagger}) &\le |R_i(\hat{f}_S) - R(\hat{f}_S)| + (R(\hat{f}_S) - R(f_i^{\dagger})) \\
&\, + |R(f_i^{\dagger}) - R_i(f_i^{\dagger})| \\
&\le 2\epsilon_m + 2 \times (\text{distribution shift bound})
\end{align*}
Substituting the expressions gives the final result. This completes the proof. 
\end{proof}
\noindent\textbf{Remark}. Theorem \ref{thm:group_risk} provides a crucial group-specific risk bound that explicitly quantifies how the statistical properties of a demographic group's data distribution directly impact its model performance. The bound decomposes the excess risk for group $a_i$ into two interpretable components: a standard statistical learning term that converges at a rate of $O(1/\sqrt{m})$ with sample size, and a distributional shift term proportional to the sum of distances (in mean $\|\mu_i - \mu_j\|_2$ and covariance $\|\Sigma_i - \Sigma_j\|_F$) between group $a_i$ and all other groups $a_j$. This formalizes the intuitive notion that a group whose feature distribution deviates significantly from the overall training population will suffer higher prediction error, even if the model is optimally trained on the aggregate data. The theorem thus provides a theoretical foundation for observed empirical disparities, particularly in contexts like racial categories where distributional differences are often pronounced, and underscores that achieving true equity requires either explicitly accounting for these distributional shifts in the learning objective or ensuring the training data itself is representative and balanced.
\begin{corollary}[Fairness-Accuracy Trade-off]\label{thm:fairness_accuracy}
Let $\mathcal{F}$ be a function space. Let $f^{\dagger} = \arg\min_{f\in\mathcal{F}} R(f)$ be the accuracy risk minimizer over the overall mixture distribution $\mathcal{D}$, and let $f_i^{\dagger} = \arg\min_{f\in\mathcal{F}} R_i(f)$ be the accuracy risk minimizer for a specific demographic group $a_i$. Under the assumptions of a mixture of normal distributions (Assumption \ref{assm:distribution}) and a Lipschitz-continuous loss function (Assumption 3.6), the trade-off between overall accuracy and group-specific accuracy is bounded by:
$$R_i(f^{\dagger}) - R_i(f_i^{\dagger}) \le 2L \sum_{j=1}^k \pi_j \left( ||\mu_i - \mu_j||_2 + \sqrt{||\Sigma_i - \Sigma_j||_F} \right)$$
where $L$ is the Lipschitz constant of the loss function.
\end{corollary}
\begin{proof}
By definition, $f_i^{\dagger}$ minimizes the group-specific risk $R_i(f)$, so the term we wish to bound, $R_i(f^{\dagger}) - R_i(f_i^{\dagger})$, is non-negative.
We can decompose this term by adding and subtracting the overall risk $R(f)$:
\begin{align*}
    \begin{split}
        R_i(f^{\dagger}) - R_i(f_i^{\dagger}) &= \left( R_i(f^{\dagger}) - R(f^{\dagger}) \right) + \left( R(f^{\dagger}) - R(f_i^{\dagger}) \right) \\
        &+ \left( R(f_i^{\dagger}) - R_i(f_i^{\dagger}) \right)
    \end{split}
\end{align*}
By the definition of $f^{\dagger}$ as the minimizer of the overall risk $R(f)$, we know that $R(f^{\dagger}) \le R(f_i^{\dagger})$. This implies that the middle term, $\left( R(f^{\dagger}) - R(f_i^{\dagger}) \right)$, is less than or equal to zero. We can thus bound the expression by dropping this non-positive term:
$$R_i(f^{\dagger}) - R_i(f_i^{\dagger}) \le \left( R_i(f^{\dagger}) - R(f^{\dagger}) \right) + \left( R(f_i^{\dagger}) - R_i(f_i^{\dagger}) \right)$$
This is equivalent to:
$$R_i(f^{\dagger}) - R_i(f_i^{\dagger}) \le |R_i(f^{\dagger}) - R(f^{\dagger})| + |R_i(f_i^{\dagger}) - R(f_i^{\dagger})|$$
Both terms on the right-hand side represent the difference between the expected loss over the group-specific distribution $\mathcal{D}_{a_i}$ and the overall mixture distribution $\mathcal{D}$ for a given function ($f^{\dagger}$ and $f_i^{\dagger}$, respectively). We can apply Lemma \ref{thm:risk_bound} to bound each of these terms:
$$|R_i(g) - R(g)| \le L \sum_{j=1}^k \pi_j \left( ||\mu_i - \mu_j||_2 + \sqrt{||\Sigma_i - \Sigma_j||_F} \right)$$
Applying this bound for both $g=f^{\dagger}$ and $g=f_i^{\dagger}$ and summing the results yields the final inequality:
$$R_i(f^{\dagger}) - R_i(f_i^{\dagger}) \le 2L \sum_{j=1}^k \pi_j \left( ||\mu_i - \mu_j||_2 + \sqrt{||\Sigma_i - \Sigma_j||_F} \right)$$
This completes the proof.
\end{proof}
\begin{theorem}[Expected Loss Bound for a Demographic Group in a Normal Mixture Distribution]\label{thm:accuracy}
Let $\mathcal{D}_{a_i} \sim \mathcal{N}(\mu_i, \Sigma_i)$ be the data distribution for group $a_i$, and let the overall data distribution be the mixture $\mathcal{D} = \sum_{j=1}^k \pi_j \mathcal{D}_{a_j}$. Let $l$ be a loss function that is $L_l$-Lipschitz and bounded by $B$. Then the expected loss of a function $f(\cdot)$ on group $a_i$ is bounded as follows:
\begin{align*}
    \begin{split}
        \mathbb{E}_{(x,y)\sim\mathcal{D}_{a_i}}[l(f(x),y)] &\le \mathbb{E}_{(x,y)\sim\mathcal{D}}[l(f(x),y)] \\
        &+ L_l \sum_{j=1}^k \pi_j \left( ||\mu_i - \mu_j||_2 + \sqrt{||\Sigma_i - \Sigma_j||_F} \right)
    \end{split}
\end{align*}
\end{theorem}
\begin{proof} The proof follows directly from Lemma \ref{thm:risk_bound}. We can write the expected loss on group $a_i$ as:
\begin{align*}
    \begin{split}
    \mathbb{E}_{\mathcal{D}_{a_i}}[l(f(x),y)] &= \mathbb{E}_{\mathcal{D}}[l(f(x),y)] \\
    &+ \left( \mathbb{E}_{\mathcal{D}_{a_i}}[l(f(x),y)]
    - \mathbb{E}_{\mathcal{D}}[l(f(x),y)] \right)
    \end{split}
\end{align*}
By taking the inequality, we get:
\begin{align*}
    \begin{split}
    \mathbb{E}_{\mathcal{D}_{a_i}}[l(f(x),y)] &\le \mathbb{E}_{\mathcal{D}}[l(f(x),y)] \\
    &+ \left| \mathbb{E}_{\mathcal{D}_{a_i}}[l(f(x),y)] - \mathbb{E}_{\mathcal{D}}[l(f(x),y)] \right|
    \end{split}
\end{align*}
Applying the bound from Lemma \ref{thm:risk_bound} to the second term on the right-hand side, where the function is the loss $l(f(x),y)$ with Lipschitz constant $L_l$, immediately gives the desired result. This completes the proof. 
\end{proof}
\noindent\textbf{Remark}. Theorem~\ref{thm:accuracy} supplies the first closed-form fairness certificate that translates readily measurable group statistics into an interpretable upper bound on the excess loss, thereby closing the gap between abstract OOD theory and practical fairness diagnostics.
Specifically, classic domain-adaptation bounds \cite{ben2010theory,muandet2013domain,quinonero2009dataset} rely on the $\mathcal{H}\Delta\mathcal{H}$-divergence (where $\mathcal{H}$ is a hypothesis class) or generic Wasserstein distance and do not isolate the first- and second-order moment mismatches ($\|\mu_{i}-\mu\|_{2}$ and $\|\Sigma_{i}-\Sigma\|_{F}$) that fairness auditors can compute without labels.
More recent distributionally-robust guarantees \cite{sinha2018certifying,duchi2021learning} guard against worst-case perturbations inside a $\chi^{2}$ or Wasserstein ball rather than against a priori fixed demographic strata.

\REVISION{
While Definition 3.1 relies on a continuous loss function to facilitate gradient-based optimization, our framework extends to deployment settings where only discrete decisions (e.g., 0-1 classification) are observable. To adopt the conclusions in such settings, the continuous loss $l$ (e.g., Cross-Entropy) serves as a differentiable, Lipschitz-continuous surrogate that upper-bounds the discrete 0-1 error. Under this interpretation, the bounds derived in Theorems 3.17 and 3.19 provide conservative guarantees: minimizing the distributional shift in feature space minimizes the upper bound on discrete decision disparities. Furthermore, as established in Eq. (2), the expected loss difference is a weighted aggregate of group-specific error rates ($L_{+, a_i}, L_{-, a_i}$); therefore, theoretical constraints on the expected loss translate directly to constraints on observable fairness metrics, such as Equal Opportunity or Equalized Odds—in black-box deployments.
}

\begin{corollary}[Correlation between Expected Loss Bound and Feature Distance]\label{thm:correlation_lossbound_featdist}
Let $f(\cdot)$ be a function that maps an input $x$ to a discriminative feature $z=f(x)$. Let $\bar{z}_i = \mathbb{E}_{(x,y)\sim\mathcal{D}_{a_i}}[f(x)]$ and $\Sigma_{z_i} = \text{Cov}_{(x,y)\sim\mathcal{D}_{a_i}}(f(x))$ be the centroid and covariance matrix of the features for demographic group $a_i$, respectively. Let $l$ be a loss function that is $L_l$-Lipschitz. The expected loss of $f(\cdot)$ on group $a_i$ is bounded as follows:
\begin{align*}
    \begin{split}
        \mathbb{E}_{(x,y)\sim\mathcal{D}_{a_i}}[l(f(x),y)] &\le \mathbb{E}_{(x,y)\sim\mathcal{D}}[l(f(x),y)] \\
        &+ L_l \sum_{j=1}^k \pi_j \left( d(\bar{z}_i, \bar{z}_j) + \sqrt{||\Sigma_{z_i} - \Sigma_{z_j}||_F} \right)
    \end{split}
\end{align*}
where $d(\cdot, \cdot)$ is the Euclidean distance between the centroids.
\end{corollary}
\begin{proof}
The proof is a direct application of Theorem \ref{thm:accuracy} to the feature space representations $z = f(x)$. The theorem provides a bound based on the means ($\mu$) and covariances ($\Sigma$) of the underlying data distributions. We apply this bound to the distributions of the features, $z$.
\\
Substituting the feature space statistics for each group $j$ (i.e., $\mu_{z_j}$, $\Sigma_{z_j}$) into the bound from Theorem \ref{thm:accuracy} gives:
\begin{align*}
    \begin{split}
        \mathbb{E}_{(x,y)\sim\mathcal{D}_{a_i}}[l(z,y)] &\le \mathbb{E}_{(x,y)\sim\mathcal{D}}[l(z,y)] \\
        &+ L_l \sum_{j=1}^k \pi_j \left( ||\mu_{z_i} - \mu_{z_j}||_2 + \sqrt{||\Sigma_{z_i} - \Sigma_{z_j}||_F} \right)
    \end{split}
\end{align*}
By definition, the feature centroid $\bar{z}_i$ is the mean of the feature distribution for group $i$, so $\bar{z}_i = \mu_{z_i}$. Therefore, the Euclidean norm of the difference in means, $||\mu_{z_i} - \mu_{z_j}||_2$, is equivalent to the distance between the feature centroids, $d(\bar{z}_i, \bar{z}_j)$. Substituting this notation gives the final result.
\end{proof}

\section{Fairness-Aware Regularization}
\label{sec:fairness_reg}
Let $\mathcal{G}$ be the set of $k$ demographic groups. For each group $a_i \in \mathcal{G}$, let $f(\cdot)$ be the feature extractor of the model, and $z = f(x) \in \mathbb{R}^d$ be the learned representation of input $x$. Let $\bar{z}_i = \mathbb{E}_{(x,y)\sim \mathcal{D}_{a_i}}[f(x)]$ and $\Sigma_{z_i} = \text{Cov}_{(x,y)\sim \mathcal{D}_{a_i}}(f(x))$ be the centroid and covariance matrix of the features for demographic group $a_i$, respectively. Let $\pi_j > 0$ be the proportion of group $a_j$ in the overall population, such that $\sum_{j=1}^{k} \pi_j = 1$.
\\
Motivated by Theorem \ref{thm:accuracy} and Corollary \ref{thm:correlation_lossbound_featdist}, which bound the expected loss on group $a_i$ as:
\begin{align*}
\begin{split}
\mathbb{E}_{(x,y)\sim \mathcal{D}_{a_i}}[\ell(f(x), y)] &\leq \mathbb{E}_{(x,y)\sim \mathcal{D}}[\ell(f(x), y)] \\
+ L_l &\sum_{j=1}^{k} \pi_j \left( \|\bar{z}_i - \bar{z}_j\|_2 + \sqrt{\|\Sigma_{z_i} - \Sigma_{z_j}\|_F} \right),
\end{split}
\end{align*}
we propose a fairness regularization term that directly minimizes the theoretical upper bound on group-wise loss disparity. This term penalizes the pairwise distributional shifts in the learned feature space between all groups:
\begin{equation}
\mathcal{R}_{\text{fair}}(f) = \sum_{i \in \mathcal{G}} \sum_{j \in \mathcal{G}} \pi_j \left( \underbrace{\|\bar{z}_i - \bar{z}_j\|_2}_{\text{centroid gap}} + \underbrace{\sqrt{\|\Sigma_{z_i} - \Sigma_{z_j}\|_F}}_{\text{covariance gap}} \right)
\label{eq:fair_reg_corrected}
\end{equation}
\\
This formulation explicitly encourages the alignment of feature distributions \emph{between every pair of demographic groups}, thereby minimizing the summation term in the theoretical bound. The overall training objective becomes:
\begin{equation}
\min_{f} \left( \frac{1}{n} \sum_{j=1}^{n} \ell(f(x_j), y_j) + \lambda \cdot \mathcal{R}_{\text{fair}}(f) \right)
\label{eq:overall_objective_corrected}
\end{equation}
where $\lambda > 0$ controls the trade-off between accuracy and fairness.
\\
In practice, the group centroids $\bar{z}_i$ and covariances $\Sigma_{z_i}$ can be estimated using mini-batch statistics or exponential moving averages during training. The gradients of $\mathcal{R}_{\text{fair}}$ with respect to the parameters of $f$ can be computed via backpropagation, enabling end-to-end optimization.
\\
The regularization is directly derived from Corollary \ref{thm:correlation_lossbound_featdist}. By minimizing $\mathcal{R}_{\text{fair}}(f)$, we directly tighten the theoretical upper bound on the expected loss for each demographic group, creating a clear and consistent link between our theoretical analysis and the proposed algorithmic solution.

    \begin{figure*}[!t]
\centering
\begin{adjustbox}{scale=0.7}
\begin{minipage}{\linewidth}
\centering
    \begin{subfigure}{0.33\textwidth}
        \centering
        \includegraphics[width=\linewidth]{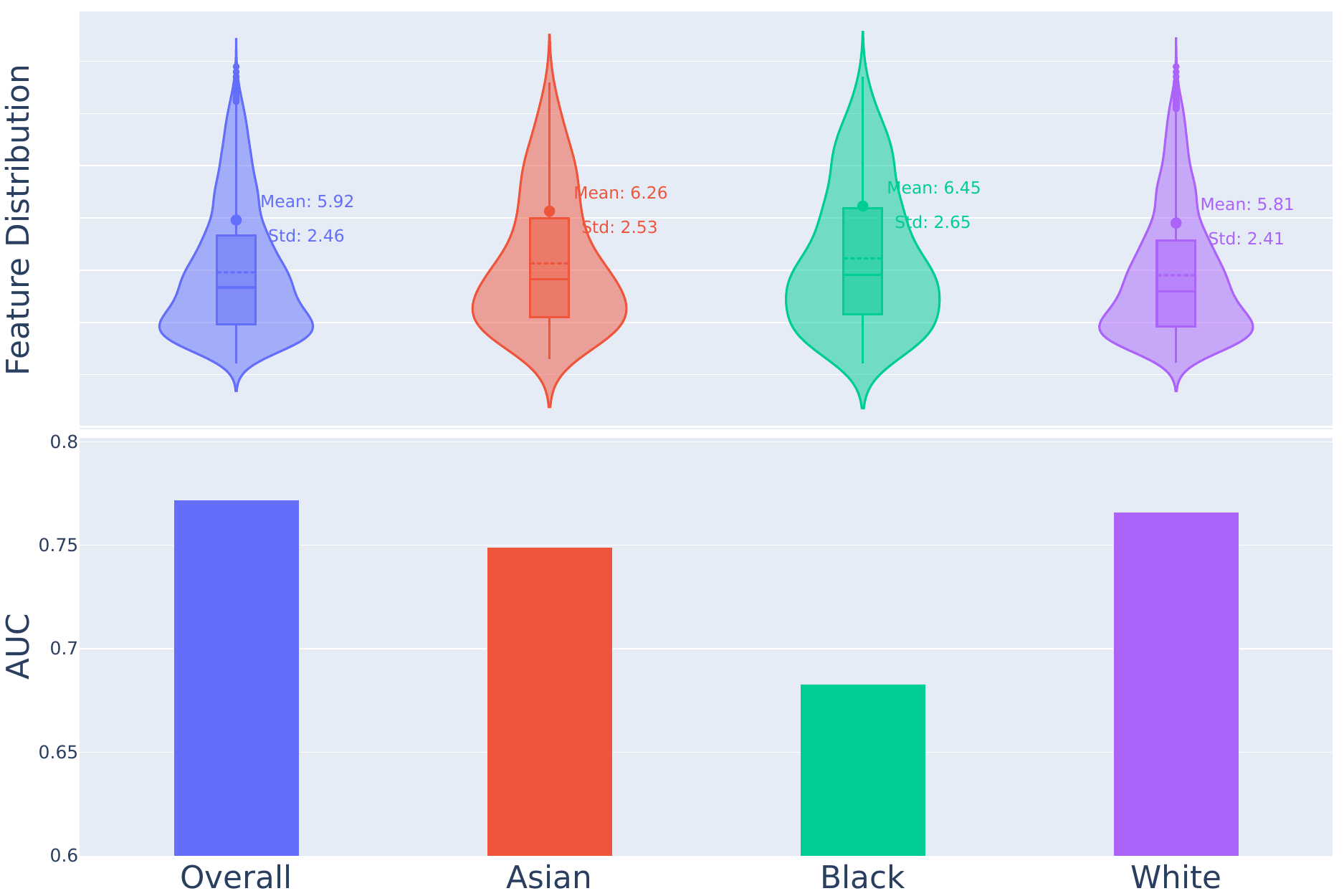}
        \caption{ViT for AMD Det. on Race}
        \label{fig:amd_vit_race}
    \end{subfigure}\hfill
    \begin{subfigure}{0.33\textwidth}
        \centering
        \includegraphics[width=\linewidth]{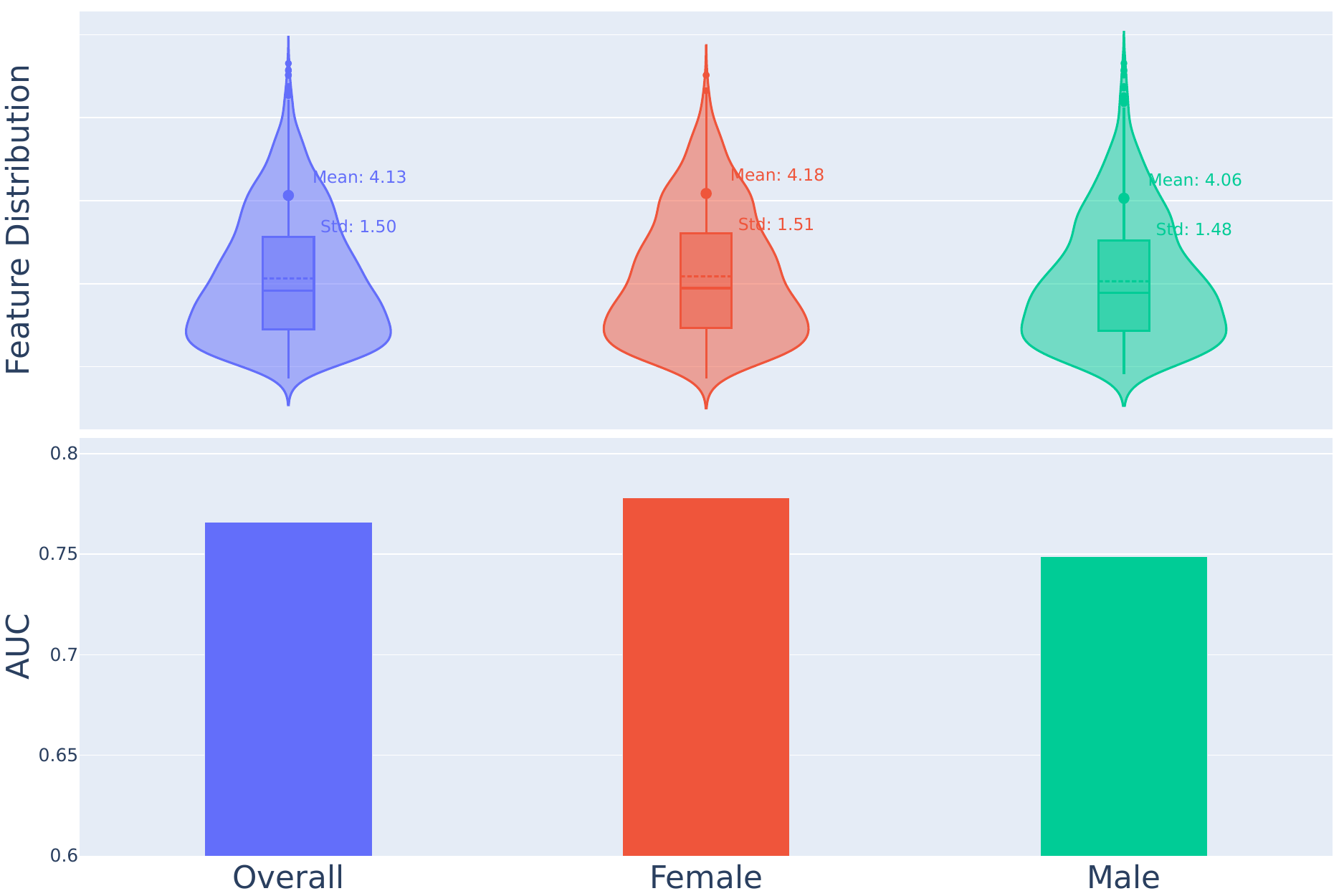}
        \caption{ViT for AMD Det. on Gender}
        \label{fig:amd_vit_gender}
    \end{subfigure}\hfill
    \begin{subfigure}{0.33\textwidth}
        \centering
        \includegraphics[width=\linewidth]{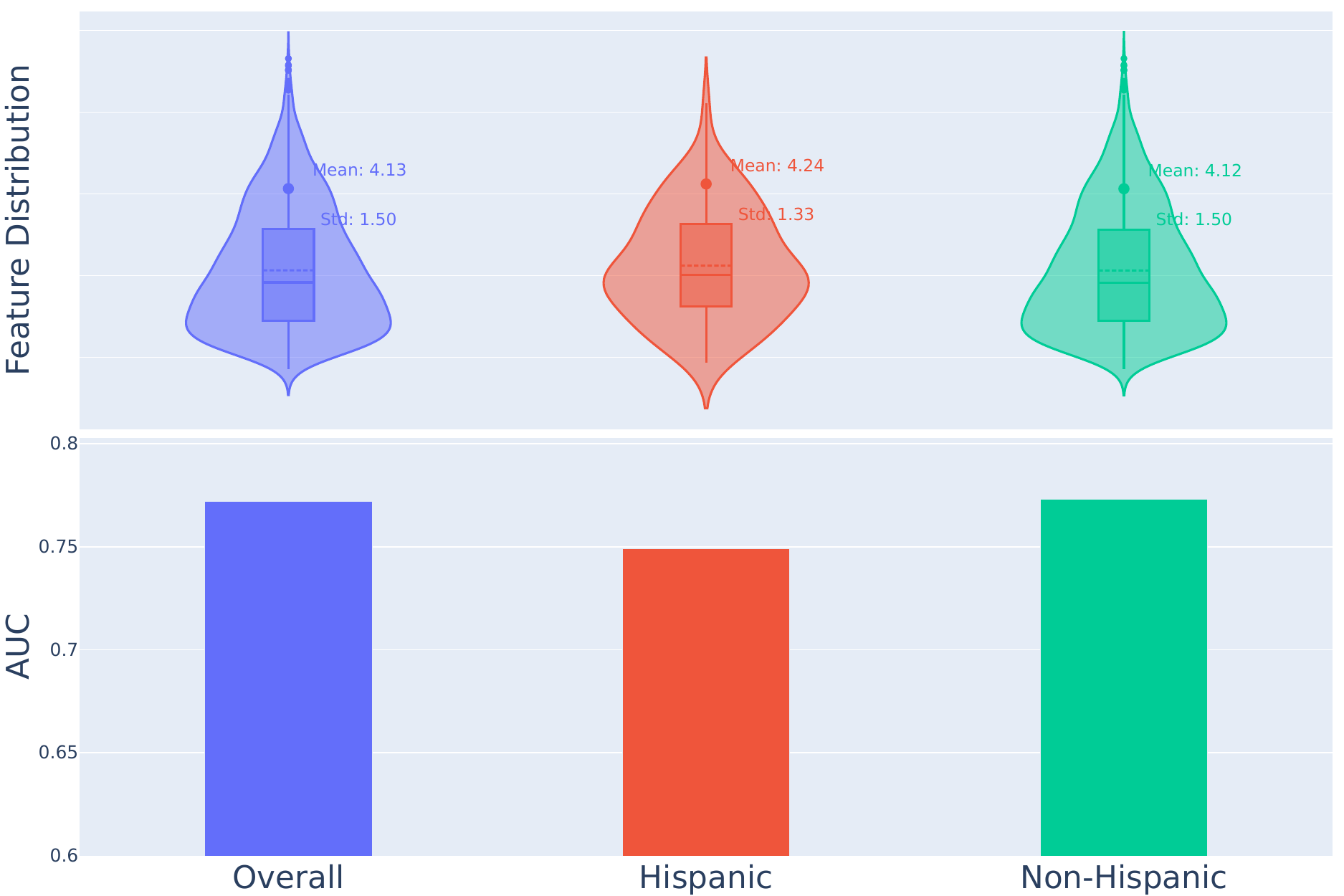}
        \caption{ViT for AMD Det. on Ethnicity}
        \label{fig:amd_vit_ethnicity}
    \end{subfigure}

    \begin{subfigure}{0.33\textwidth}
        \centering
        \includegraphics[width=\linewidth]{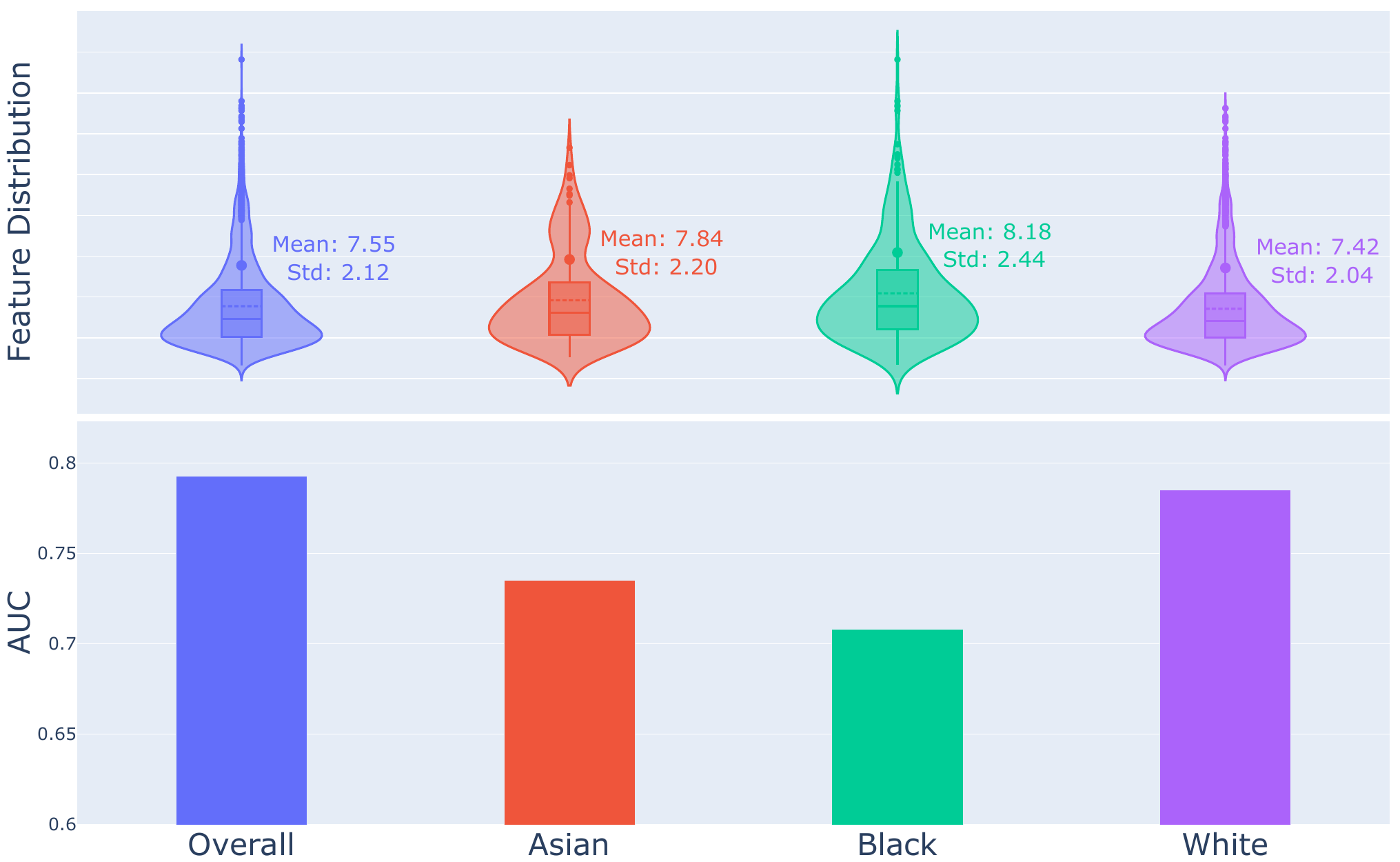}
        \caption{EffNet for AMD Det. on Race}
        \label{fig:amd_eff_race}
    \end{subfigure}\hfill
    \begin{subfigure}{0.33\textwidth}
        \centering
        \includegraphics[width=\linewidth]{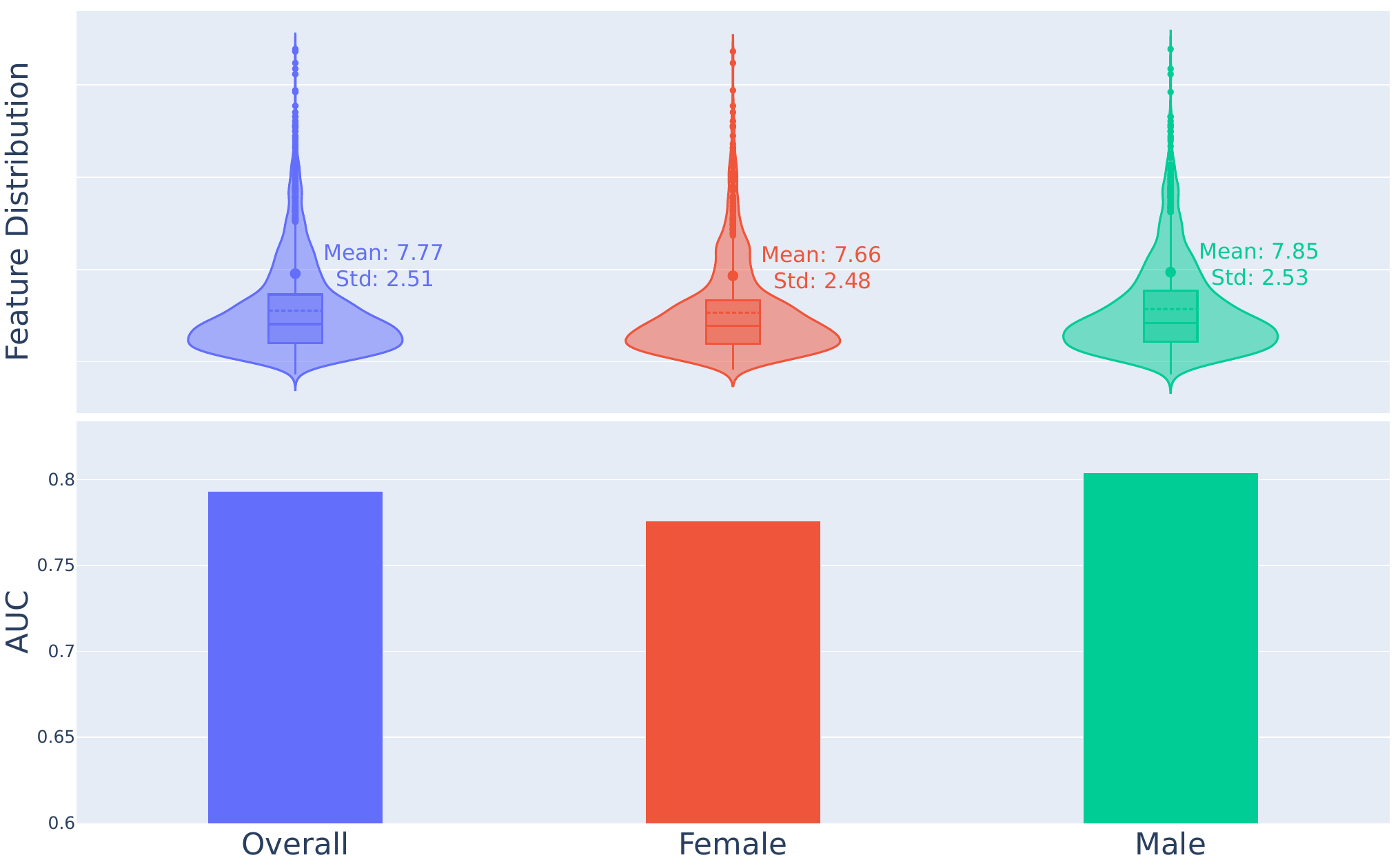}
        \caption{EffNet for AMD Det. on Gender}
        \label{fig:amd_eff_gender}
    \end{subfigure}\hfill
    \begin{subfigure}{0.33\textwidth}
        \centering
        \includegraphics[width=\linewidth]{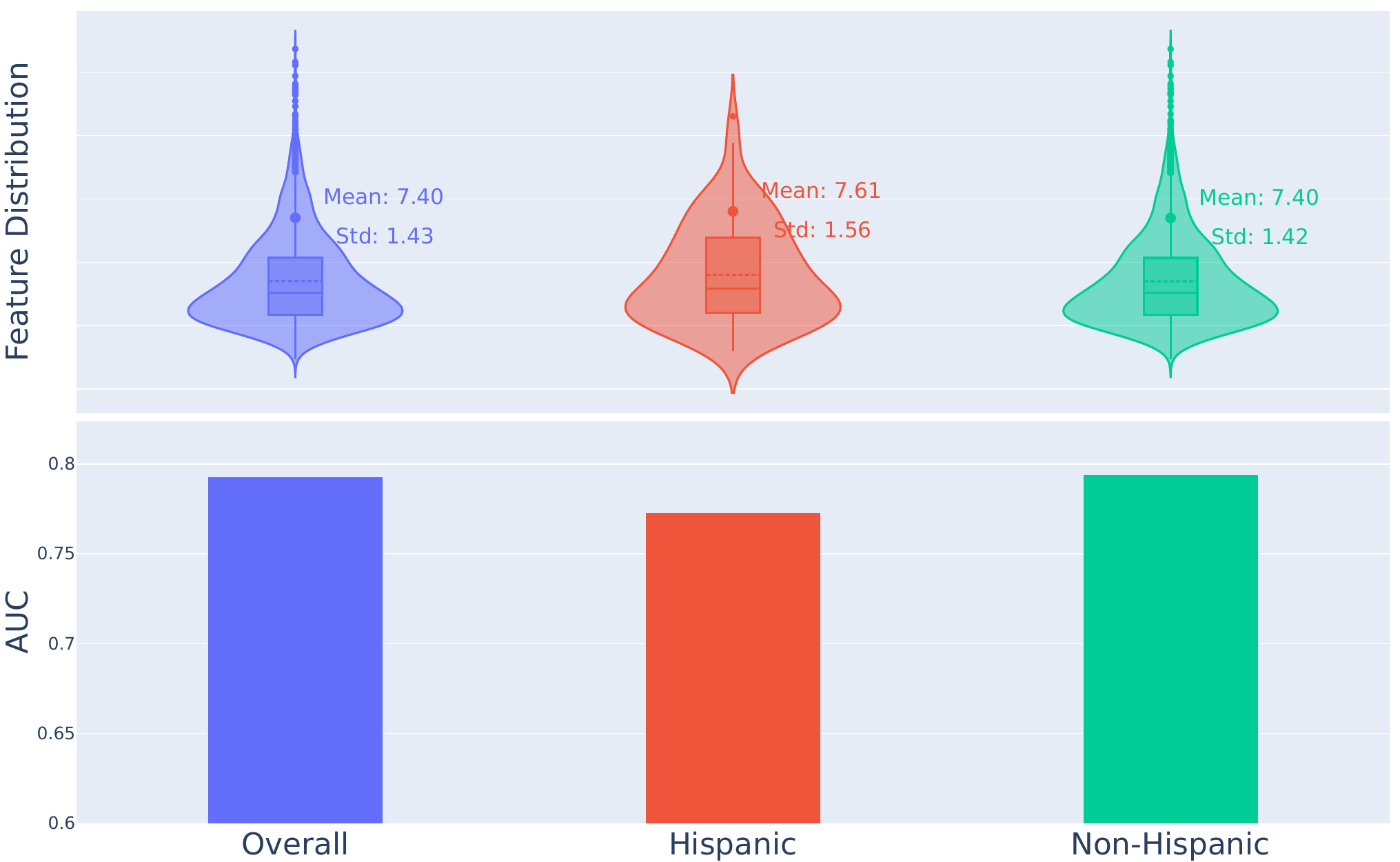}
        \caption{EffNet for AMD Det. on Ethnicity}
        \label{fig:amd_eff_ethnicity}
    \end{subfigure}

    \begin{subfigure}{0.33\textwidth}
        \centering
        \includegraphics[width=\linewidth]{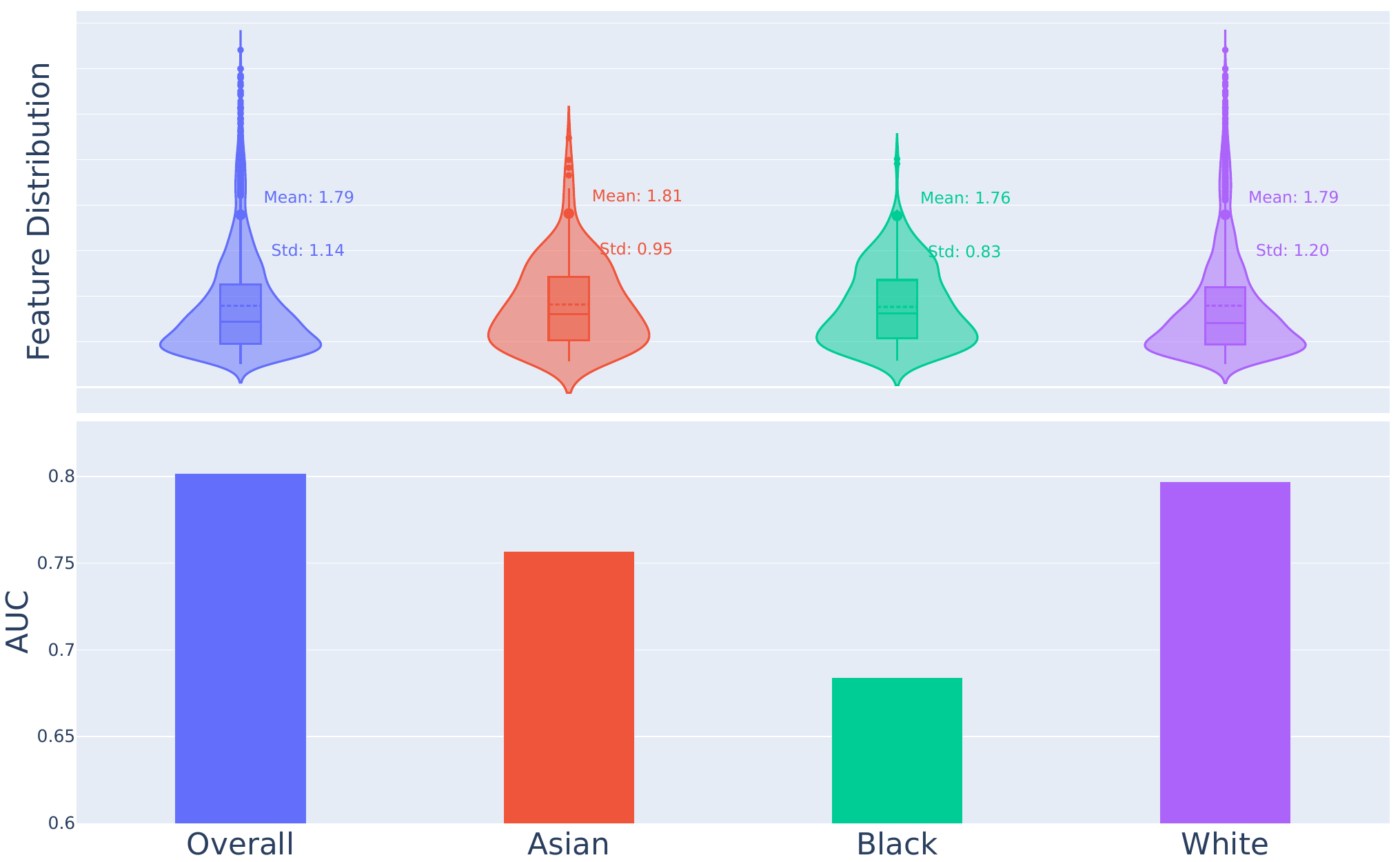}
        \caption{ViT-FAR for AMD Det. on Race}
        \label{fig:amd_vitfar_race}
    \end{subfigure}\hfill
    \begin{subfigure}{0.33\textwidth}
        \centering
        \includegraphics[width=\linewidth]{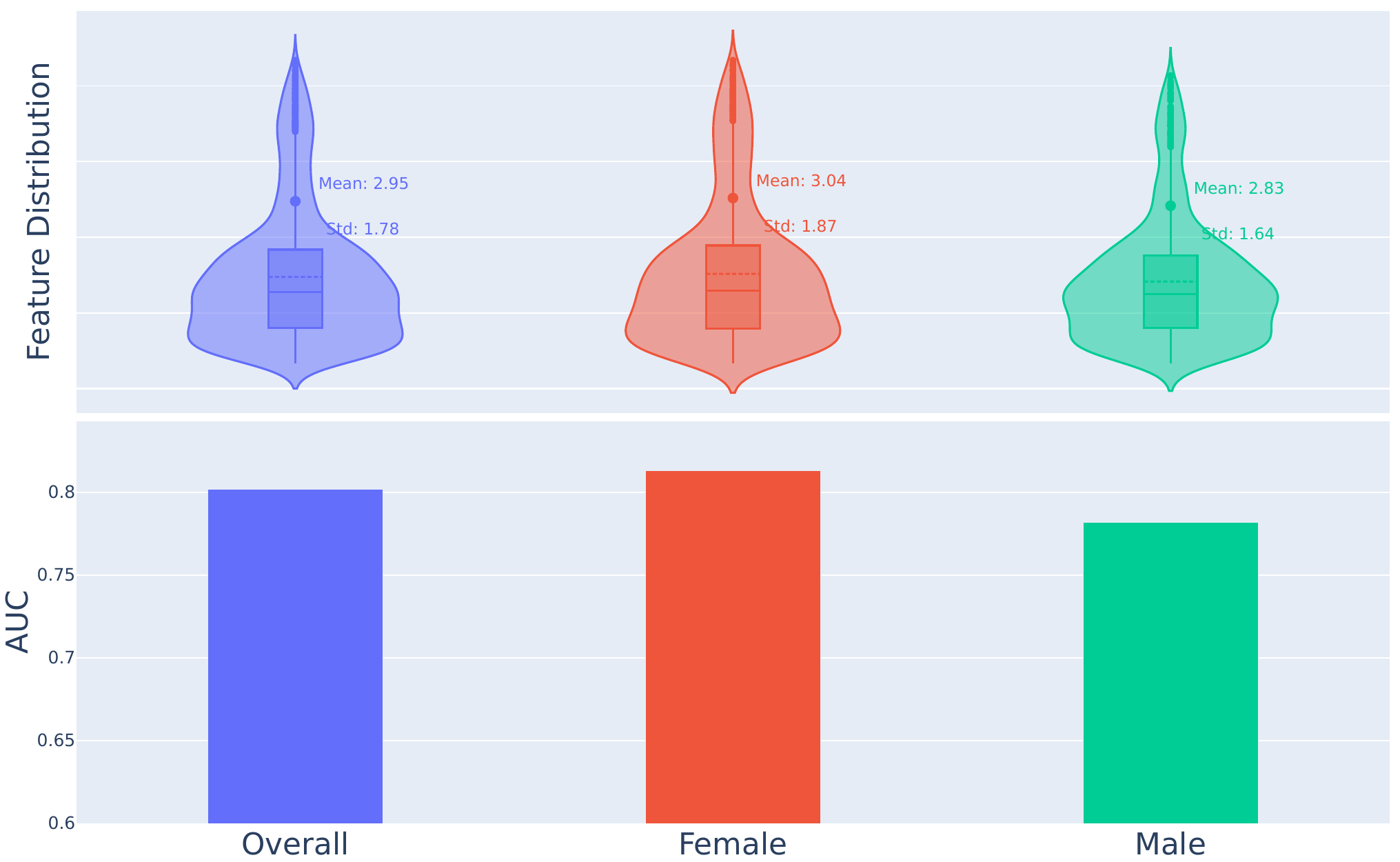}
        \caption{ViT-FAR for AMD Det. on Gender}
        \label{fig:amd_vitfar_gender}
    \end{subfigure}\hfill
    \begin{subfigure}{0.33\textwidth}
        \centering
        \includegraphics[width=\linewidth]{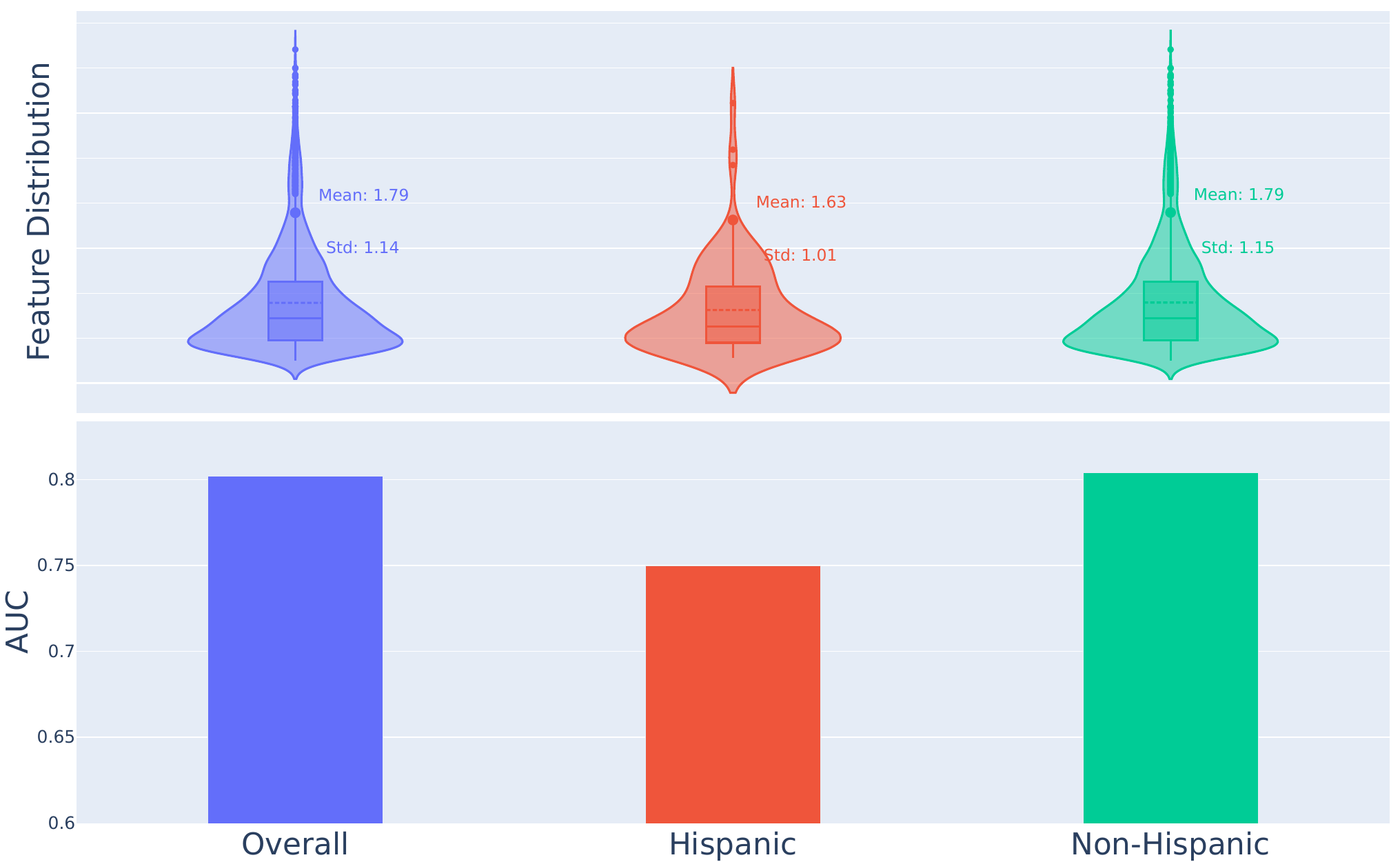}
        \caption{ViT-FAR for AMD Det. on Ethnicity}
        \label{fig:amd_vitfar_ethnicity}
    \end{subfigure}

    \begin{subfigure}{0.33\textwidth}
        \centering
        \includegraphics[width=\linewidth]{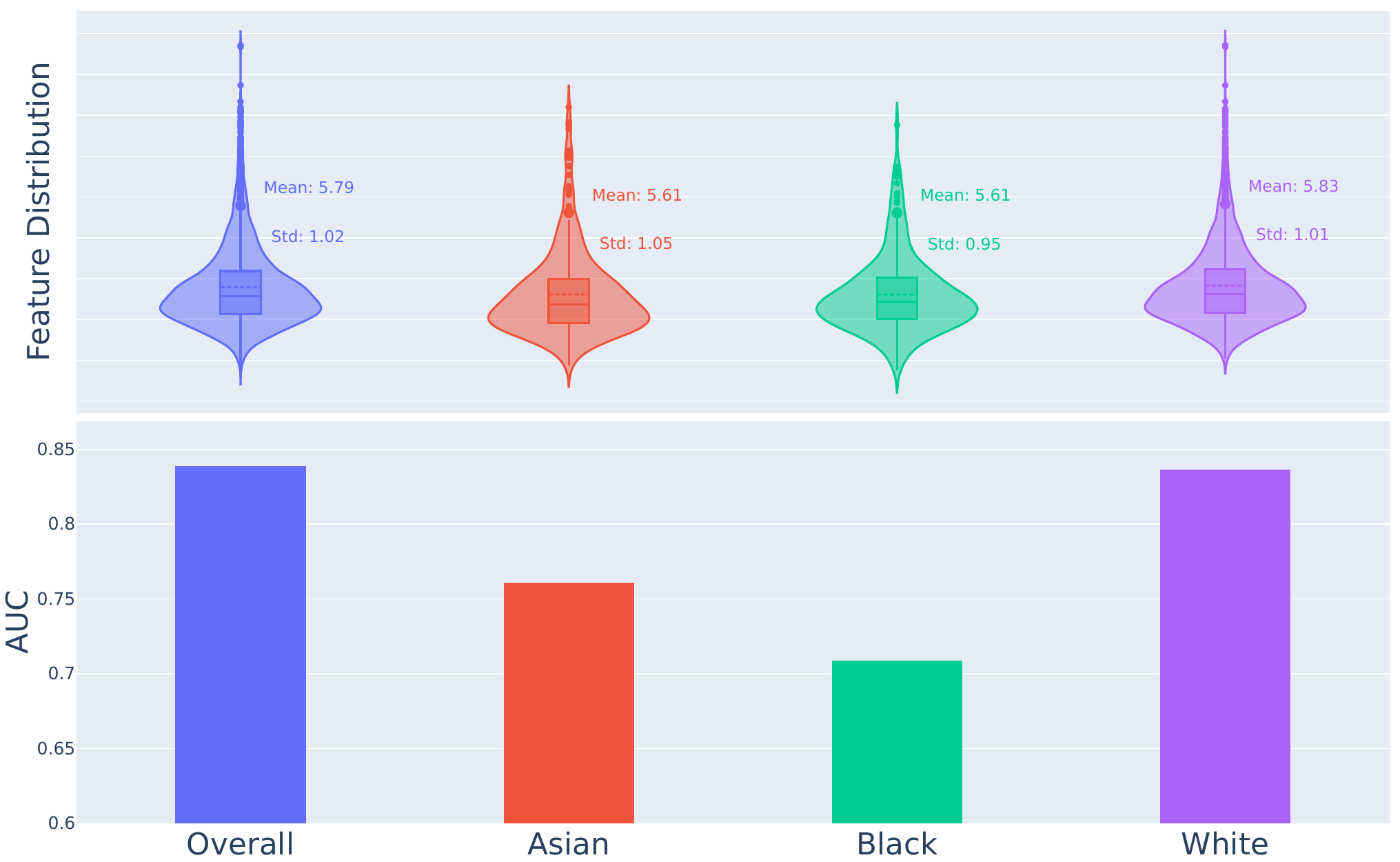}
        \caption{EffNet-FAR for AMD Det. on Race}
        \label{fig:amd_effnetfar_race}
    \end{subfigure}\hfill
    \begin{subfigure}{0.33\textwidth}
        \centering
        \includegraphics[width=\linewidth]{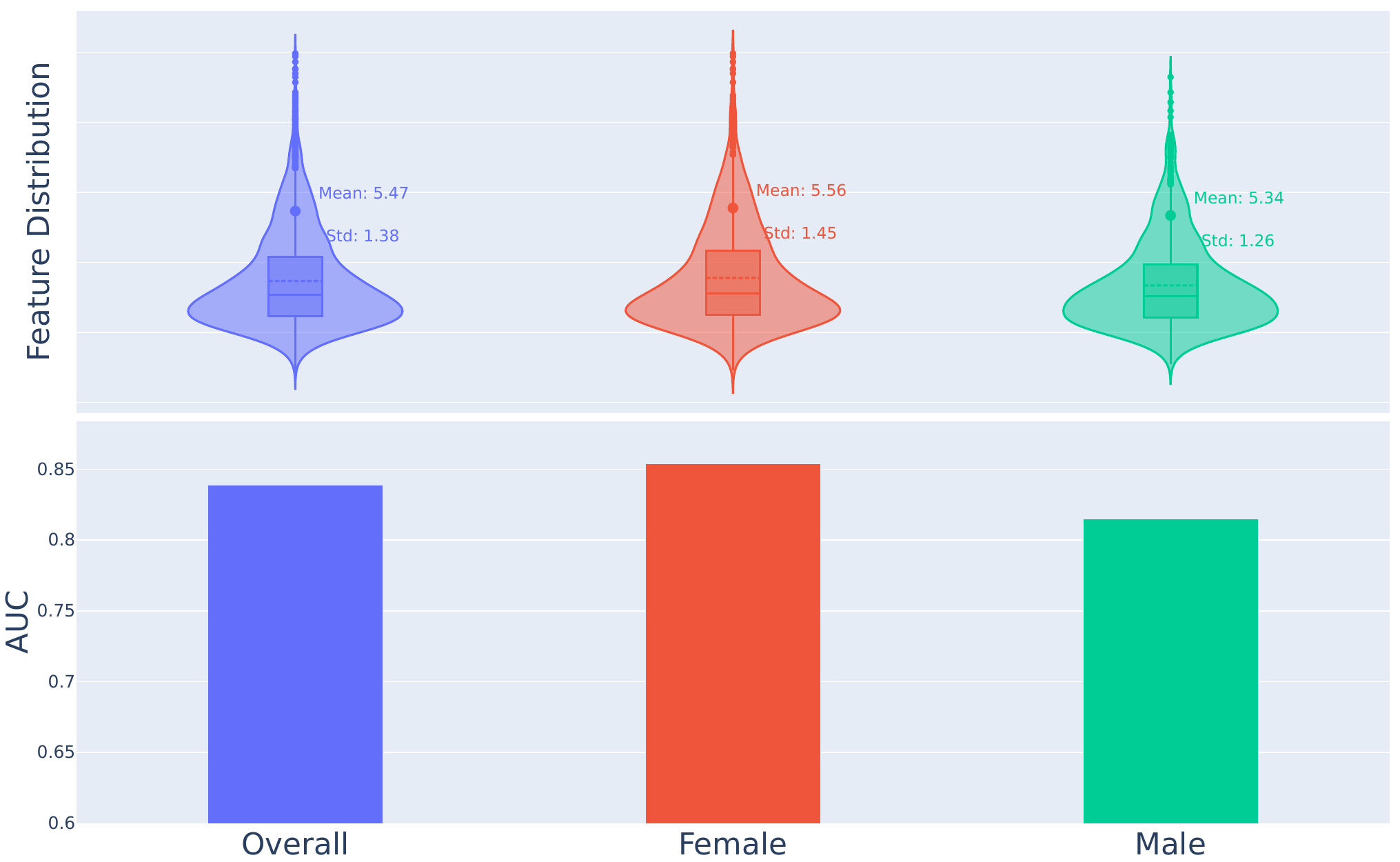}
        \caption{EffNet-FAR for AMD Det. on Gender}
        \label{fig:amd_effnetfar_gender}
    \end{subfigure}\hfill
    \begin{subfigure}{0.33\textwidth}
        \centering
        \includegraphics[width=\linewidth]{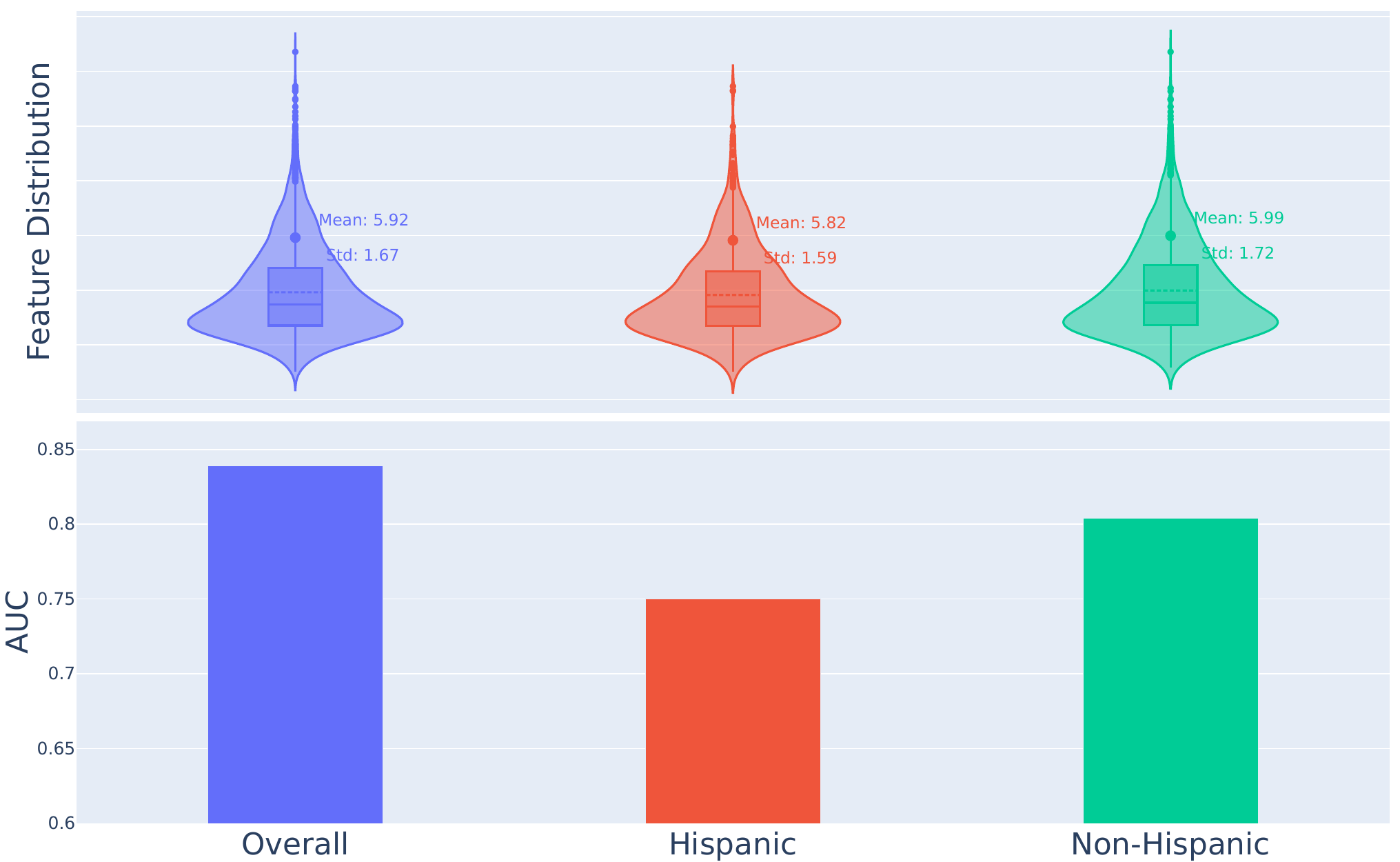}
        \caption{EffNet-FAR for AMD Det. on Ethnicity}
        \label{fig:amd_effnetfar_ethnicity}
    \end{subfigure}

    \begin{subfigure}{0.48\textwidth}
        \centering
        \includegraphics[width=\linewidth]{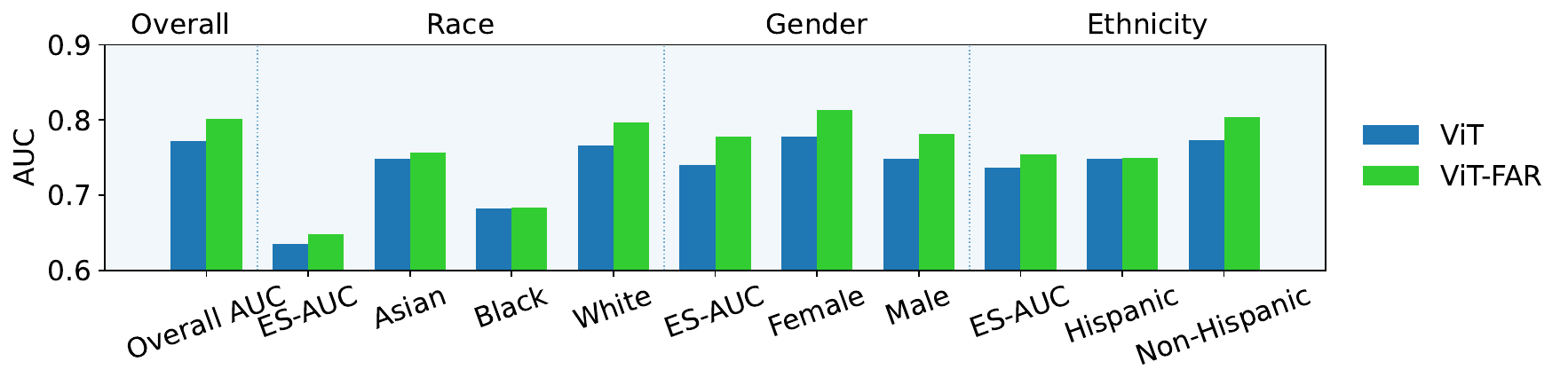}
        \caption{Comparison of ViT and ViT-FAR on AMD Det.}
        \label{fig:amd_vit_comparison}
    \end{subfigure}\hfill
    \begin{subfigure}{0.48\textwidth}
        \centering
        \includegraphics[width=\linewidth]{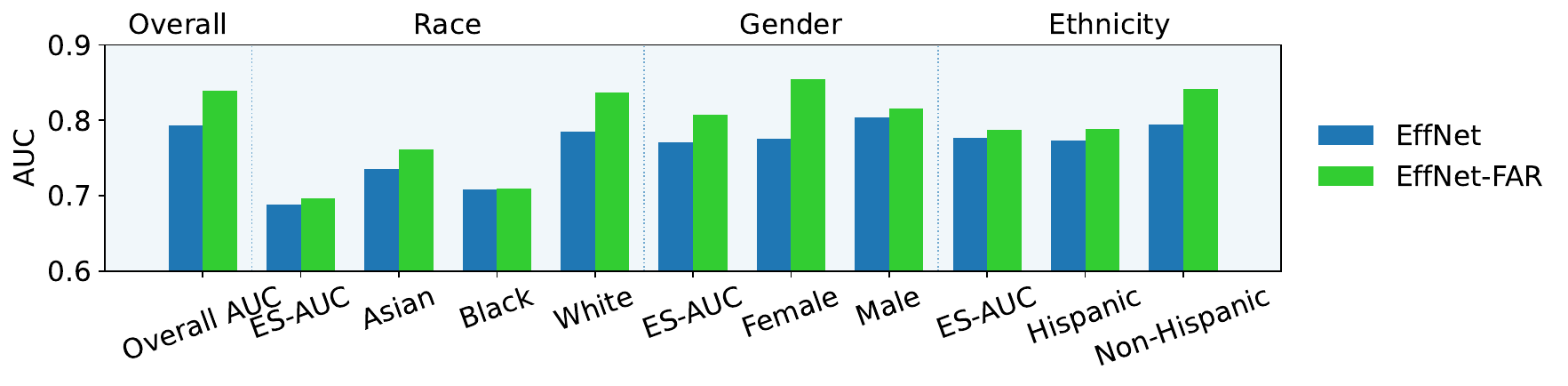}
        \caption{Comparison of EffNet and EffNet-FAR on AMD Det.}
        \label{fig:amd_eff_comparison}
    \end{subfigure}
\end{minipage}
\end{adjustbox}

\caption{Feature distribution, AUC performance, and AUC comparison of ViT, EfficientNet, and their fairness-aware regularization (FAR) variants (ViT-FAR and EffNet-FAR) for AMD detection across three demographic attributes, including Race, Gender, and Ethnicity, on \textbf{FairVision}.}
\label{fig:amd}
\end{figure*}

\section{Experiments}
\label{sec:expr}
\noindent\textbf{Datasets}. \textit{FairVision} contains 30,000 2D scanning laser ophthalmoscopy (SLO) fundus images from 30,000 patients, and each patient has six demographic identity attributes available. This dataset features three common ophthalmic diseases:  Diabetic Retinopathy (DR), Age-related Macular Degeneration (AMD), and Glaucoma, with 10,000 samples for each disease \cite{luo2023fairvision}. According to the official configuration, for each disease, 6,000 samples are used as the training set, 1,000 as the validation set, and 3,000 as the test set. We select the three demographic attributes, including race, gender, and ethnicity, with SLO fundus images as the focus of our study.

\textit{CheXpert}~\cite{irvin2019chexpert} is a large chest X-ray dataset labeled for 14 pathologies, with demographic attributes (age, gender, race). Following~\cite{gichoya2022ai,glocker2023algorithmic}, we use 42,884 patients (127,118 scans): 76,205 training, 12,673 validation, and 38,240 test images. We study pleural effusion detection across race and gender.

\textit{HAM10000} contains 10,015 dermatoscopic images of pigmented skin lesions collected over 20 years. After filtering missing attributes as in~\cite{zong2022medfair}, we use 9,948 images, relabeled into benign/malignant categories. We evaluate skin cancer detection with respect to gender and age.

\textit{FairFace} combines 13,000 new images with a reannotated IJB-C subset~\cite{maze2018iarpa}, yielding 152,917 images from ~6,100 identities (100,186 train / 17,138 val / 35,593 test). It includes annotations for gender, skin color, age, and eyeglasses. We treat glasses detection as binary classification, focusing on fairness across age, skin color, and gender.

\textit{ACS Income}~\cite{ding2021retiring} contains 1.66M records from the 2018 U.S. Census ACS, capturing demographic and socioeconomic features. We cast income prediction as binary classification (>$\$50$k/year) and assess fairness across race and gender.

\textit{CivilComments-WILDS} includes 450,000 crowdsourced comments labeled for toxicity (majority vote, $\ge$10 workers per comment). With 269,038 training, 45,180 validation, 133,782 testing samples, we frame toxicity detection as binary classification and evaluate fairness across race and gender.

\begin{figure*}[!h]
\centering
\begin{adjustbox}{scale=0.7}
\begin{minipage}{\linewidth}
\centering
    \begin{subfigure}{0.33\textwidth}
        \centering
        \includegraphics[width=\linewidth]{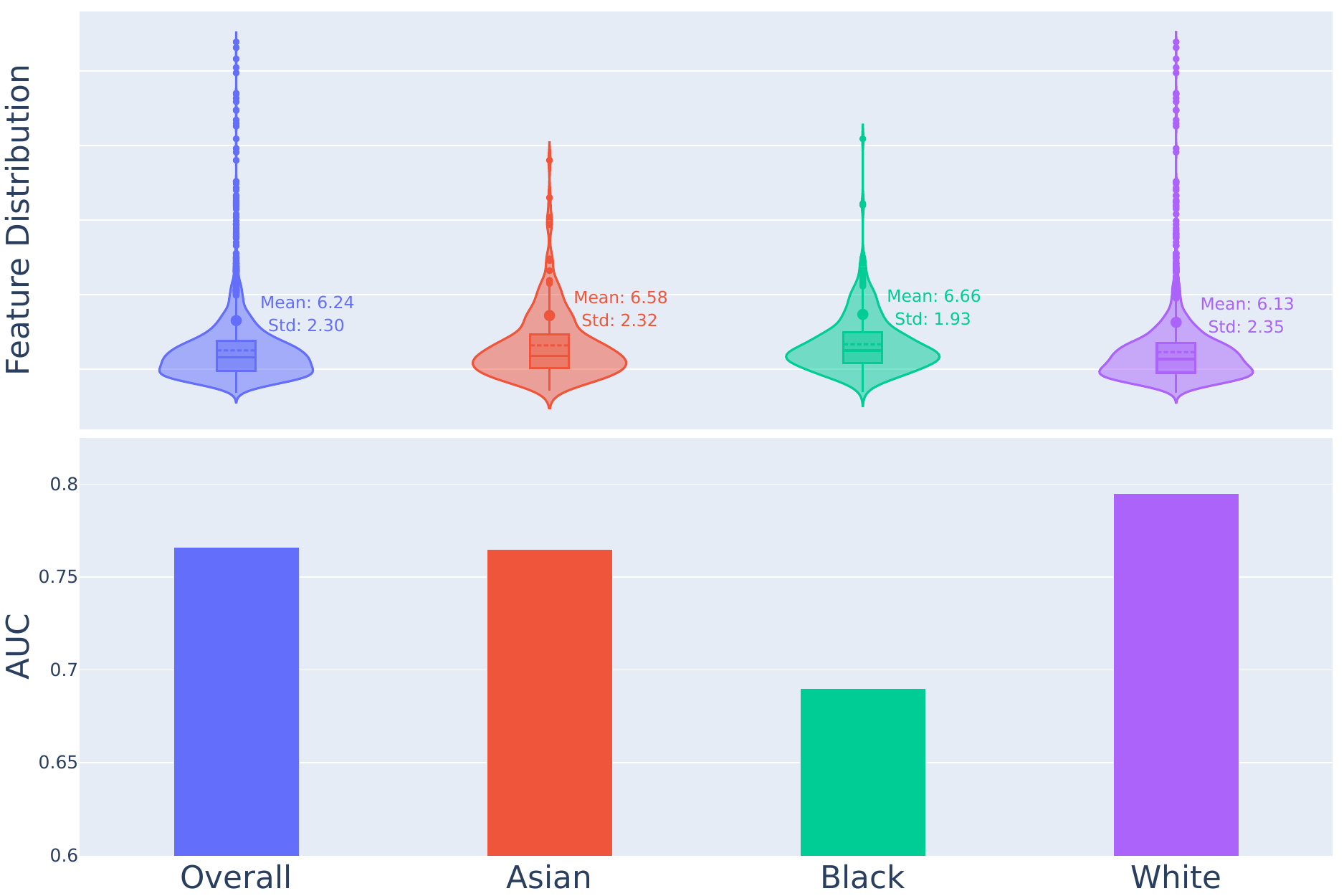}
        \caption{ViT for DR Det. on Race}
        \label{fig:dr_vit_race}
    \end{subfigure}\hfill
    \begin{subfigure}{0.33\textwidth}
        \centering
        \includegraphics[width=\linewidth]{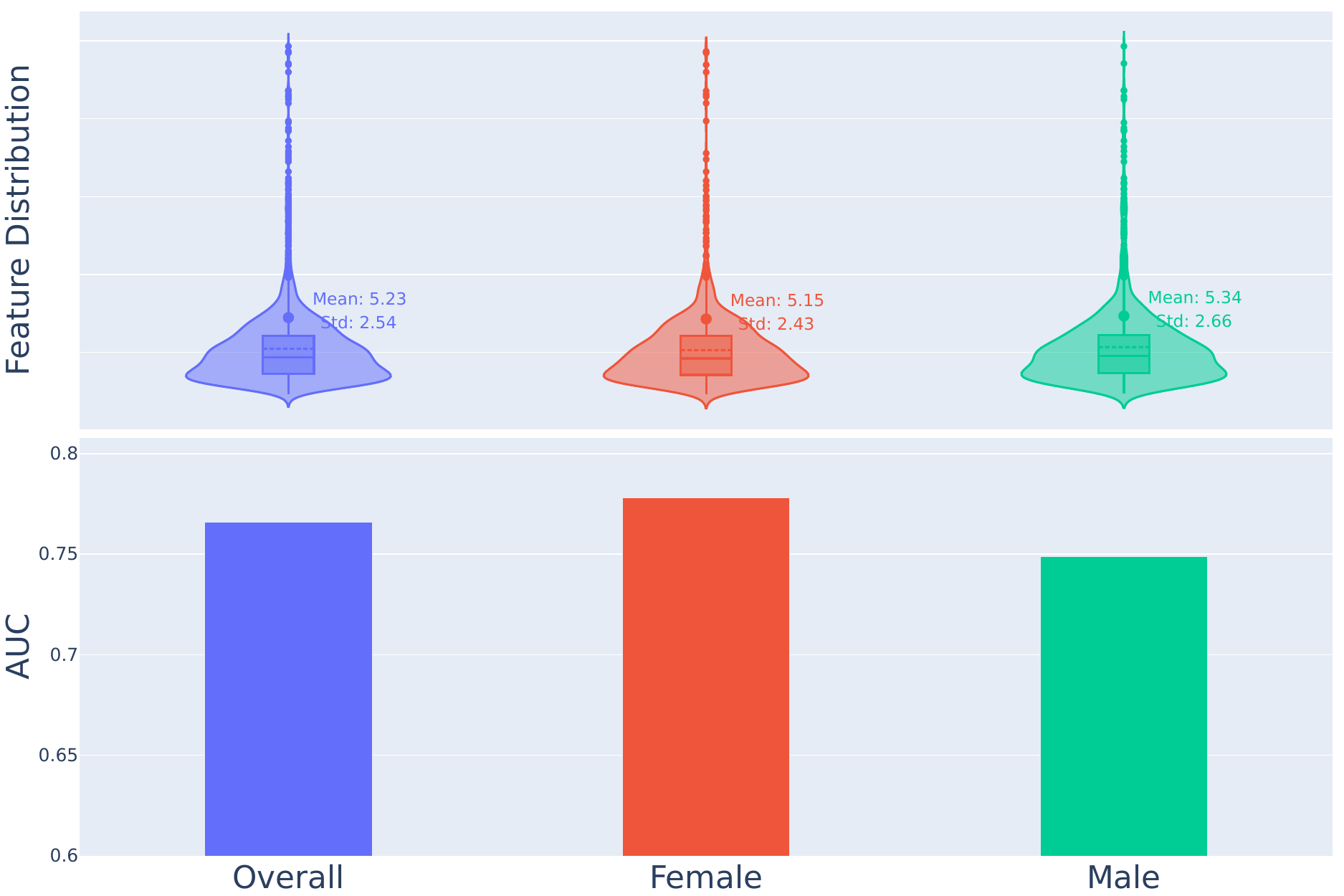}
        \caption{ViT for DR Det. on Gender}
        \label{fig:dr_vit_gender}
    \end{subfigure}\hfill
    \begin{subfigure}{0.33\textwidth}
        \centering
        \includegraphics[width=\linewidth]{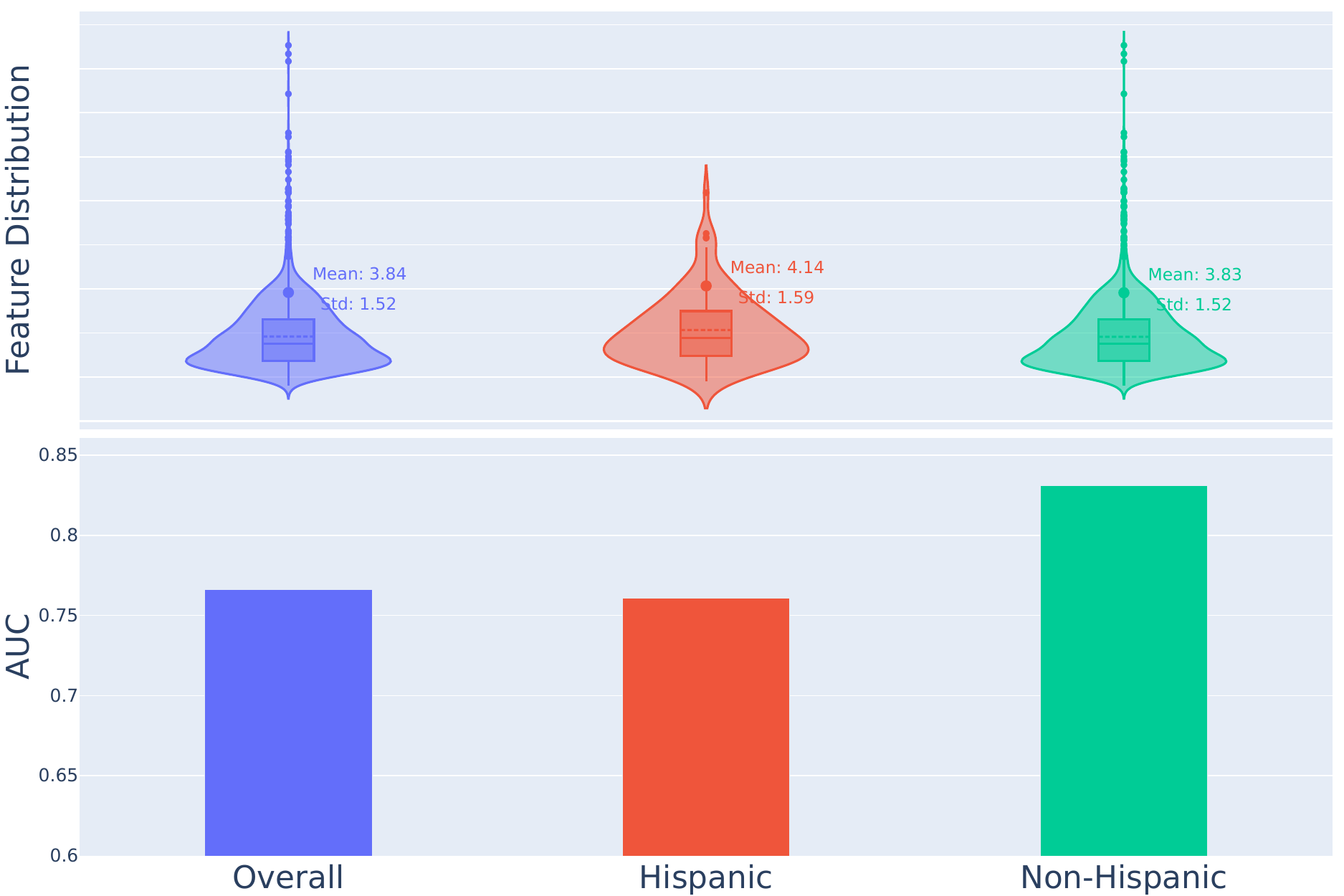}
        \caption{ViT for DR Det. on Ethnicity}
        \label{fig:dr_vit_ethnicity}
    \end{subfigure}

    \begin{subfigure}{0.33\textwidth}
        \centering
        \includegraphics[width=\linewidth]{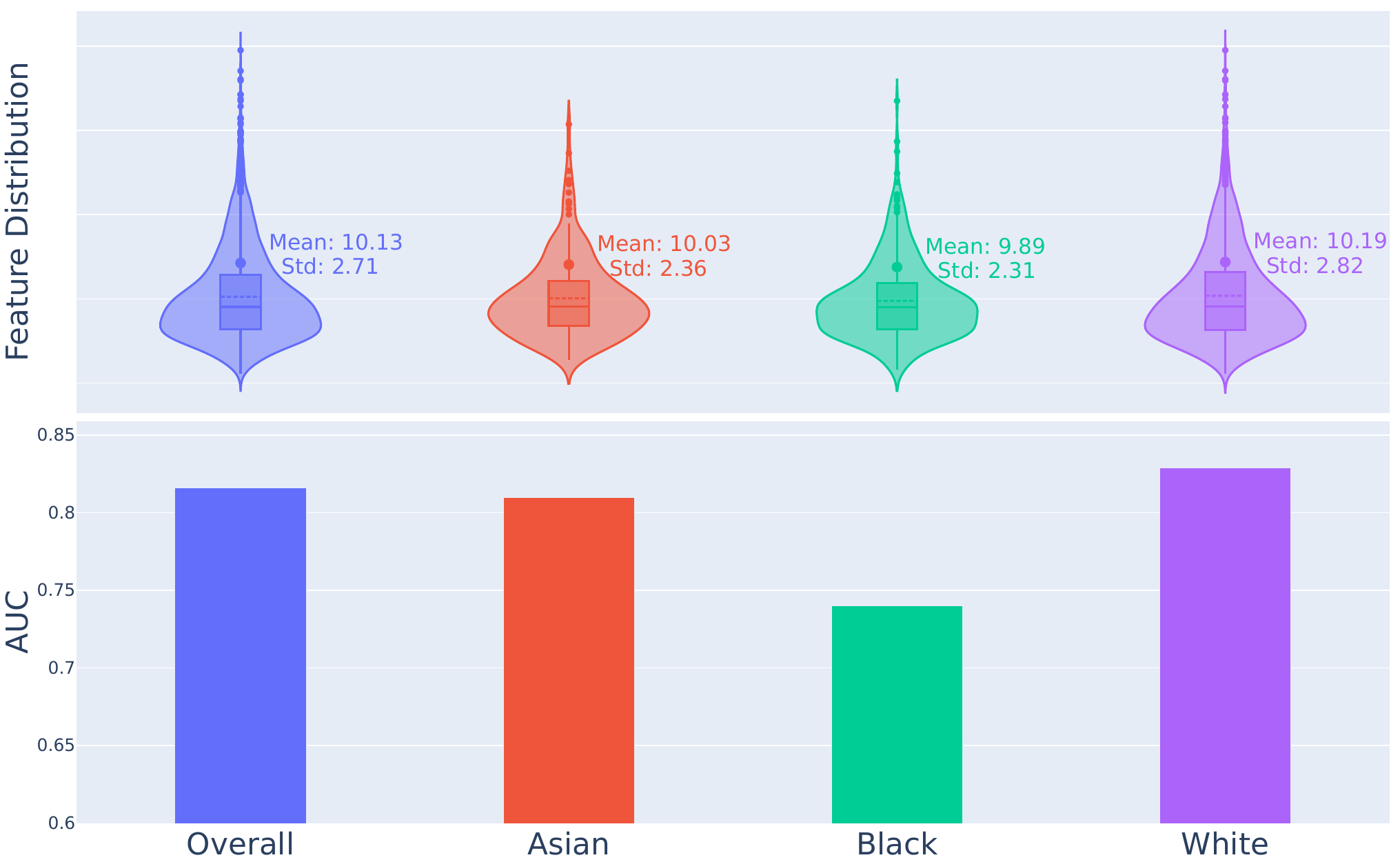}
        \caption{EffNet for DR Det. on Race}
        \label{fig:dr_eff_race}
    \end{subfigure}\hfill
    \begin{subfigure}{0.33\textwidth}
        \centering
        \includegraphics[width=\linewidth]{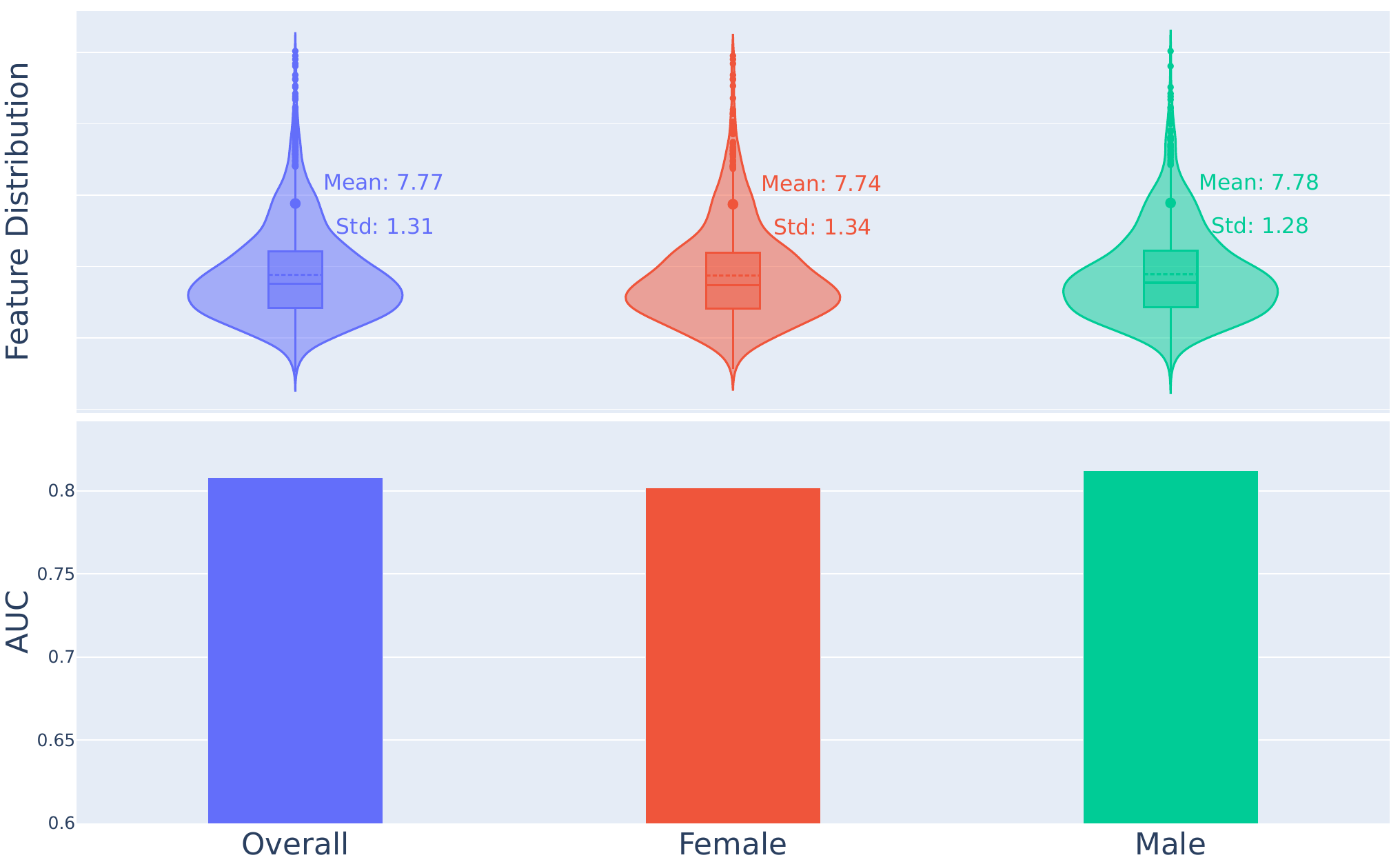}
        \caption{EffNet for DR Det. on Gender}
        \label{fig:dr_eff_gender}
    \end{subfigure}\hfill
    \begin{subfigure}{0.33\textwidth}
        \centering
        \includegraphics[width=\linewidth]{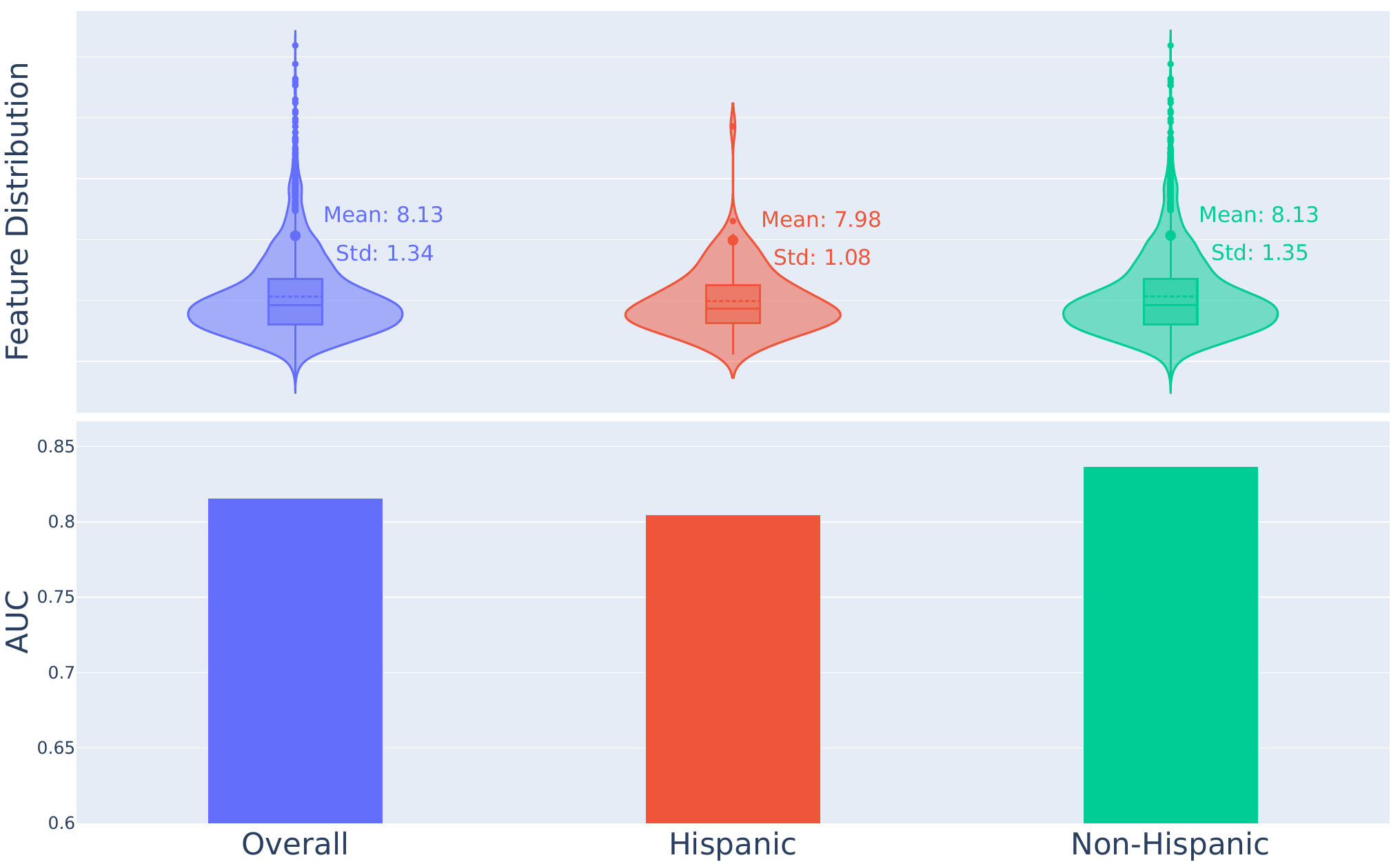}
        \caption{EffNet for DR Det. on Ethnicity}
        \label{fig:dr_eff_ethnicity}
    \end{subfigure}

    \begin{subfigure}{0.33\textwidth}
        \centering
        \includegraphics[width=\linewidth]{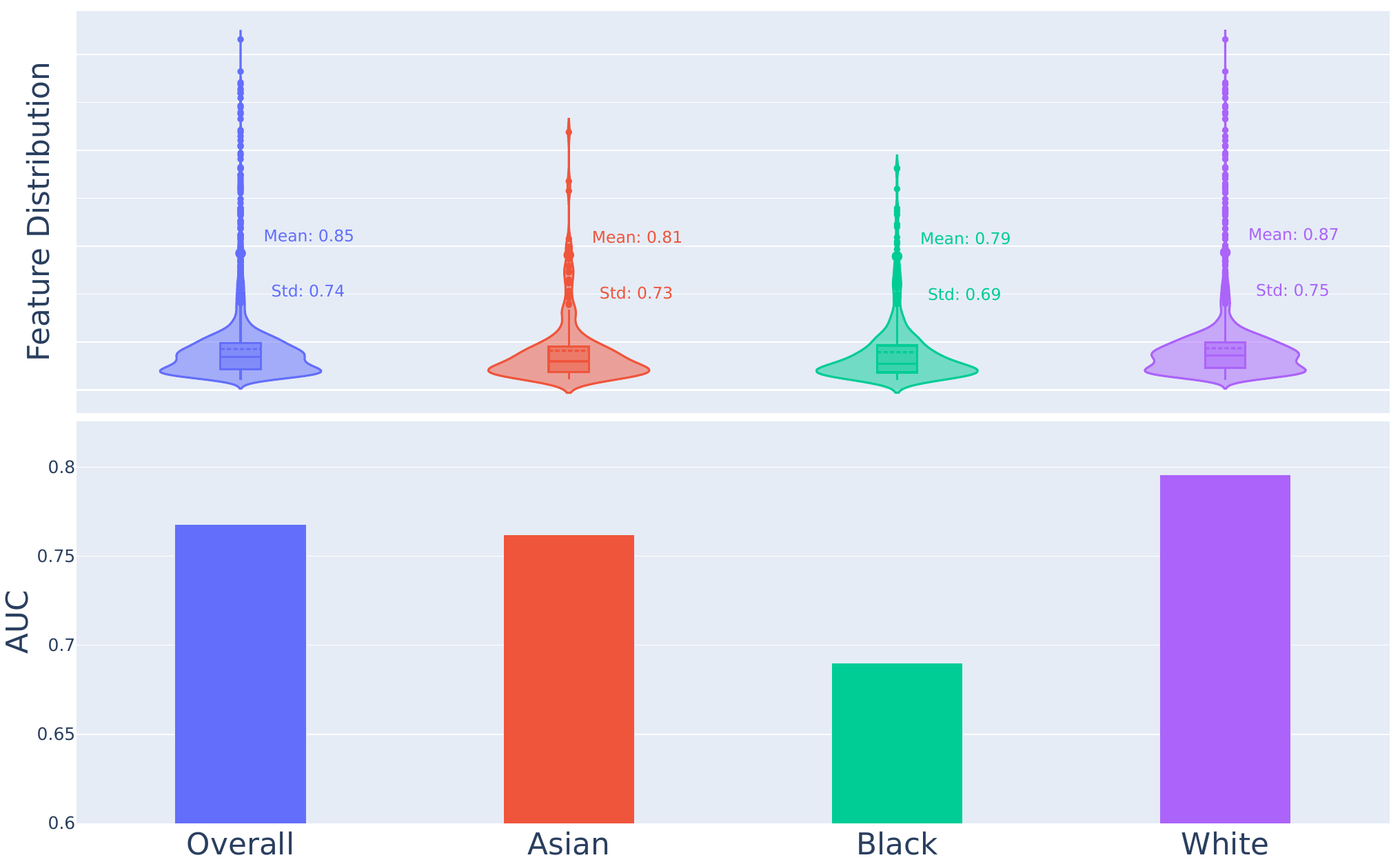}
        \caption{ViT-FAR for DR Det. on Race}
        \label{fig:dr_vitfar_race}
    \end{subfigure}\hfill
    \begin{subfigure}{0.33\textwidth}
        \centering
        \includegraphics[width=\linewidth]{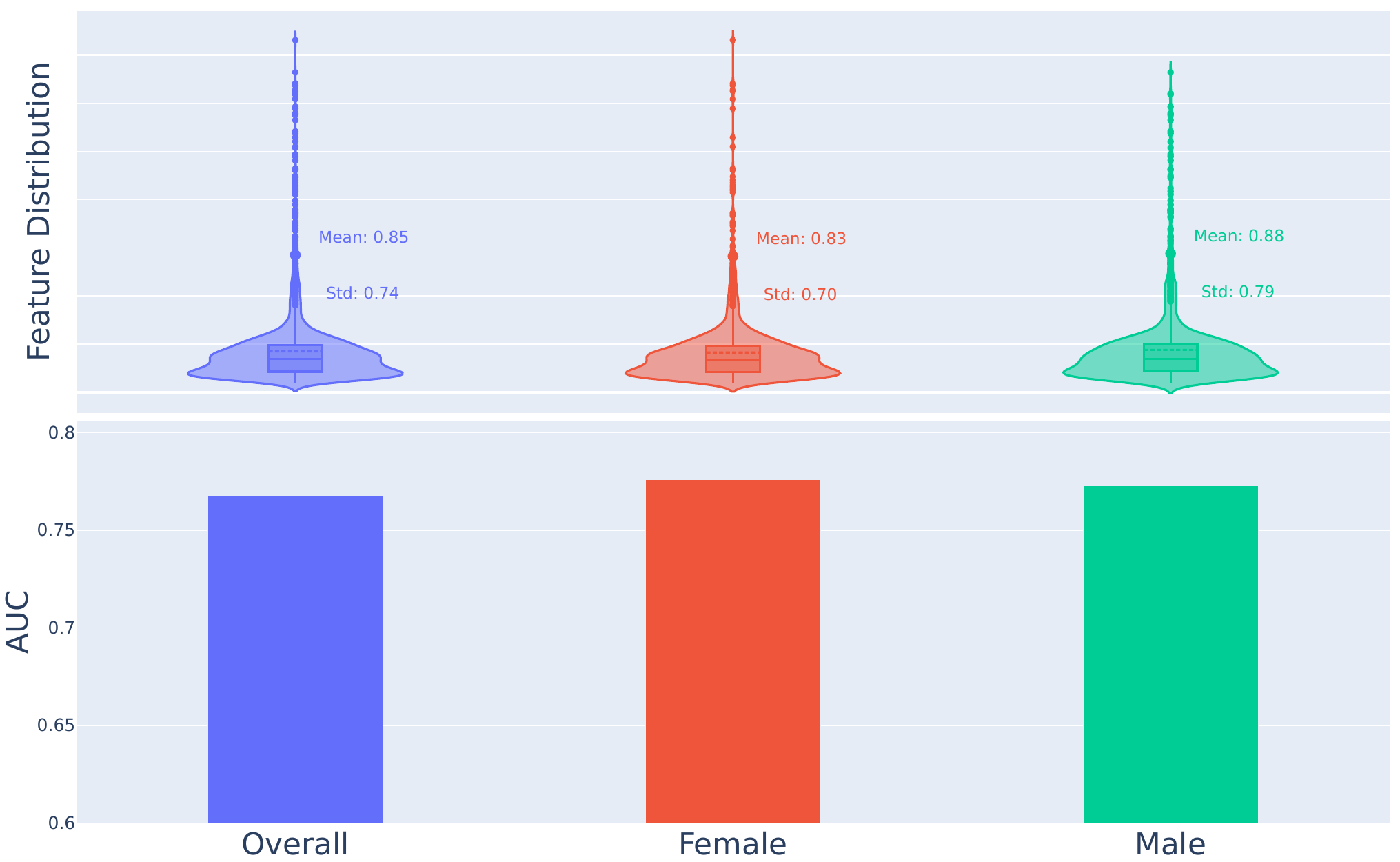}
        \caption{ViT-FAR for DR Det. on Gender}
        \label{fig:dr_vitfar_gender}
    \end{subfigure}\hfill
    \begin{subfigure}{0.33\textwidth}
        \centering
        \includegraphics[width=\linewidth]{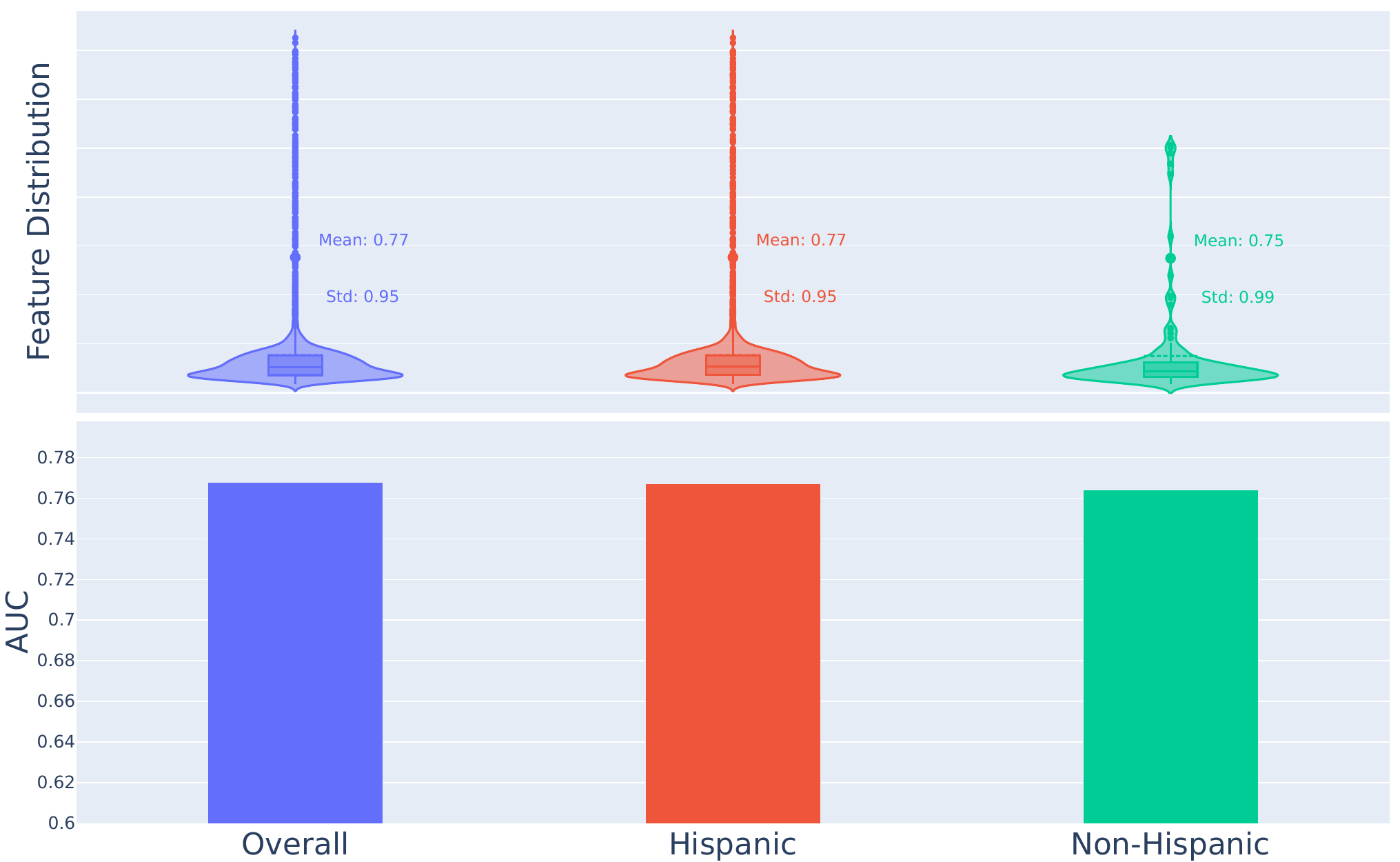}
        \caption{ViT-FAR for DR Det. on Ethnicity}
        \label{fig:dr_vitfar_ethnicity}
    \end{subfigure}

    \begin{subfigure}{0.33\textwidth}
        \centering
        \includegraphics[width=\linewidth]{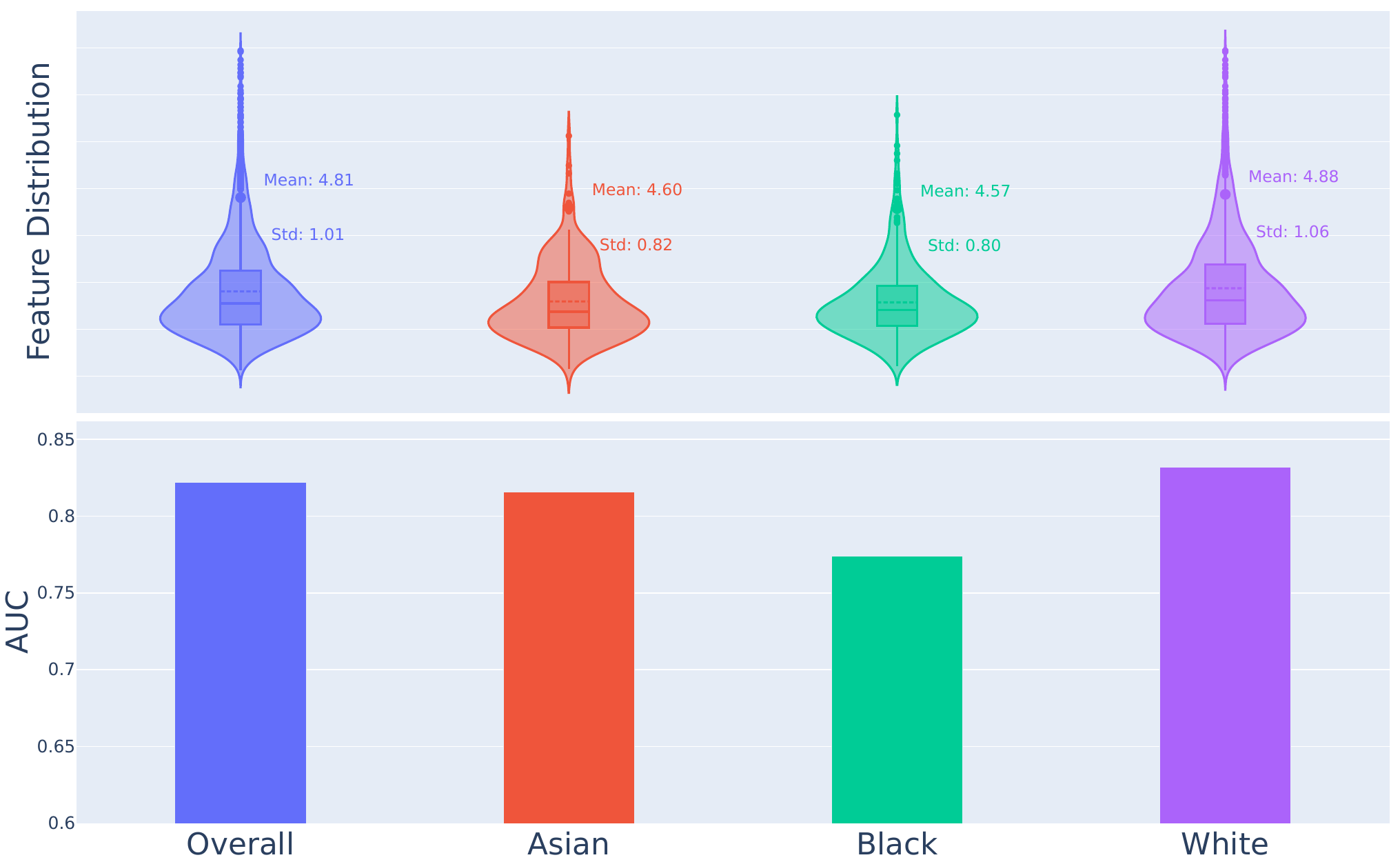}
        \caption{EffNet-FAR for DR Det. on Race}
        \label{fig:dr_effnetfar_race}
    \end{subfigure}\hfill
    \begin{subfigure}{0.33\textwidth}
        \centering
        \includegraphics[width=\linewidth]{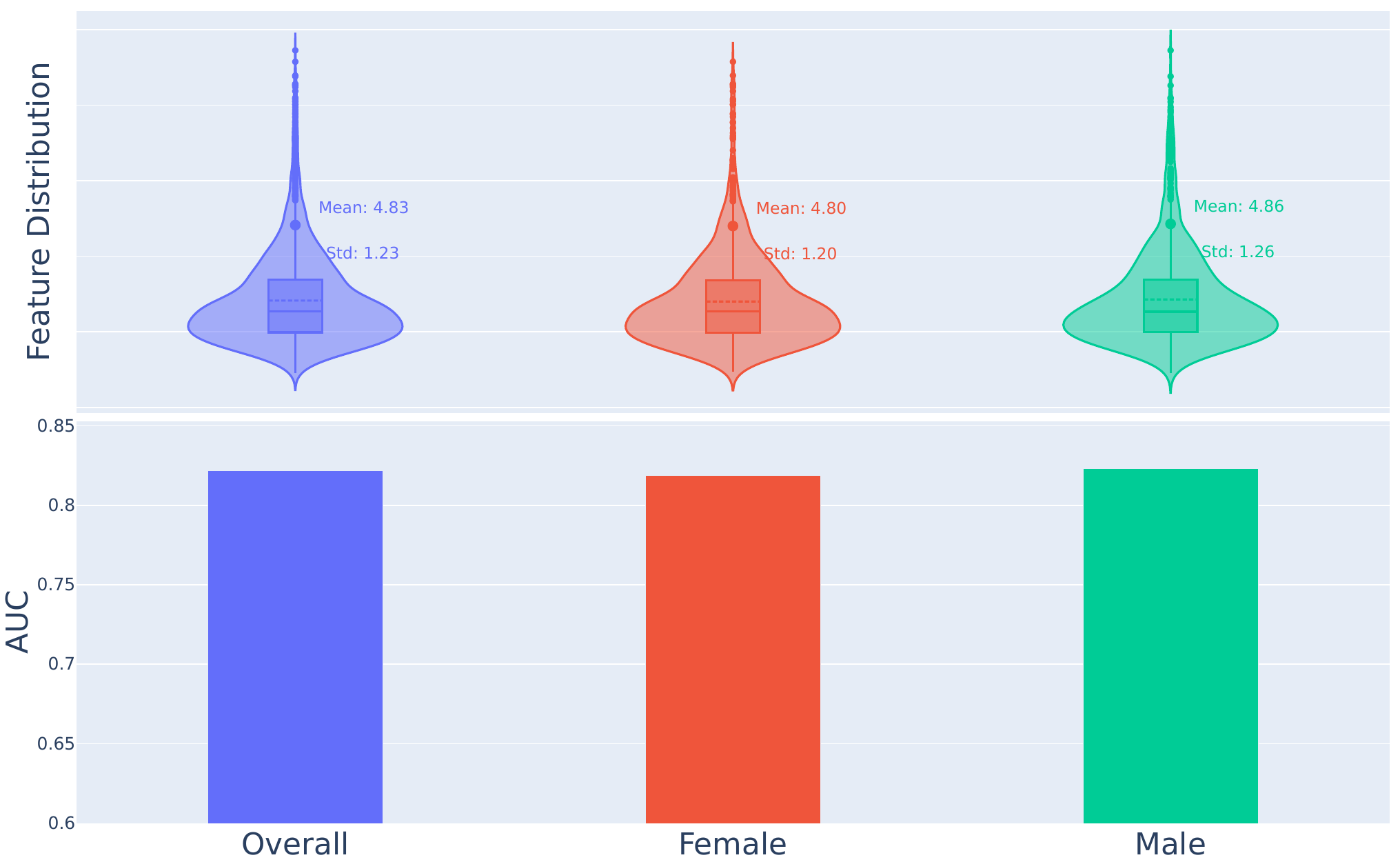}
        \caption{EffNet-FAR for DR Det. on Gender}
        \label{fig:dr_effnetfar_gender}
    \end{subfigure}\hfill
    \begin{subfigure}{0.33\textwidth}
        \centering
        \includegraphics[width=\linewidth]{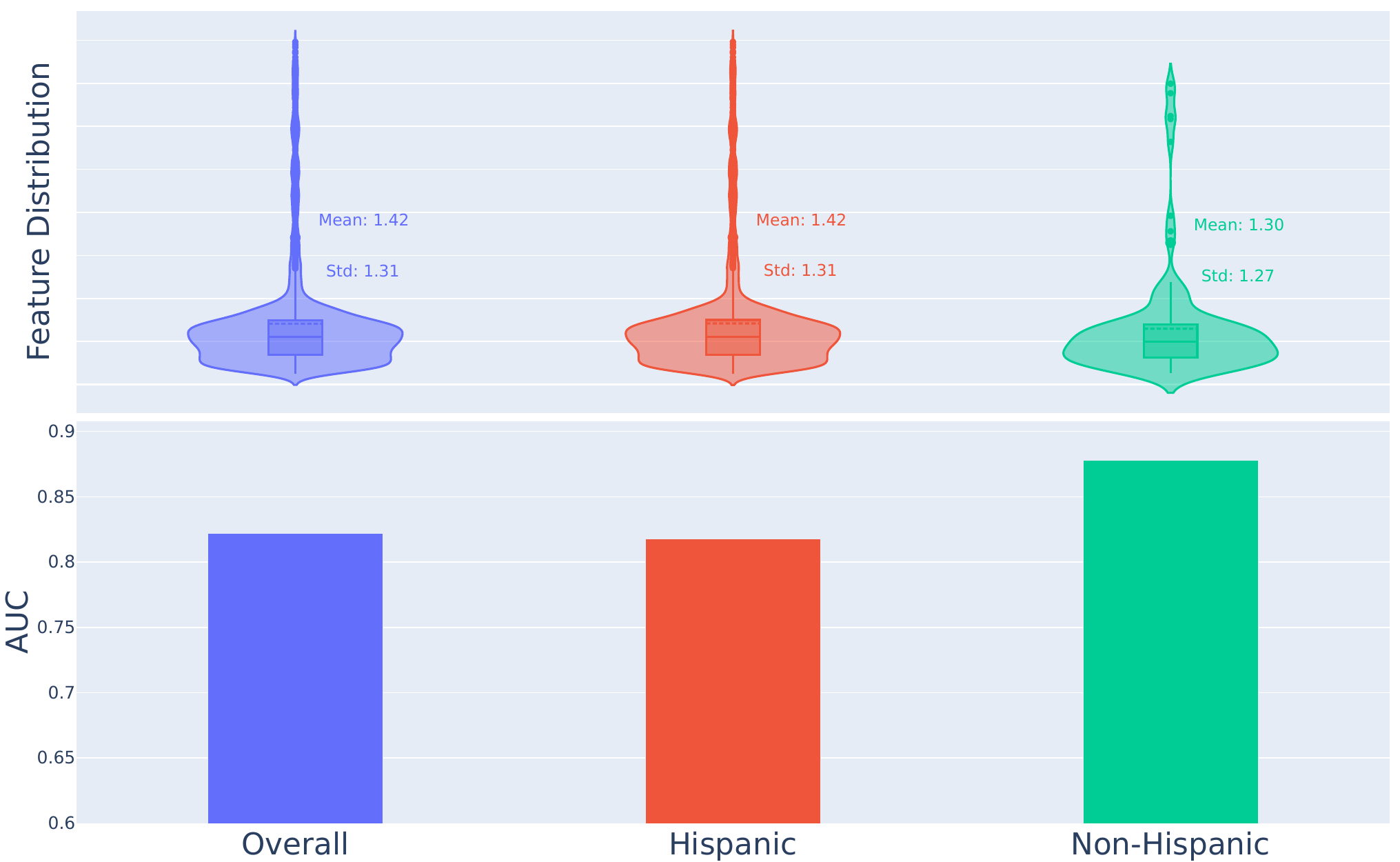}
        \caption{EffNet-FAR for DR Det. on Ethnicity}
        \label{fig:dr_effnetfar_ethnicity}
    \end{subfigure}

    \begin{subfigure}{0.48\textwidth}
        \centering
        \includegraphics[width=\linewidth]{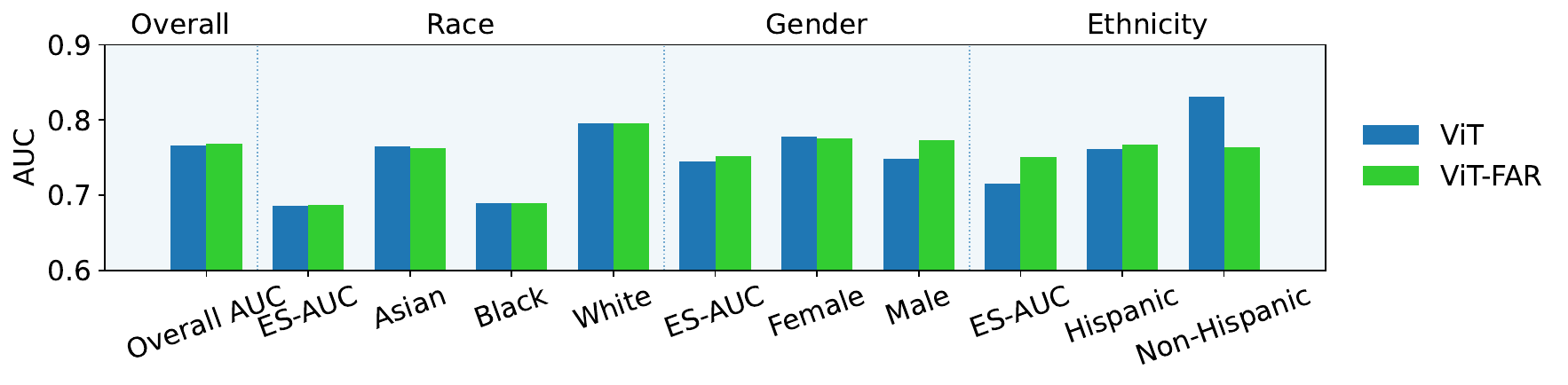}
        \caption{Comparison of ViT and ViT-FAR on DR Det.}
        \label{fig:dr_vit_comparison}
    \end{subfigure}\hfill
    \begin{subfigure}{0.48\textwidth}
        \centering
        \includegraphics[width=\linewidth]{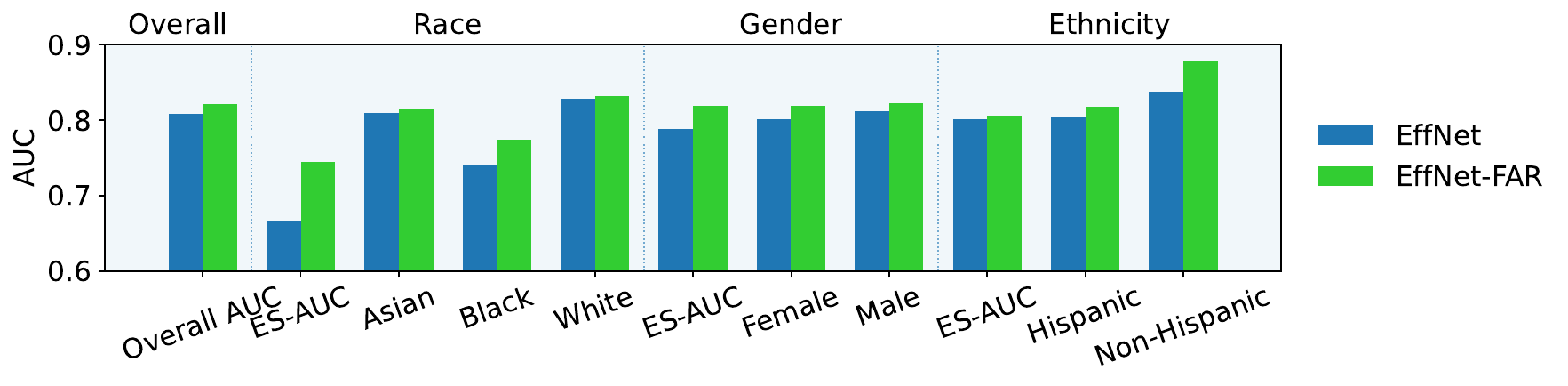}
        \caption{Comparison of EffNet and EffNet-FAR on DR Det.}
        \label{fig:dr_eff_comparison}
    \end{subfigure}
\end{minipage}
\end{adjustbox}

\caption{Feature distribution, AUC performance, and AUC comparison of ViT, EfficientNet, and their fairness-aware regularization (FAR) variants (ViT-FAR and EffNet-FAR) for DR detection across three demographic attributes, including Race, Gender, and Ethnicity, on \textbf{FairVision}.}
\label{fig:dr}
\end{figure*}

\begin{figure*}[!h]
\centering
\begin{adjustbox}{scale=0.7}
\begin{minipage}{\linewidth}
\centering
    \begin{subfigure}{0.33\textwidth}
        \centering
        \includegraphics[width=\linewidth]{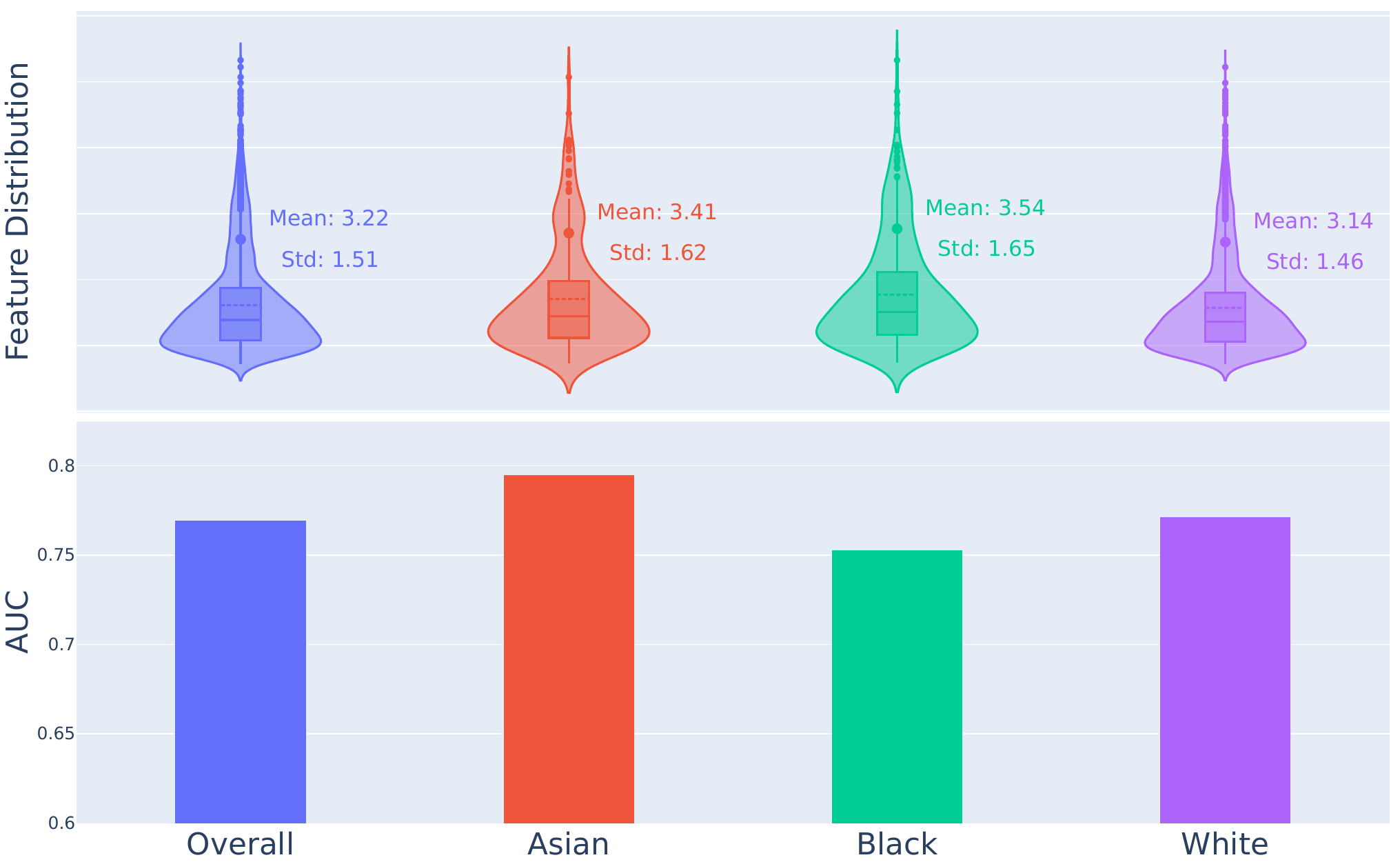}
        \caption{ViT for GL Det. on Race}
        \label{fig:glaucoma_vit_race}
    \end{subfigure}\hfill
    \begin{subfigure}{0.33\textwidth}
        \centering
        \includegraphics[width=\linewidth]{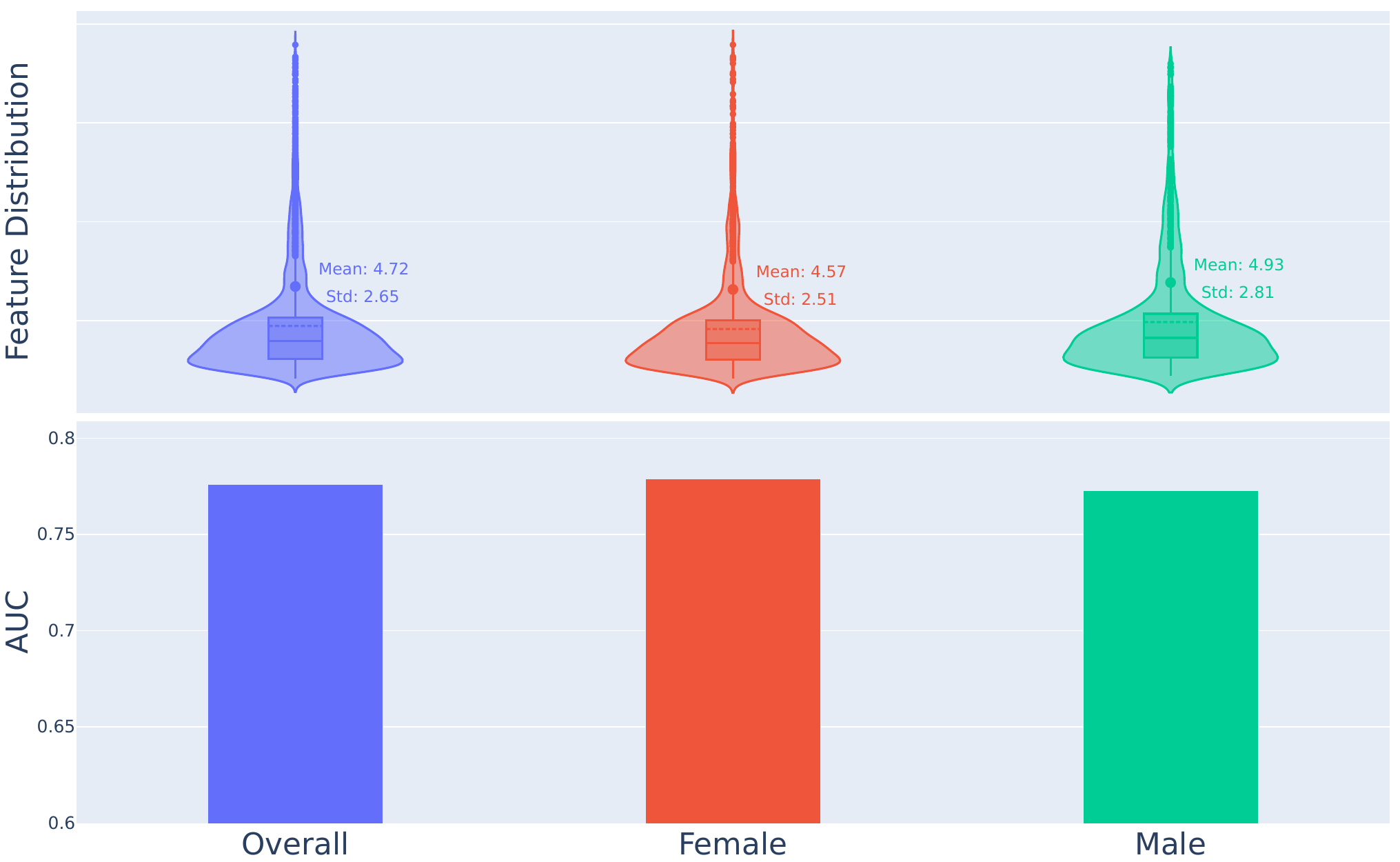}
        \caption{ViT for GL Det. on Gender}
        \label{fig:glaucoma_vit_gender}
    \end{subfigure}\hfill
    \begin{subfigure}{0.33\textwidth}
        \centering
        \includegraphics[width=\linewidth]{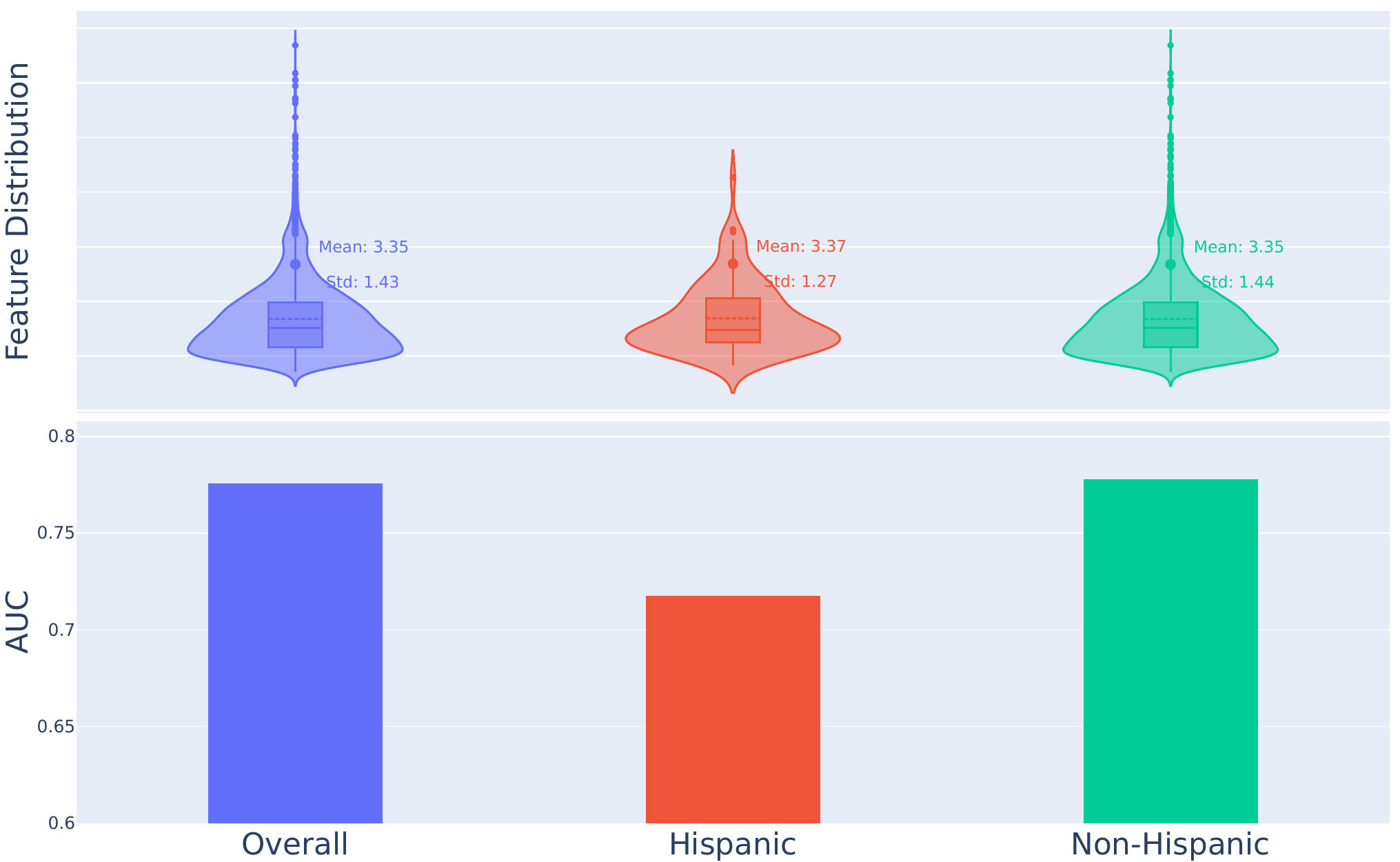}
        \caption{ViT for GL Det. on Ethnicity}
        \label{fig:glaucoma_vit_ethnicity}
    \end{subfigure}

    \begin{subfigure}{0.33\textwidth}
        \centering
        \includegraphics[width=\linewidth]{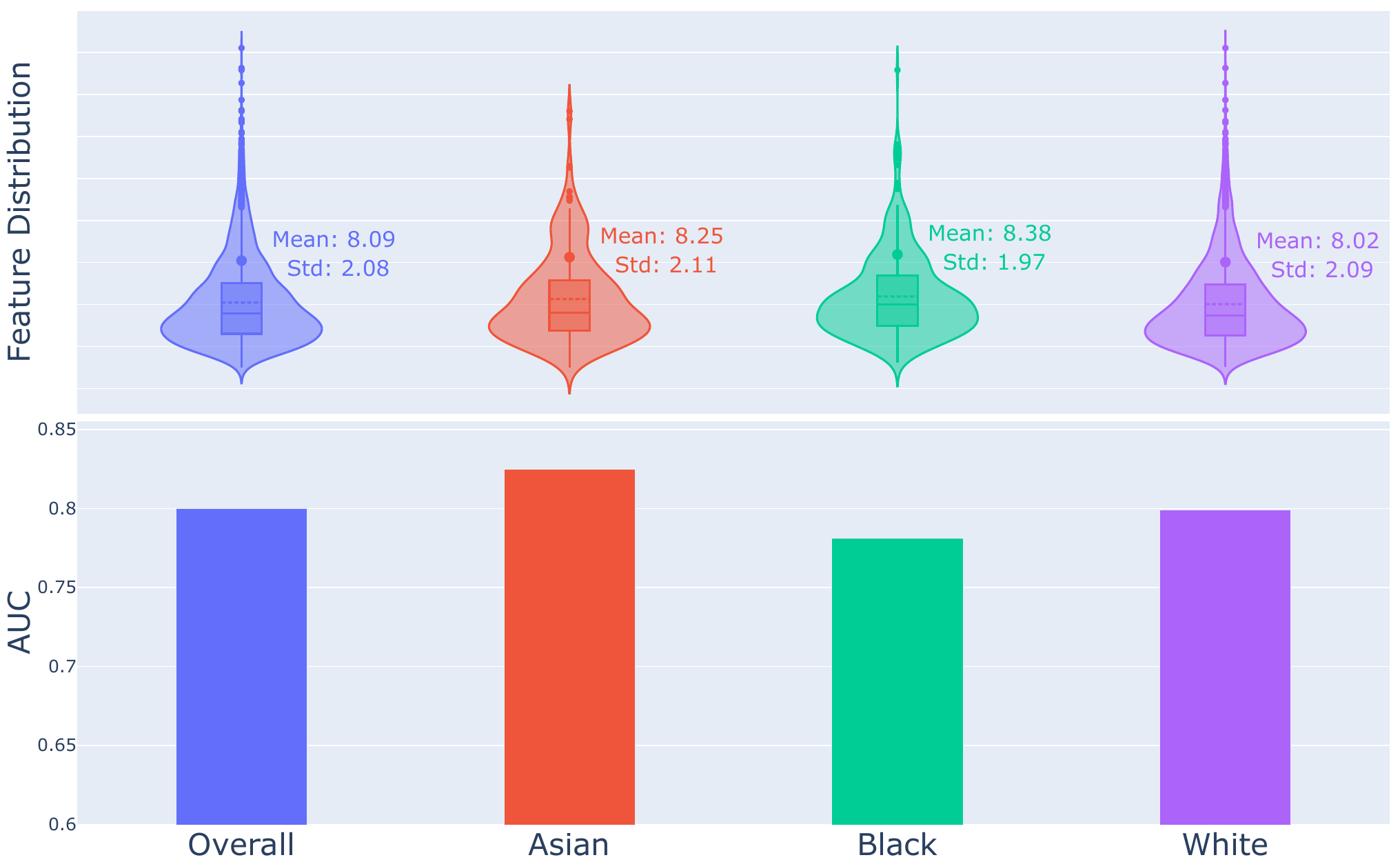}
        \caption{EffNet for GL Det. on Race}
        \label{fig:glaucoma_eff_race}
    \end{subfigure}\hfill
    \begin{subfigure}{0.33\textwidth}
        \centering
        \includegraphics[width=\linewidth]{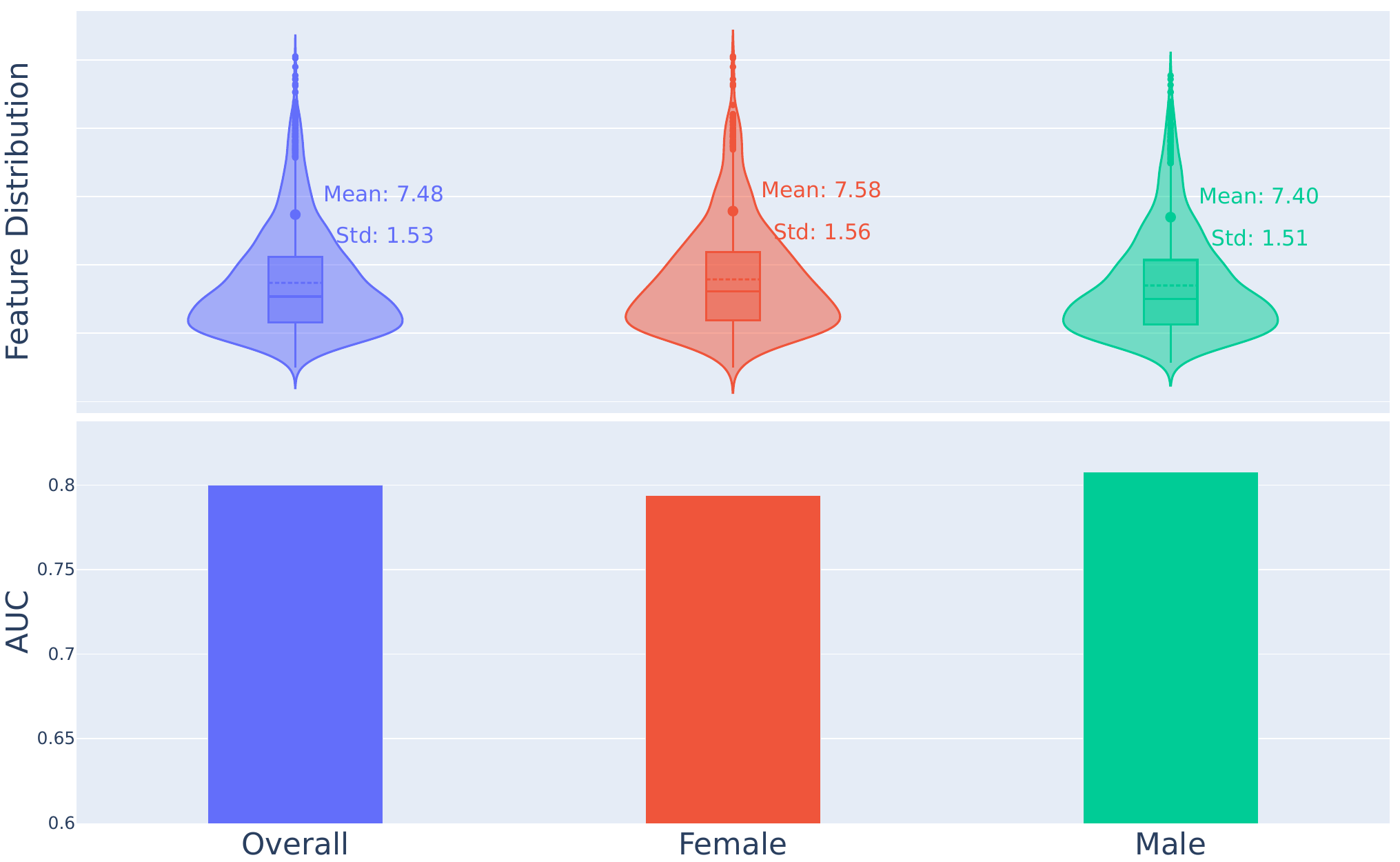}
        \caption{EffNet for GL Det. on Gender}
        \label{fig:glaucoma_eff_gender}
    \end{subfigure}\hfill
    \begin{subfigure}{0.33\textwidth}
        \centering
        \includegraphics[width=\linewidth]{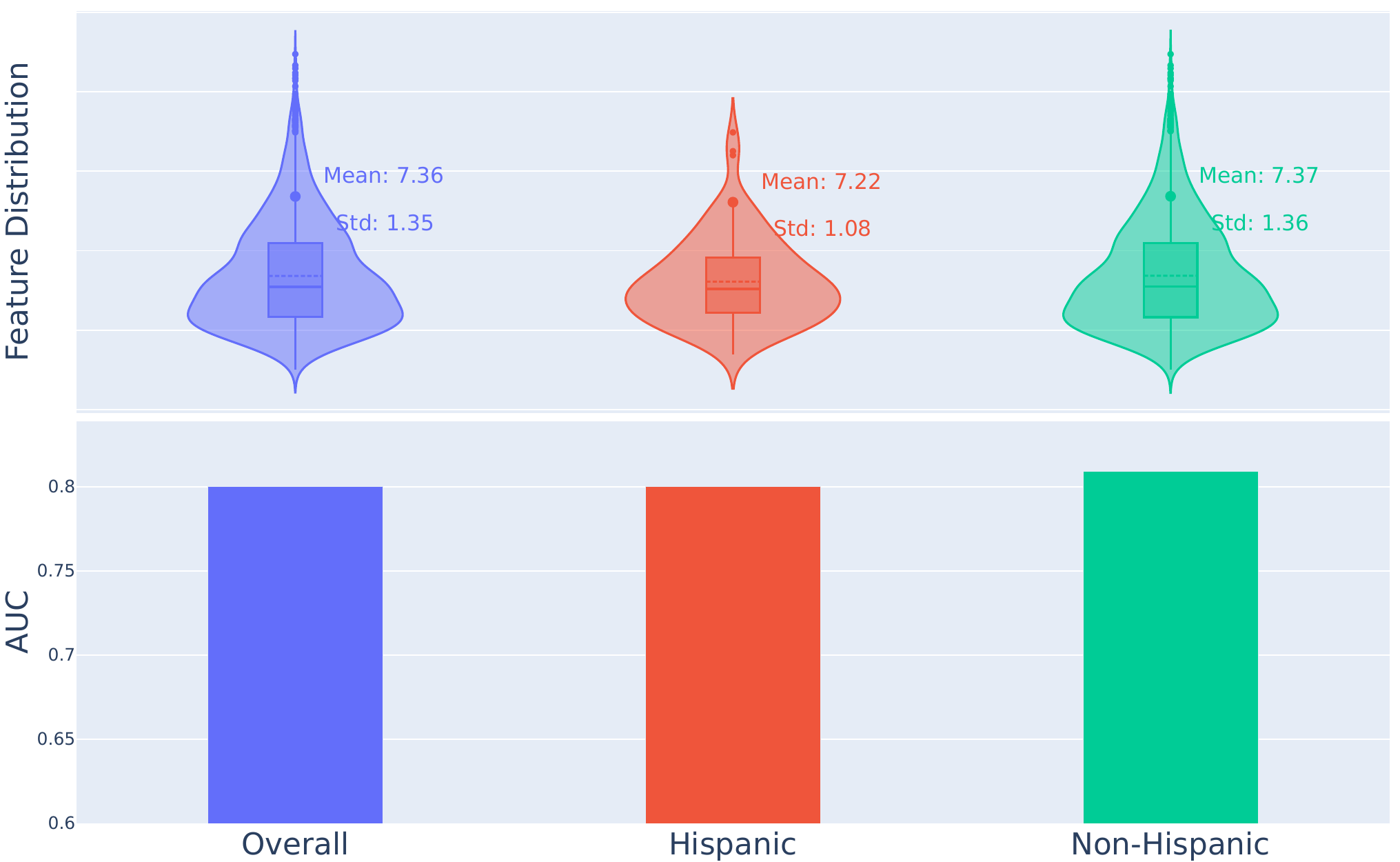}
        \caption{EffNet for GL Det. on Ethnicity}
        \label{fig:glaucoma_eff_ethnicity}
    \end{subfigure}

    \begin{subfigure}{0.33\textwidth}
        \centering
        \includegraphics[width=\linewidth]{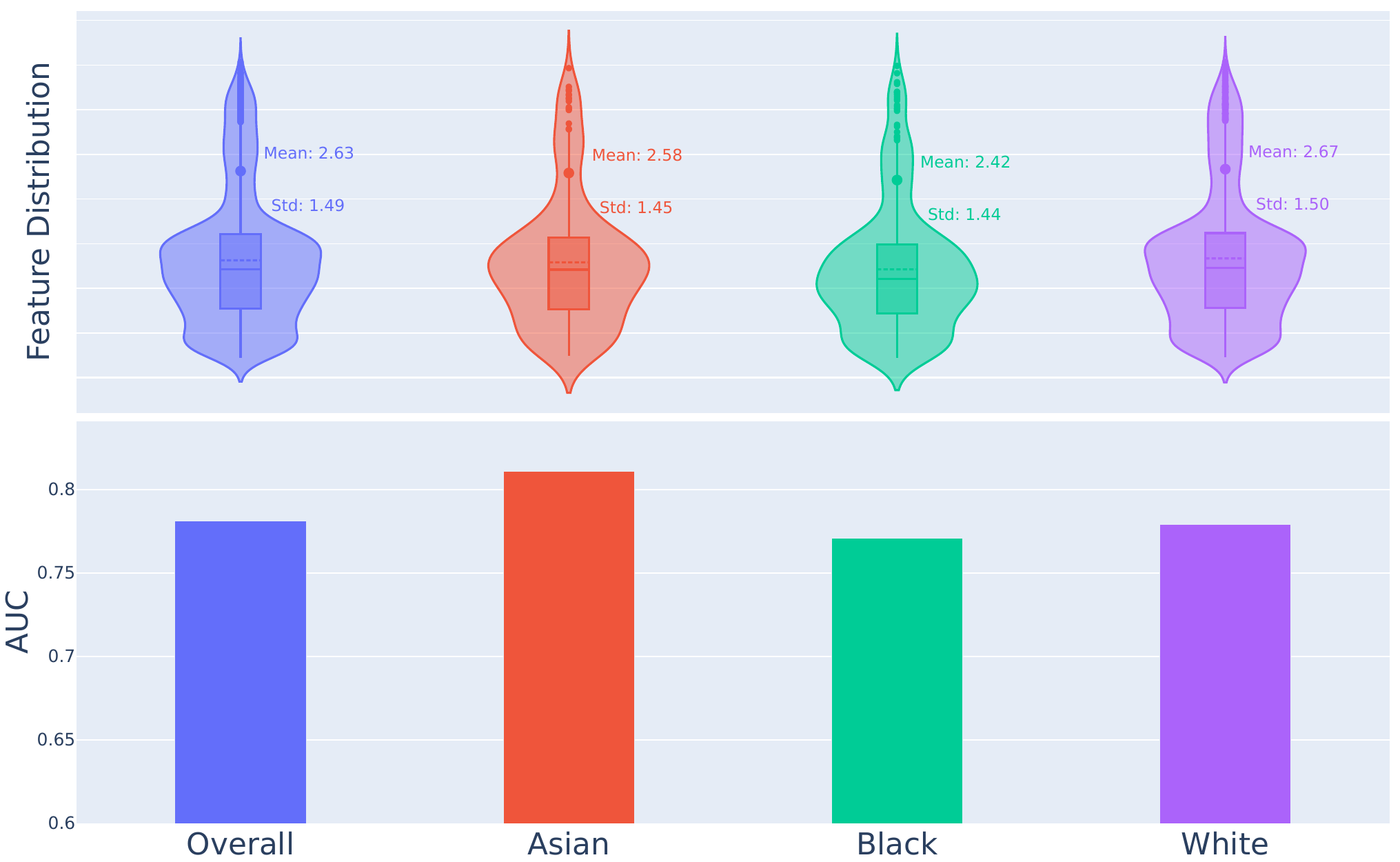}
        \caption{ViT-FAR for GL Det. on Race}
        \label{fig:glaucoma_vitfar_race}
    \end{subfigure}\hfill
    \begin{subfigure}{0.33\textwidth}
        \centering
        \includegraphics[width=\linewidth]{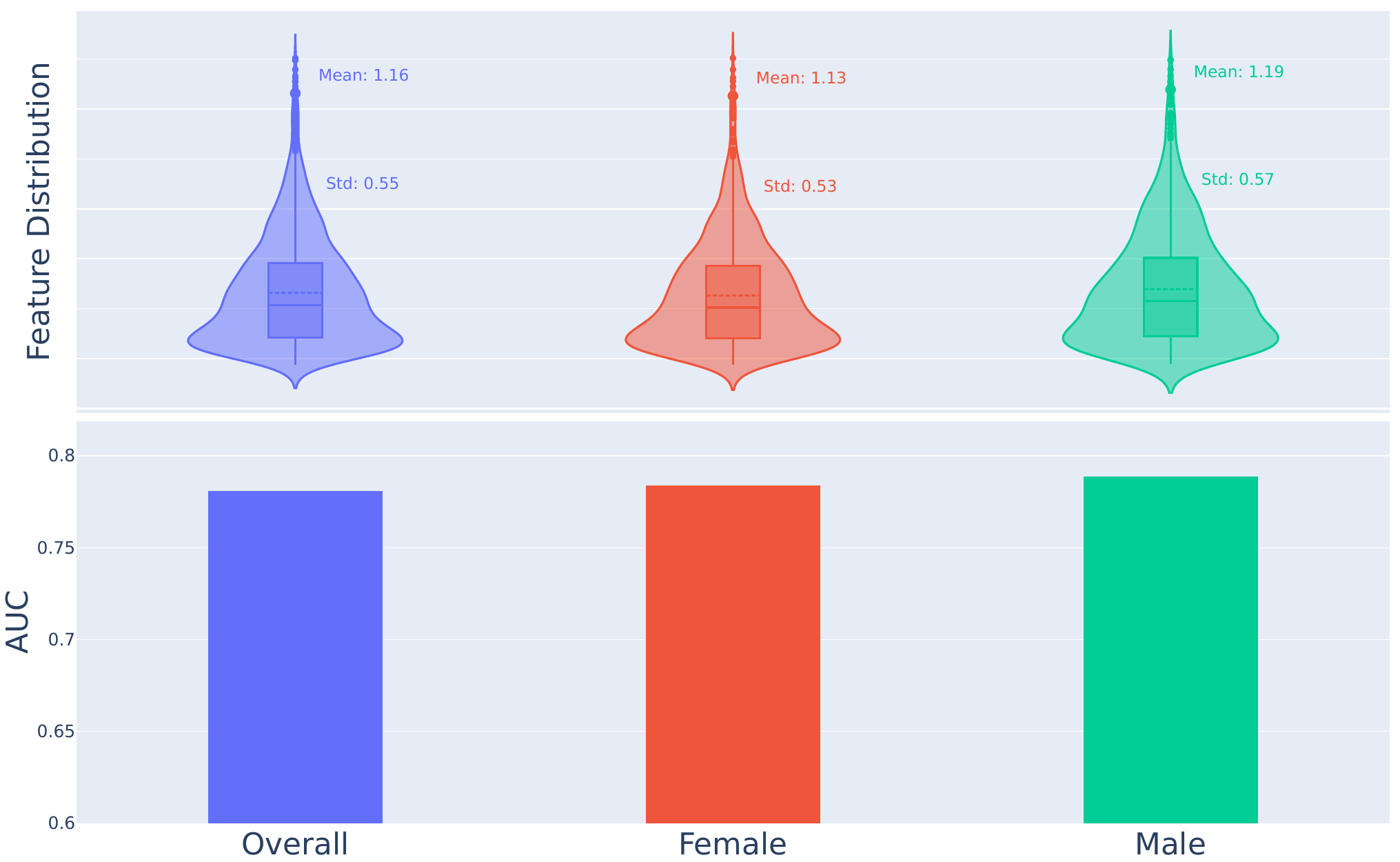}
        \caption{ViT-FAR for GL Det. on Gender}
        \label{fig:glaucoma_vitfar_gender}
    \end{subfigure}\hfill
    \begin{subfigure}{0.33\textwidth}
        \centering
        \includegraphics[width=\linewidth]{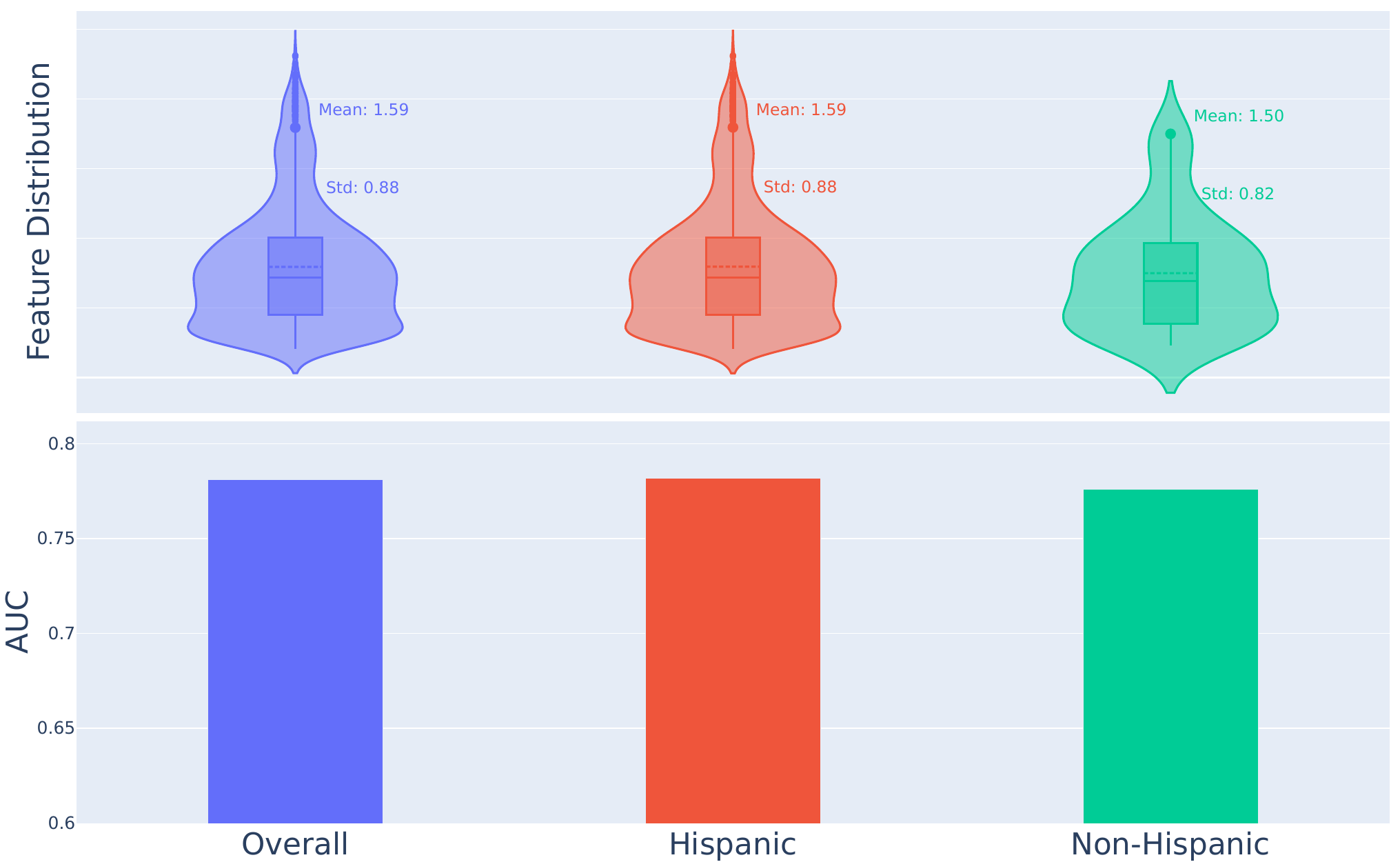}
        \caption{ViT-FAR for GL Det. on Ethnicity}
        \label{fig:glaucoma_vitfar_ethnicity}
    \end{subfigure}

    \begin{subfigure}{0.33\textwidth}
        \centering
        \includegraphics[width=\linewidth]{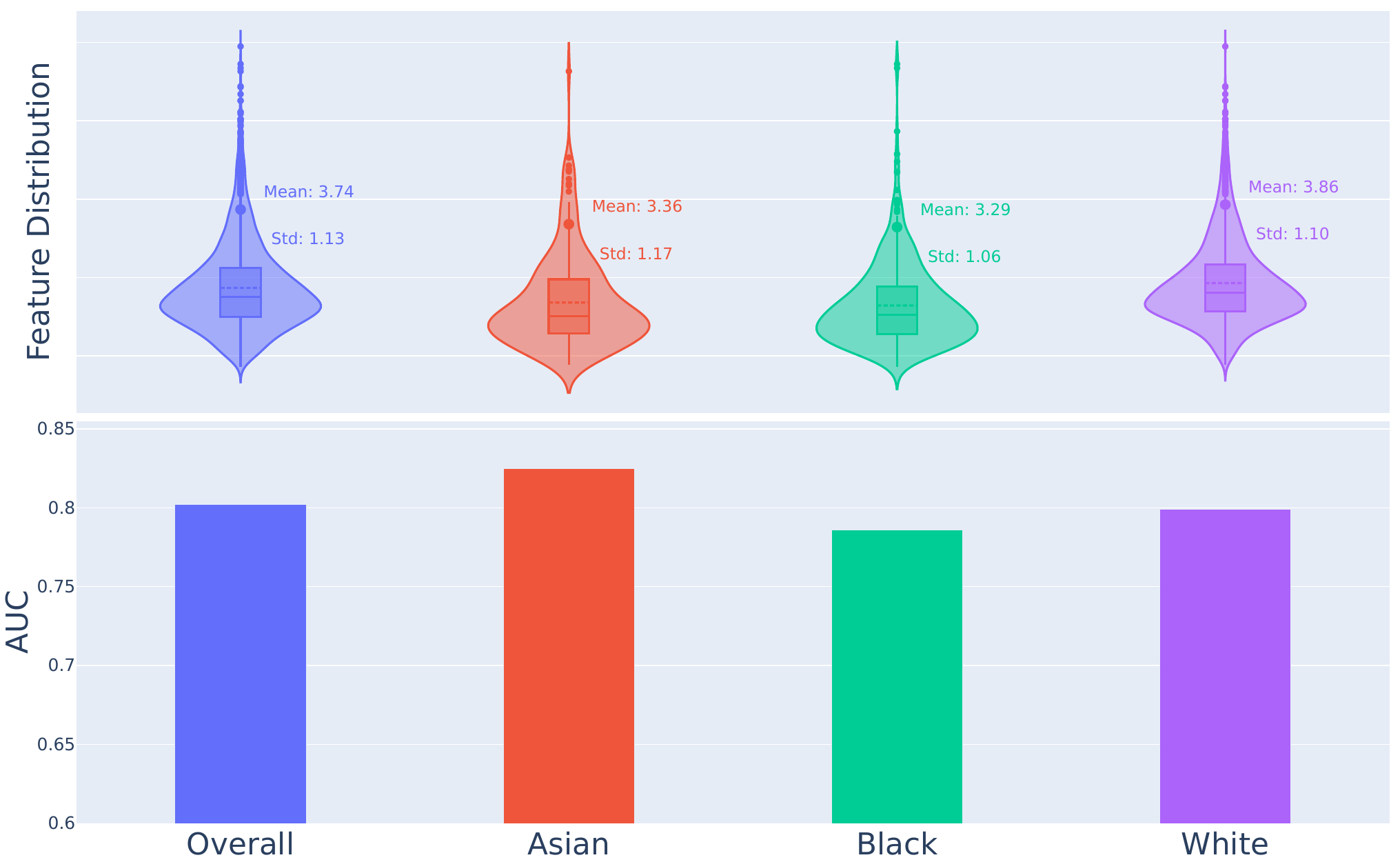}
        \caption{EffNet-FAR for GL Det. on Race}
        \label{fig:glaucoma_effnetfar_race}
    \end{subfigure}\hfill
    \begin{subfigure}{0.33\textwidth}
        \centering
        \includegraphics[width=\linewidth]{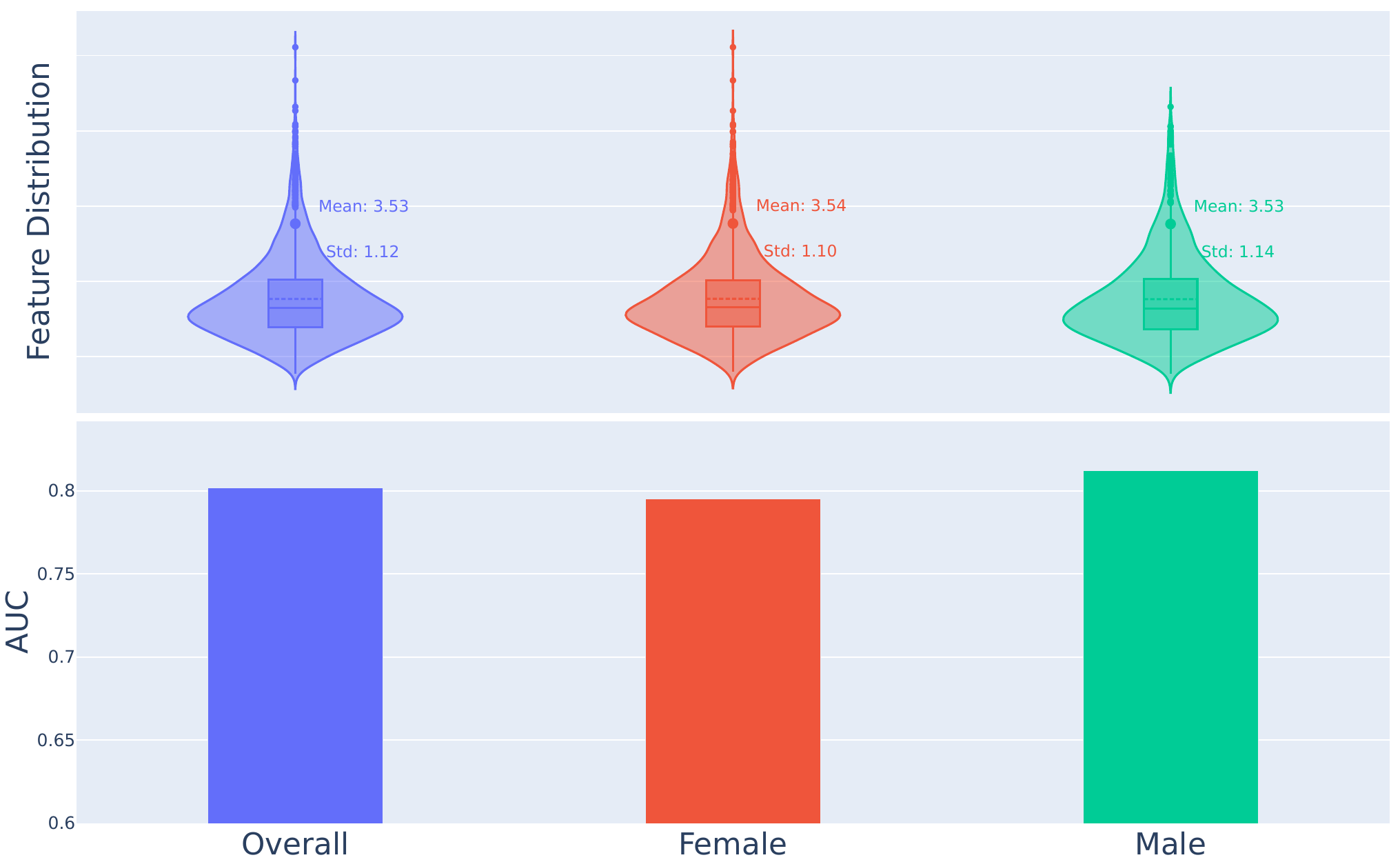}
        \caption{EffNet-FAR for GL Det. on Gender}
        \label{fig:glaucoma_effnetfar_gender}
    \end{subfigure}\hfill
    \begin{subfigure}{0.33\textwidth}
        \centering
        \includegraphics[width=\linewidth]{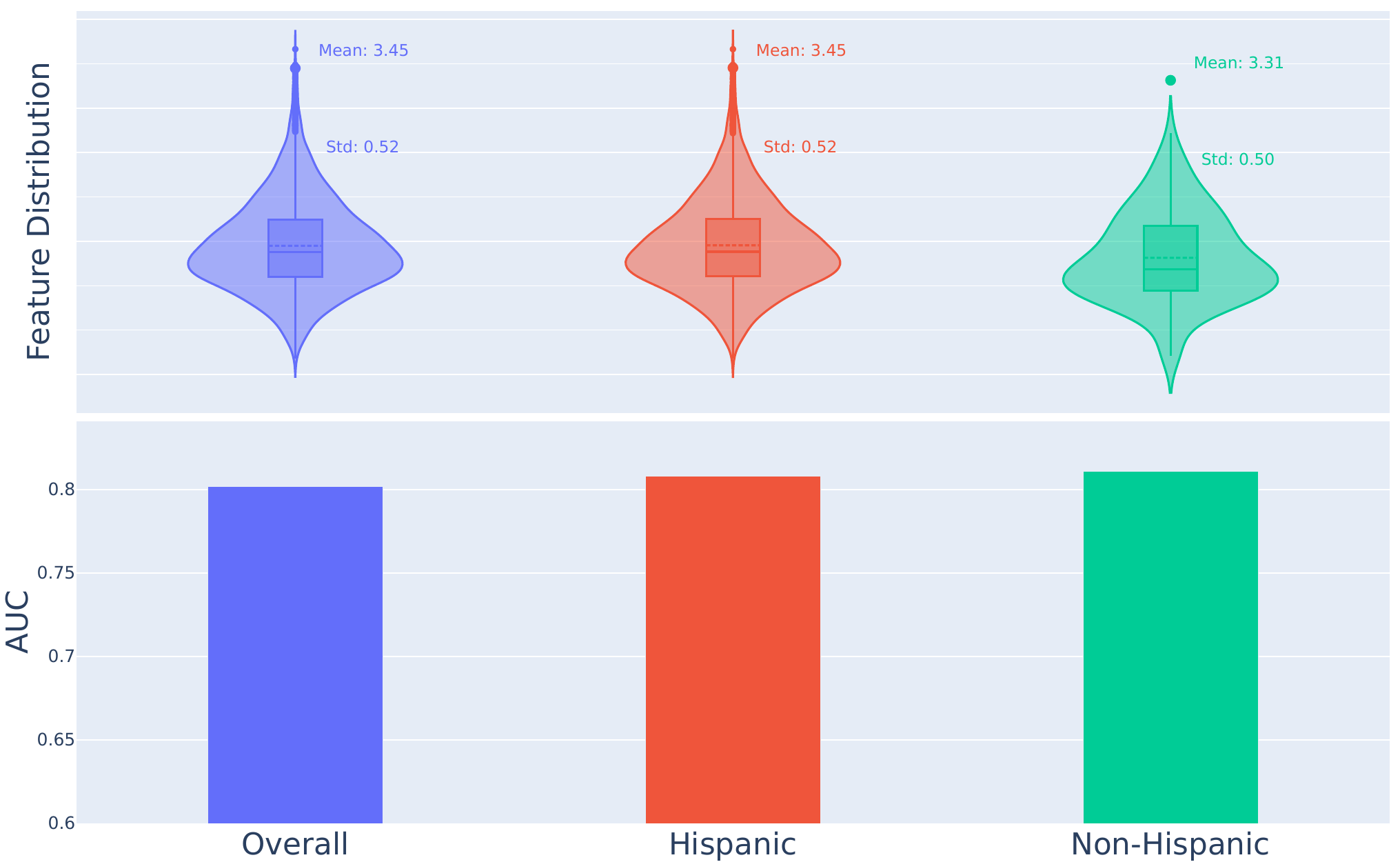}
        \caption{EffNet-FAR for GL Det. on Ethnicity}
        \label{fig:glaucoma_effnetfar_ethnicity}
    \end{subfigure}

    \begin{subfigure}{0.48\textwidth}
        \centering
        \includegraphics[width=\linewidth]{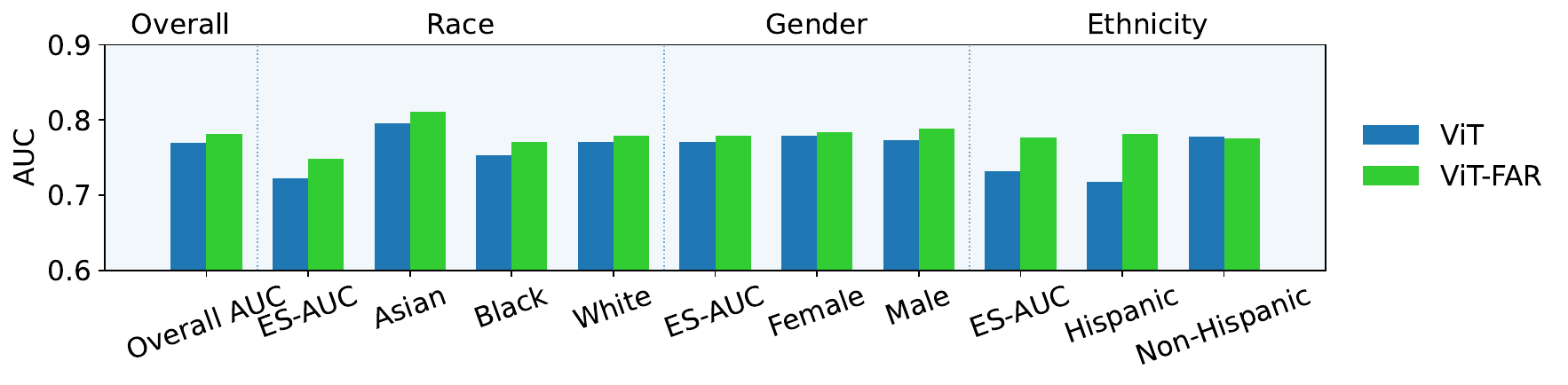}
        \caption{Comparison of ViT and ViT-FAR on GL Det.}
        \label{fig:gl_vit_comparison}
    \end{subfigure}\hfill
    \begin{subfigure}{0.48\textwidth}
        \centering
        \includegraphics[width=\linewidth]{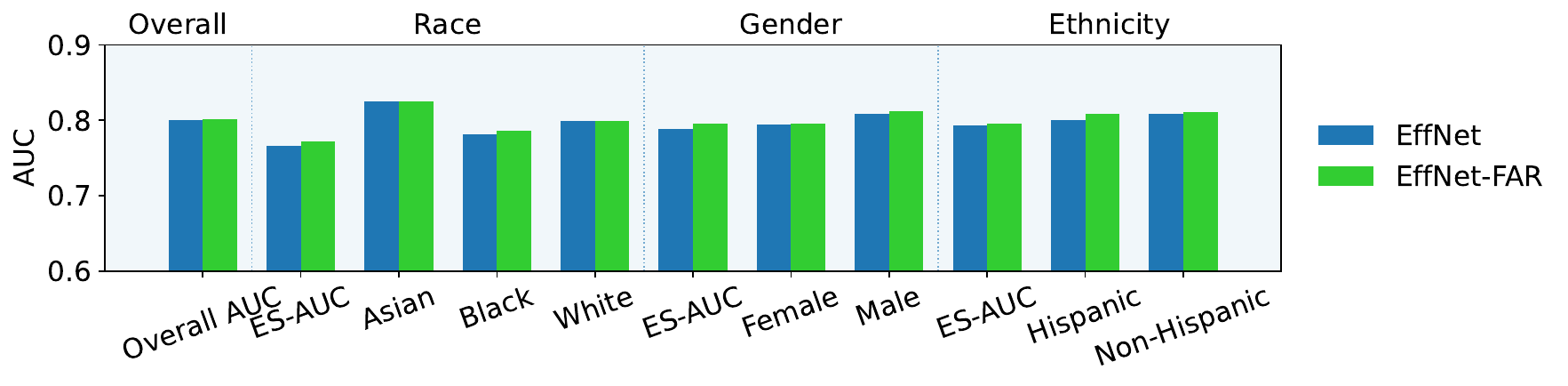}
        \caption{Comparison of EffNet and EffNet-FAR on GL Det.}
        \label{fig:gl_eff_comparison}
    \end{subfigure}
\end{minipage}
\end{adjustbox}

\caption{Feature distribution, AUC performance, and AUC comparison of ViT, EfficientNet, and their fairness-aware regularization (FAR) variants (ViT-FAR and EffNet-FAR) for Glaucoma (GL) detection across three demographic attributes, including Race, Gender, and Ethnicity, on \textbf{FairVision}.}
\label{fig:gl}
\end{figure*}

\noindent\textbf{Implementation Details}. 
For image datasets, we adopt two baseline models with distinct architectures: the CNN-based EfficientNet~\cite{Tan_ICML_2019} and the Transformer-based ViT~\cite{dosovitskiy2020image}. For tabular datasets, we employ TabTransformer~\cite{huang2020tabtransformer}, and for natural language datasets, we use RoBERTa~\cite{liu2019roberta}. All experiments are conducted on an NVIDIA A100 GPU with 80 GB of memory. EfficientNet and ViT are initialized with pre-trained weights from TorchVision and fine-tuned for 10 epochs with a learning rate of 1e-4; the batch size is set to 10 for EfficientNet and 50 for ViT. TabTransformer is trained for 50 epochs with a learning rate of 1e-4 and a batch size of 1,000. RoBERTa is fine-tuned for 50 epochs with a learning rate of 1e-5 and a batch size of 64.


\noindent\textbf{Results}. 
We first trained ViT and EfficientNet models on the training sets of the FairVision dataset for detecting DR, AMD, and Glaucoma. After training, the models were applied to the corresponding test sets, where we obtained the encoded features and predicted labels. Using these outputs, we calculated the Euclidean distance of the features to the mean, representing the empirical feature distribution, and evaluated the Area Under the Curve (AUC) for both overall and subgroup performance. The feature distributions and AUC results for DR, AMD, and Glaucoma detection are illustrated in Fig.~\ref{fig:amd}, Fig.~\ref{fig:dr}, and Fig.~\ref{fig:gl}. From the figures, it is evident that both models achieved reasonable detection performance across all three diseases, with EfficientNet consistently outperforming ViT. Additionally, we observed significant disparities in AUC values across demographic groups, underscoring the importance of studying fairness in AI systems. These disparities are particularly pronounced in the results stratified by race, where the Black subgroup exhibits markedly lower detection accuracy compared to the Asian and White subgroups. Notably, incorporating fairness-aware regularization (FAR) improves ES-AUC across demographic groups, demonstrating enhanced balance between subpopulations.

\begin{figure*}[h]
\centering
\begin{adjustbox}{scale=0.7}
\begin{minipage}{\linewidth}
\centering
    \begin{subfigure}{0.5\textwidth}
        \centering
        \includegraphics[width=\linewidth]{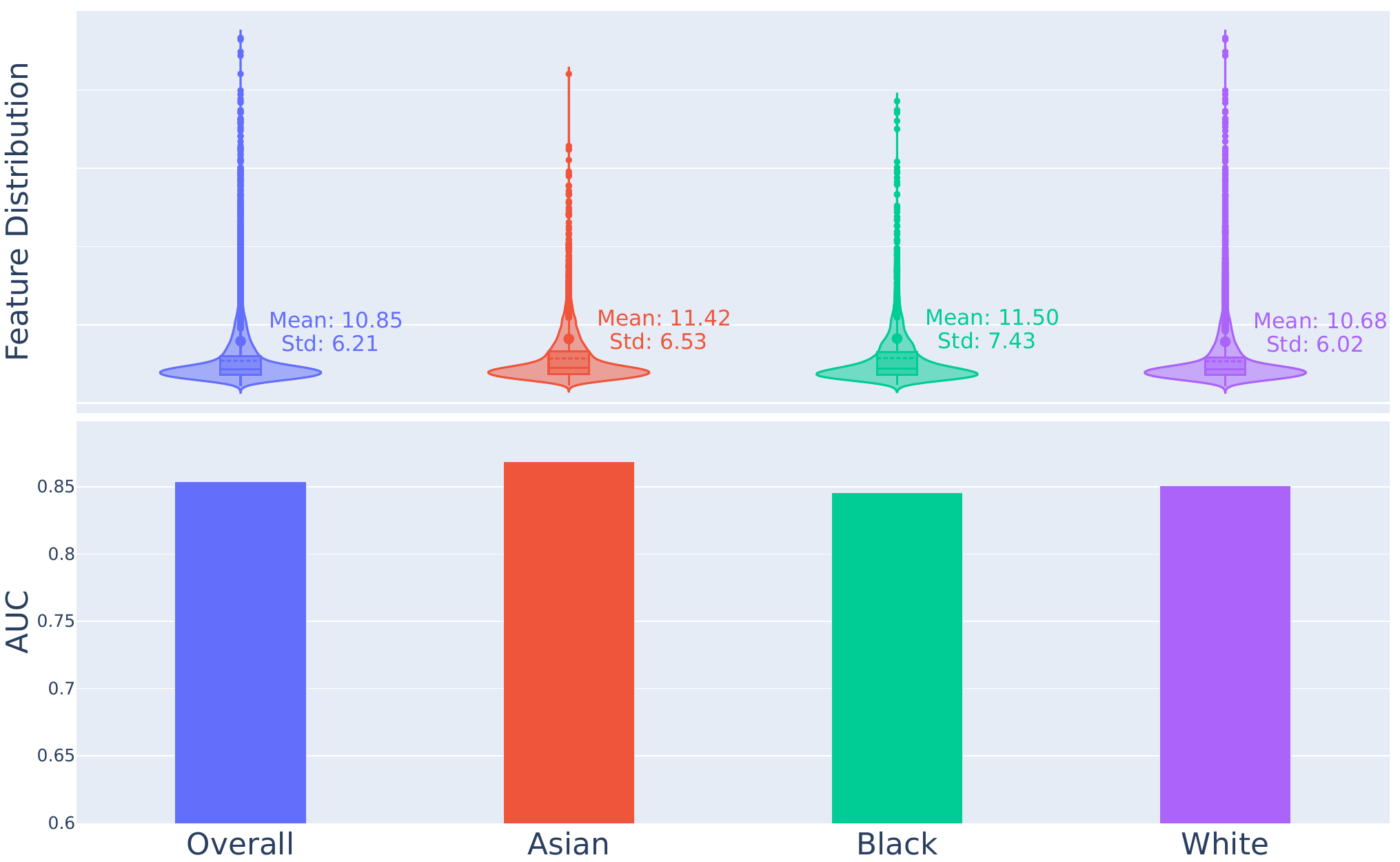}
        \caption{EffNet for PE Det. on Race}
        \label{fig:chexpert_eff_race}
    \end{subfigure}%
    \begin{subfigure}{0.5\textwidth}
        \centering
        \includegraphics[width=\linewidth]{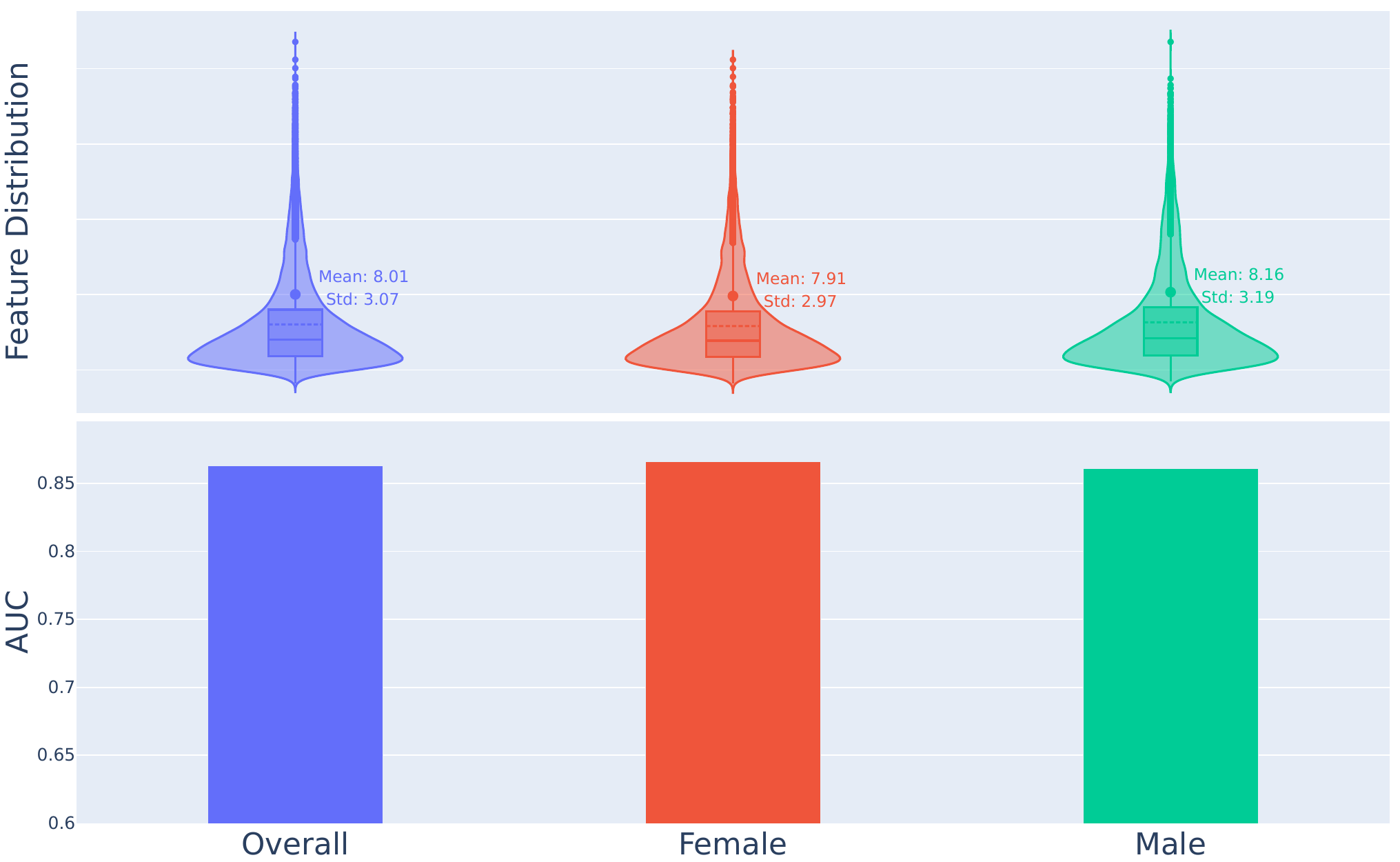}
        \caption{EffNet for PE Det. on Gender}
        \label{fig:chexpert_eff_gender}
    \end{subfigure}

    \begin{subfigure}{0.5\textwidth}
        \centering
        \includegraphics[width=\linewidth]{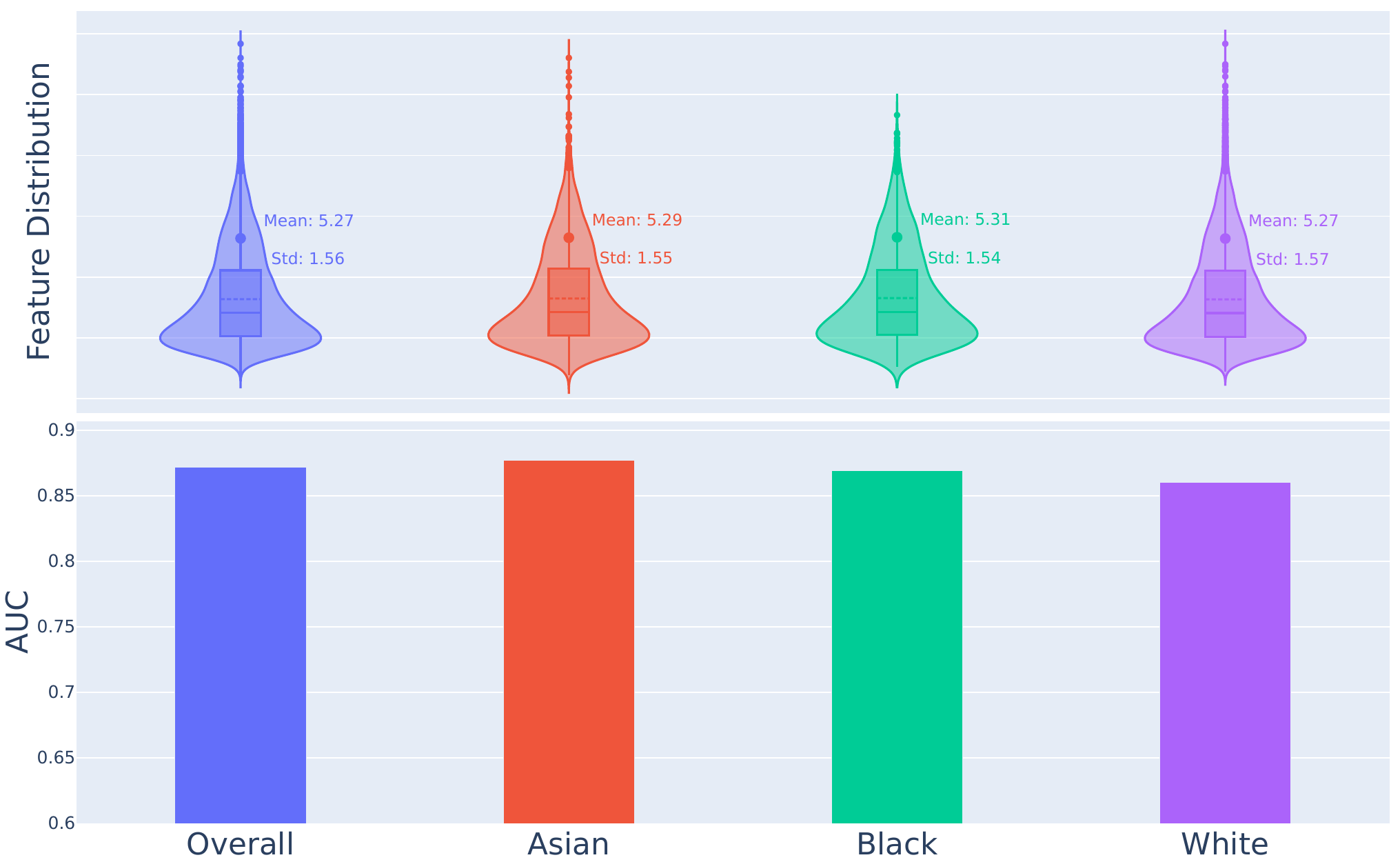}
        \caption{EffNet-FAR for PE Det. on Race}
        \label{fig:chexpert_efffar_race}
    \end{subfigure}%
    \begin{subfigure}{0.5\textwidth}
        \centering
        \includegraphics[width=\linewidth]{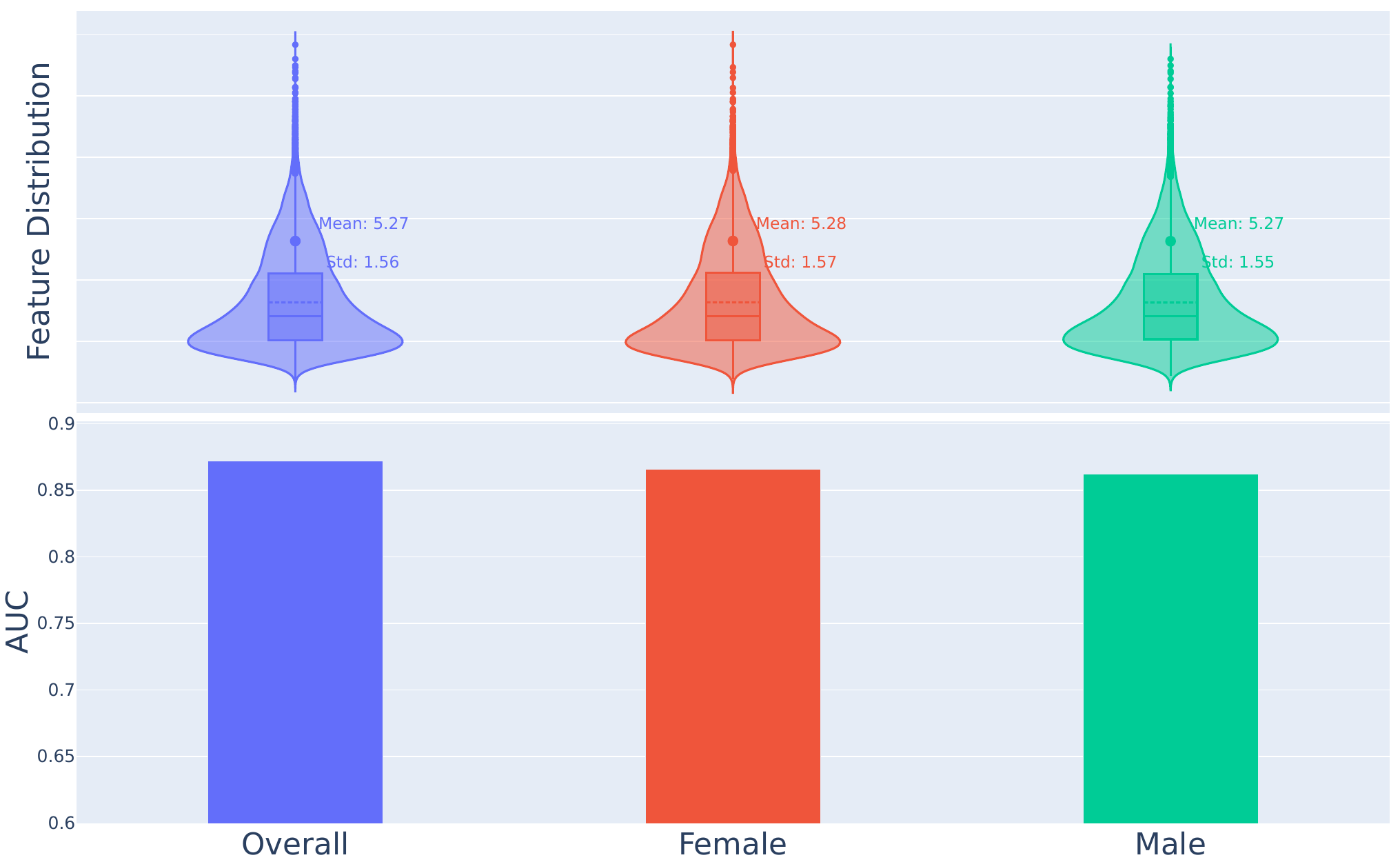}
        \caption{EffNet-FAR for PE Det. on Gender}
        \label{fig:chexpert_efffar_gender}
    \end{subfigure}

    \begin{subfigure}{0.8\textwidth}
        \centering
        \includegraphics[width=\linewidth]{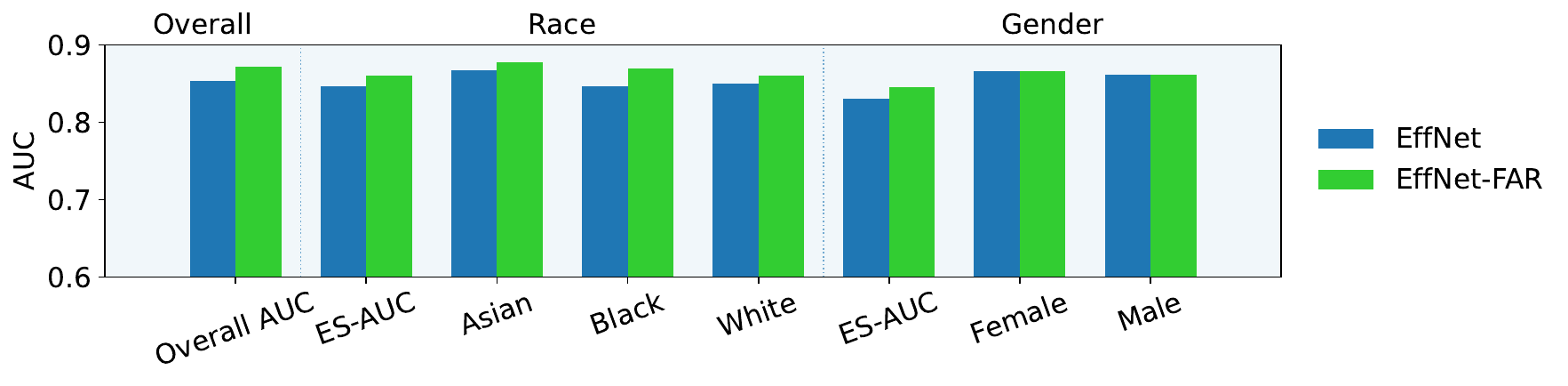}
        \caption{Comparison of EffNet and EffNet-FAR on PE Det.}
        \label{fig:chexpert_eff_comparison}
    \end{subfigure}
\end{minipage}
\end{adjustbox}

\caption{Feature distribution, AUC performance, and AUC comparison of EfficientNet and EffNet-FAR for Pleural Effusion (PE) detection across two demographic attributes, including Race and Gender, on \textbf{Chexpert}.}
\label{fig:chexpert}
\end{figure*}

\begin{figure*}[h]
\centering
\begin{adjustbox}{scale=0.7}
\begin{minipage}{\linewidth}
\centering
    \begin{subfigure}{0.5\textwidth}
        \centering
        \includegraphics[width=\linewidth]{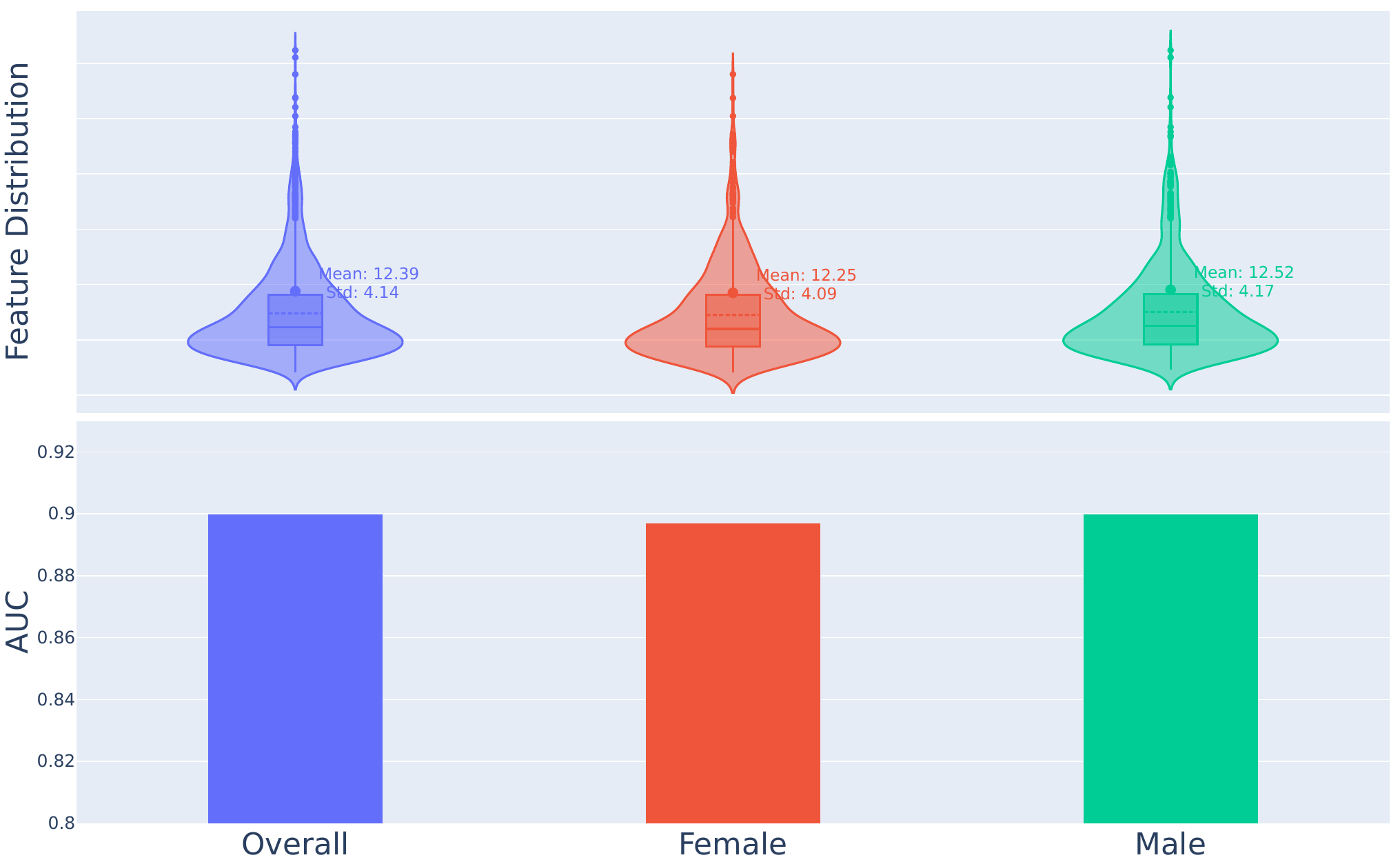}
        \caption{EffNet for SC Det. on Gender}
        \label{fig:ham_eff_gender}
    \end{subfigure}%
    \begin{subfigure}{0.5\textwidth}
        \centering
        \includegraphics[width=\linewidth]{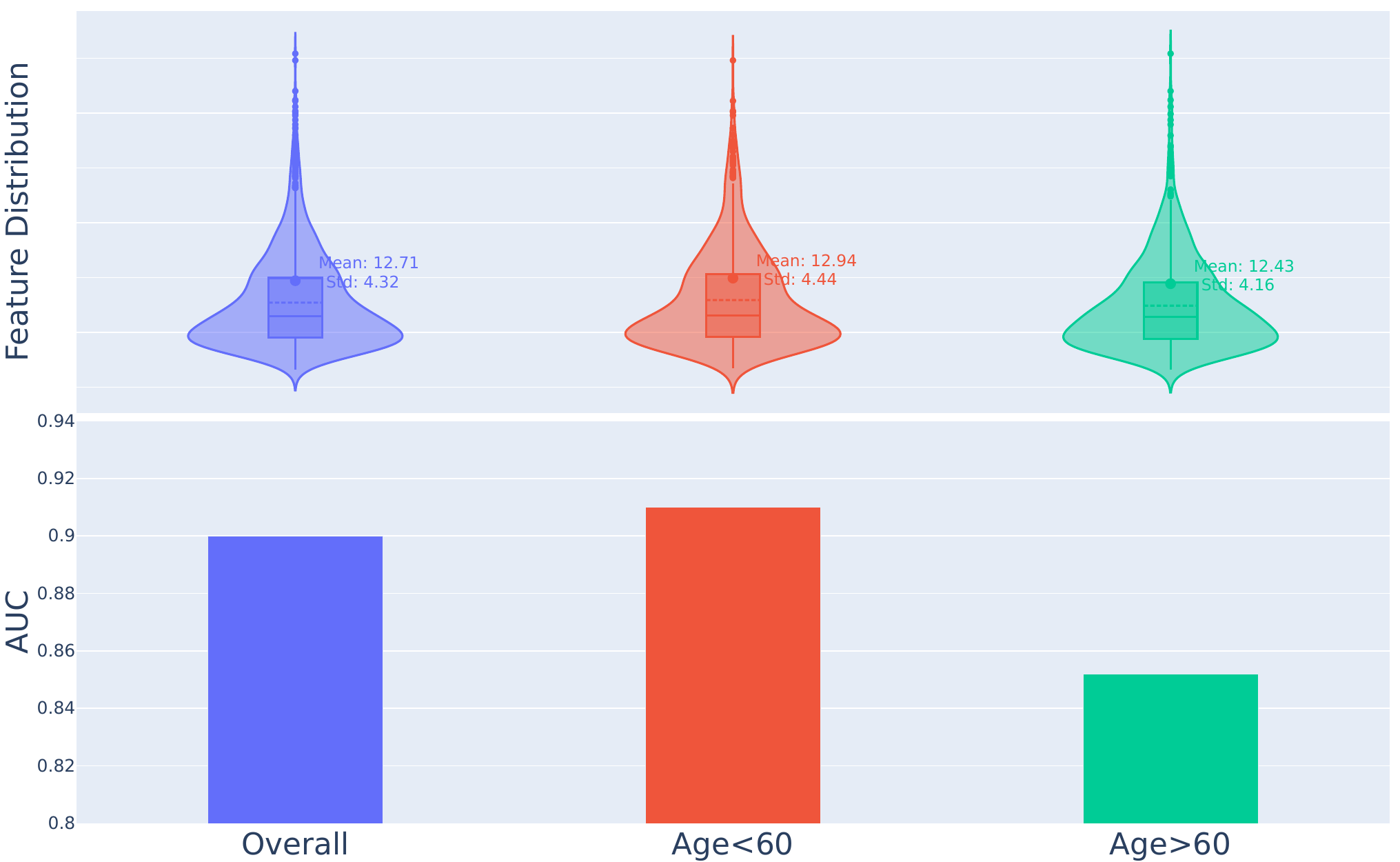}
        \caption{EffNet for SC Det. on Age}
        \label{fig:ham_eff_age}
    \end{subfigure}

    \begin{subfigure}{0.5\textwidth}
        \centering
        \includegraphics[width=\linewidth]{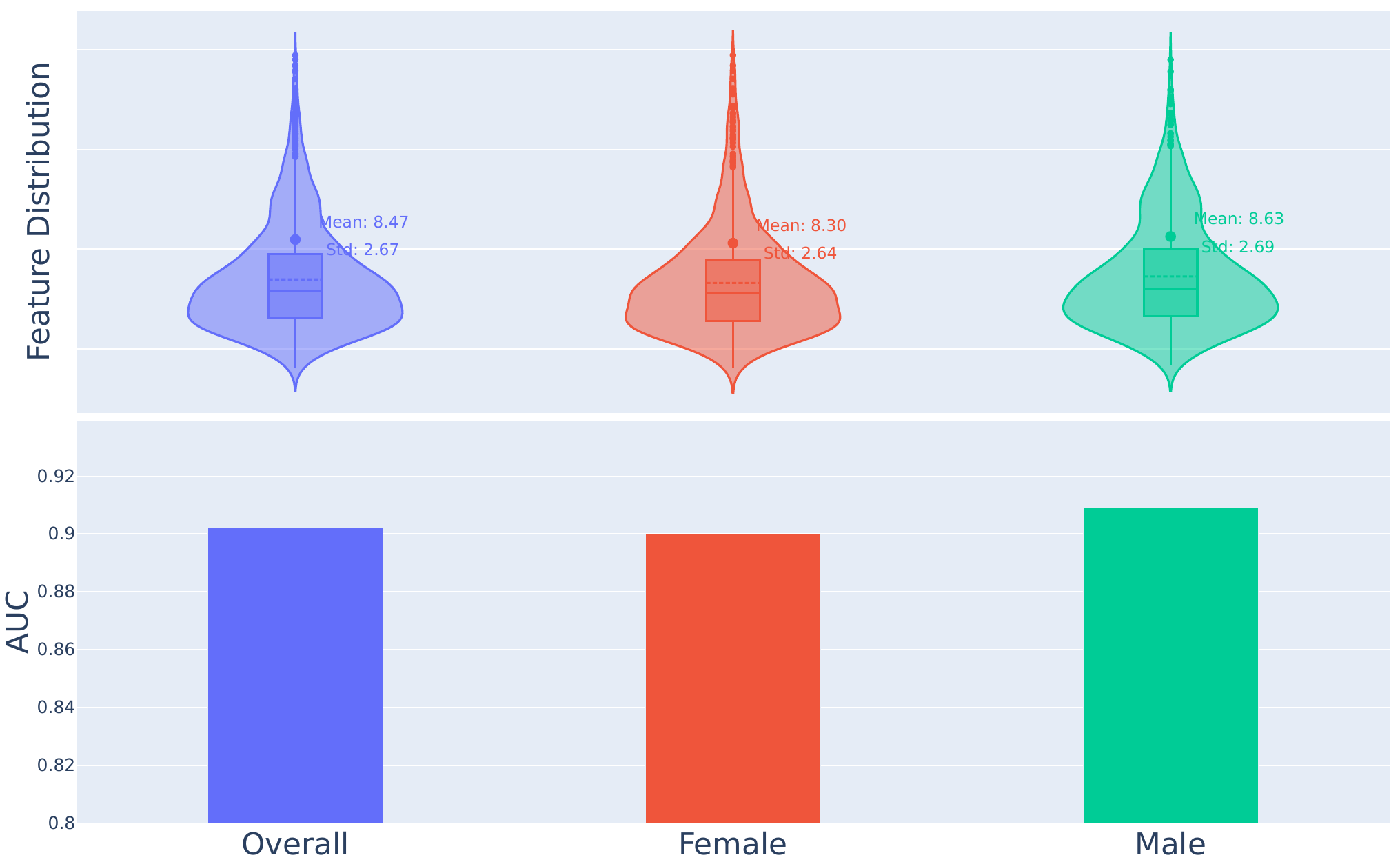}
        \caption{EffNet-FAR for SC Det. on Gender}
        \label{fig:ham_efffar_gender}
    \end{subfigure}%
    \begin{subfigure}{0.5\textwidth}
        \centering
        \includegraphics[width=\linewidth]{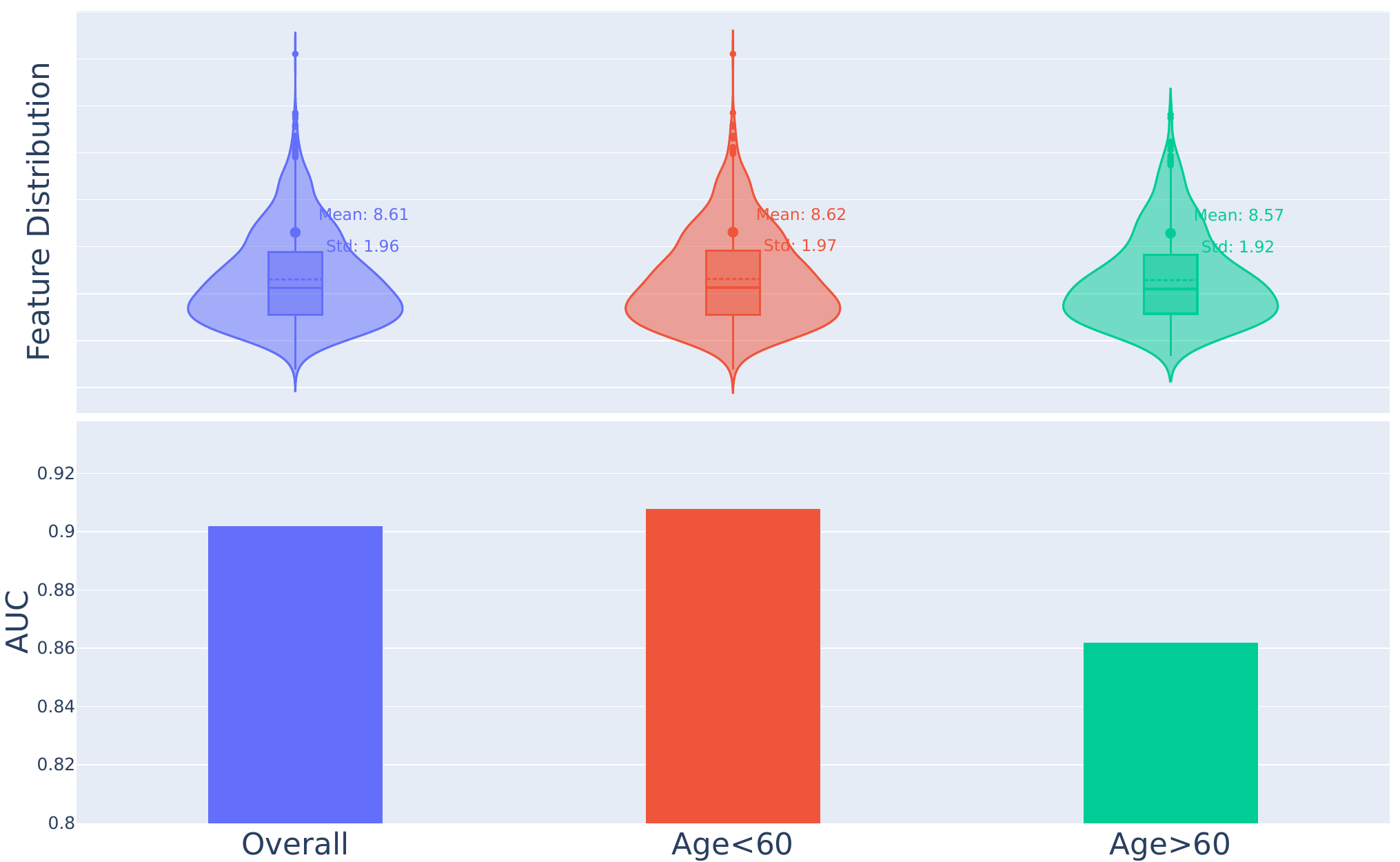}
        \caption{EffNet-FAR for SC Det. on Age}
        \label{fig:ham_efffar_age}
    \end{subfigure}

    \begin{subfigure}{0.8\textwidth}
        \centering
        \includegraphics[width=\linewidth]{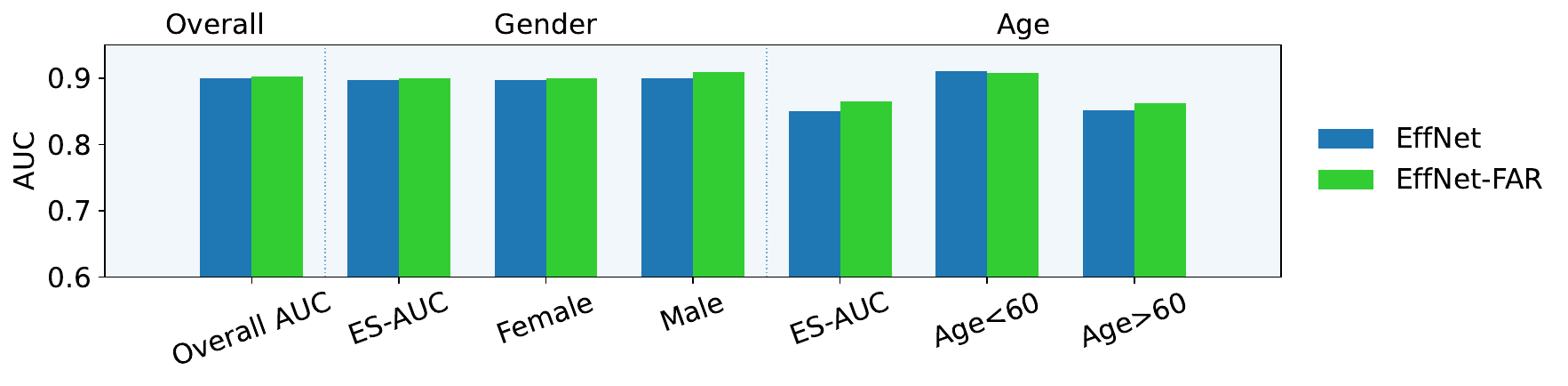}
        \caption{Comparison of EffNet and EffNet-FAR on SC Det.}
        \label{fig:ham_eff_comparison}
    \end{subfigure}
\end{minipage}
\end{adjustbox}

\caption{Feature distribution, AUC performance, and AUC comparison of EfficientNet and EffNet-FAR for Skin Cancer (SC) detection across two demographic attributes, including Gender and Age, on \textbf{HAM10000}.}
\label{fig:ham}
\end{figure*}

\begin{figure*}[h]
\centering
\begin{adjustbox}{scale=0.7}
\begin{minipage}{\linewidth}
\centering
    \begin{subfigure}{0.33\textwidth}
        \centering
        \includegraphics[width=\linewidth]{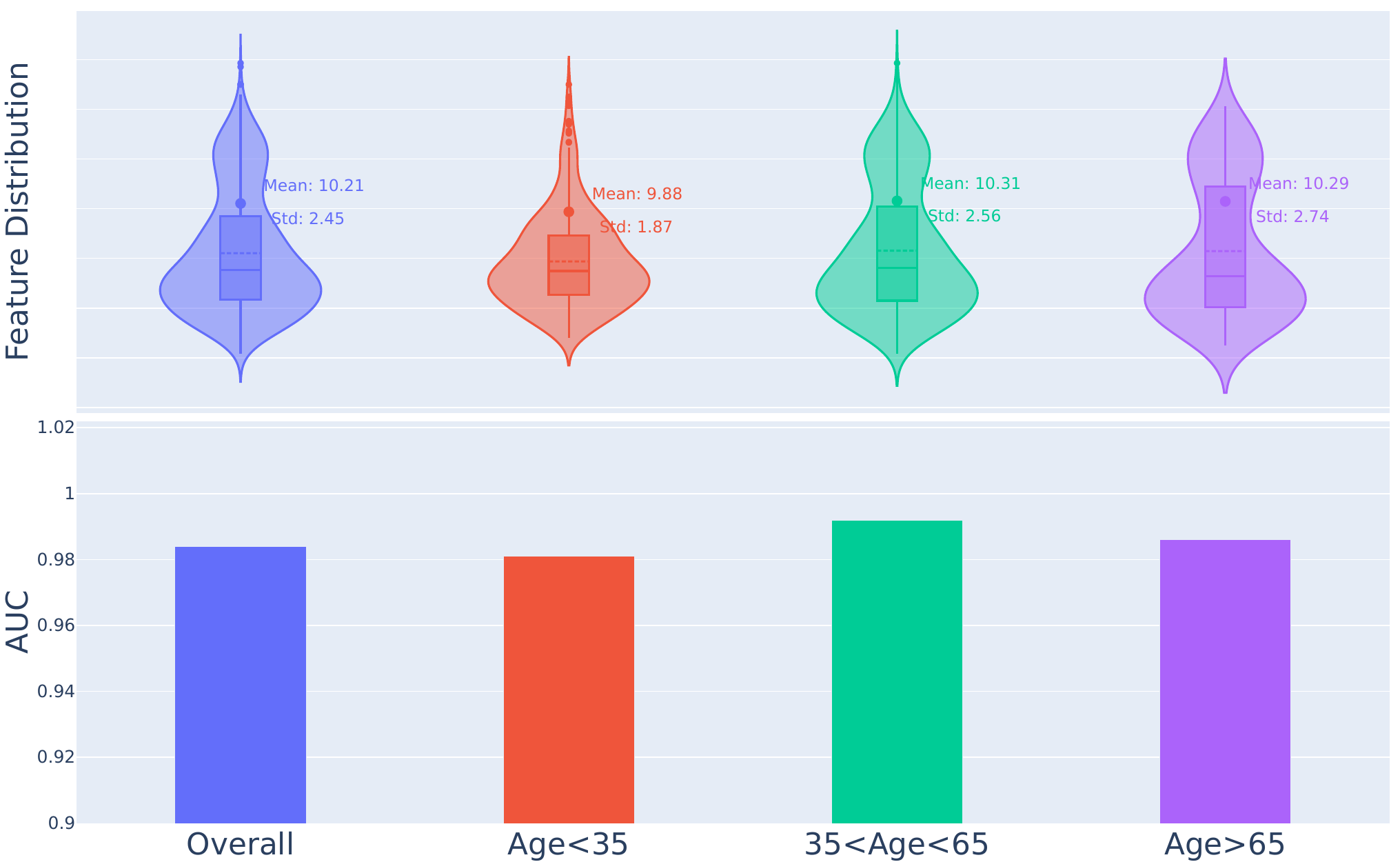}
        \caption{EffNet for Glasses Det. on Age}
        \label{fig:fairface_eff_age}
    \end{subfigure}%
    \begin{subfigure}{0.33\textwidth}
        \centering
        \includegraphics[width=\linewidth]{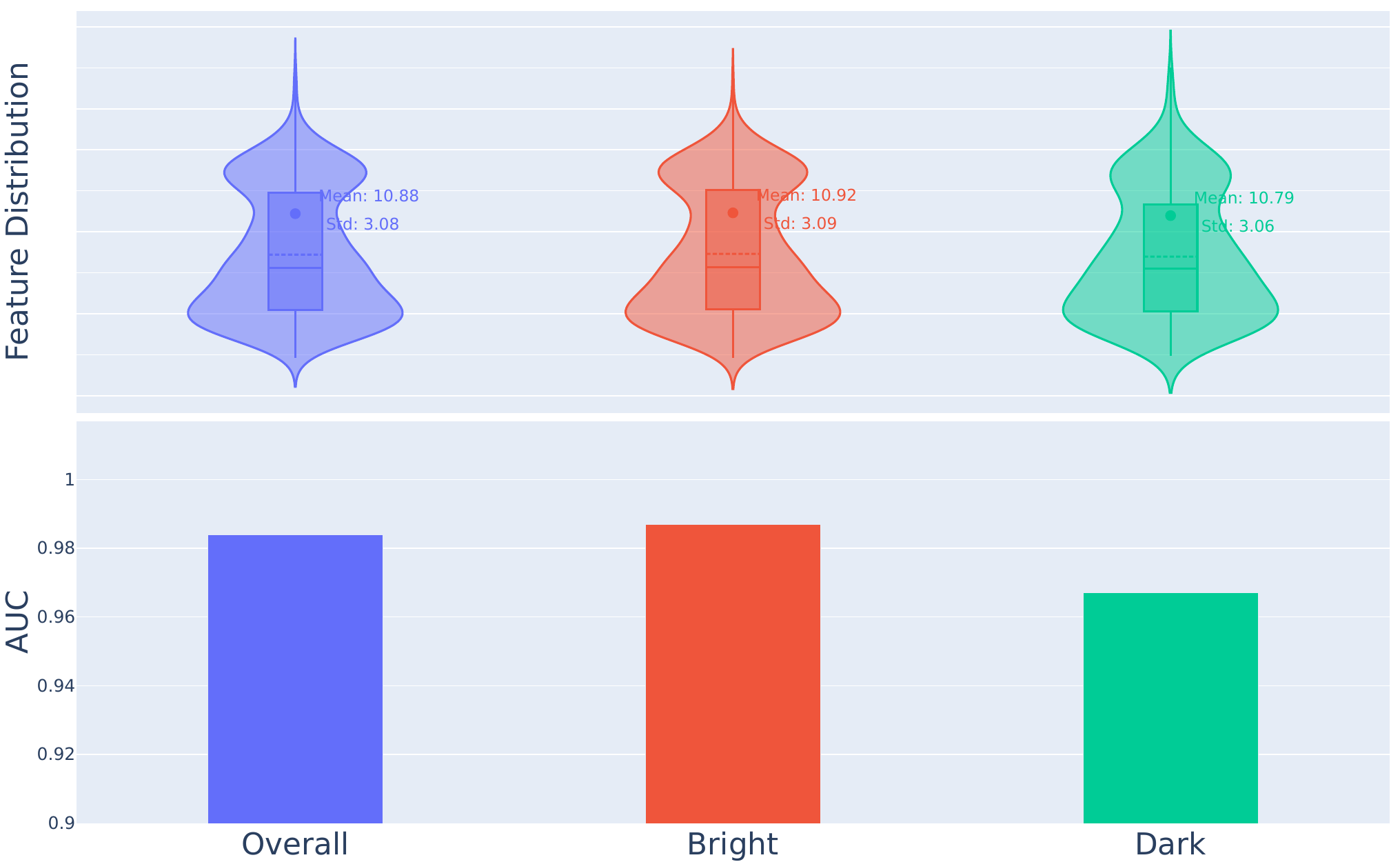}
        \caption{EffNet for Glasses Det. on ST}
        \label{fig:fairface_eff_skin}
    \end{subfigure}%
    \begin{subfigure}{0.33\textwidth}
        \centering
        \includegraphics[width=\linewidth]{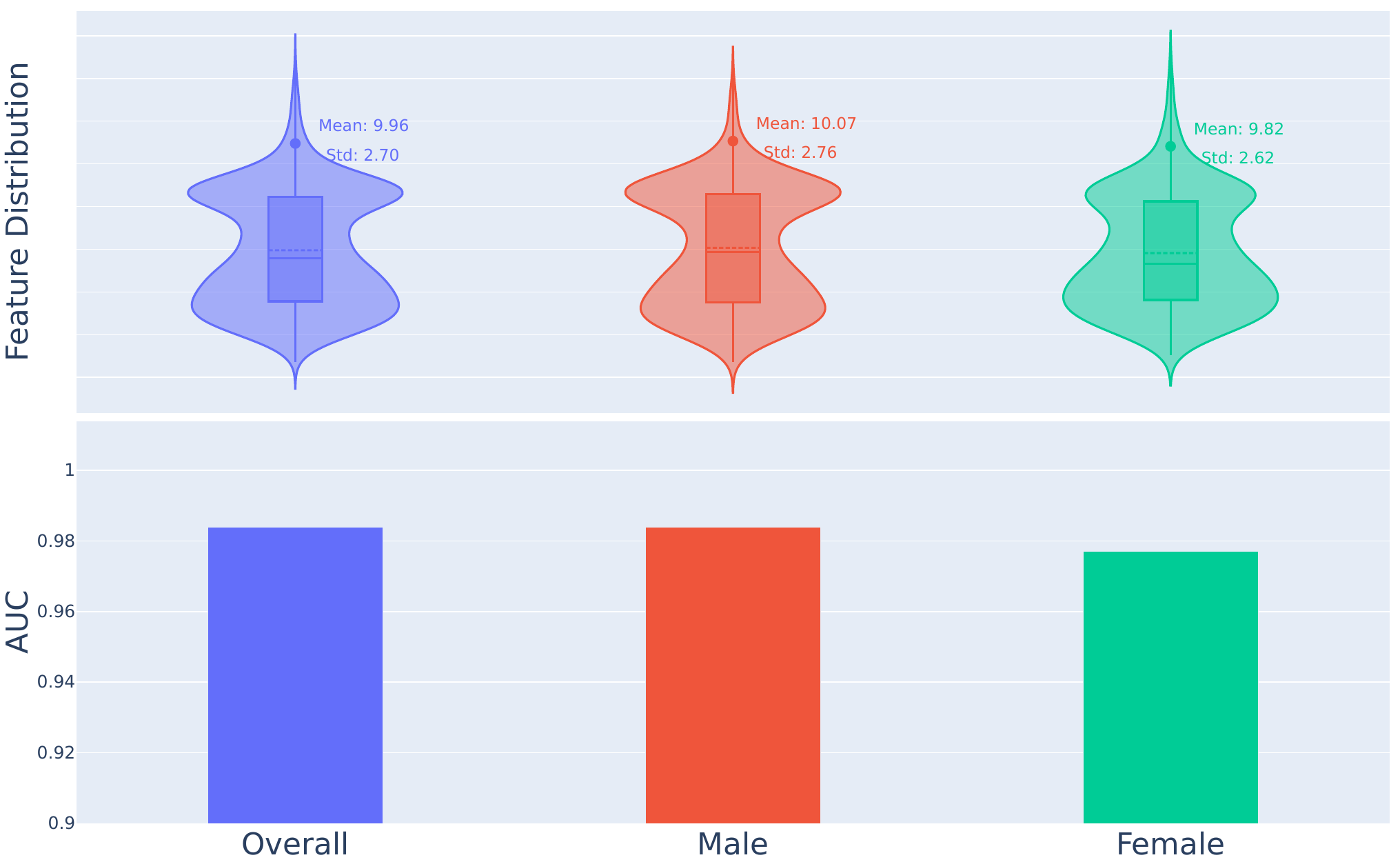}
        \caption{EffNet for Glasses Det. on Gender}
        \label{fig:fairface_eff_gender}
    \end{subfigure}

    \begin{subfigure}{0.33\textwidth}
        \centering
        \includegraphics[width=\linewidth]{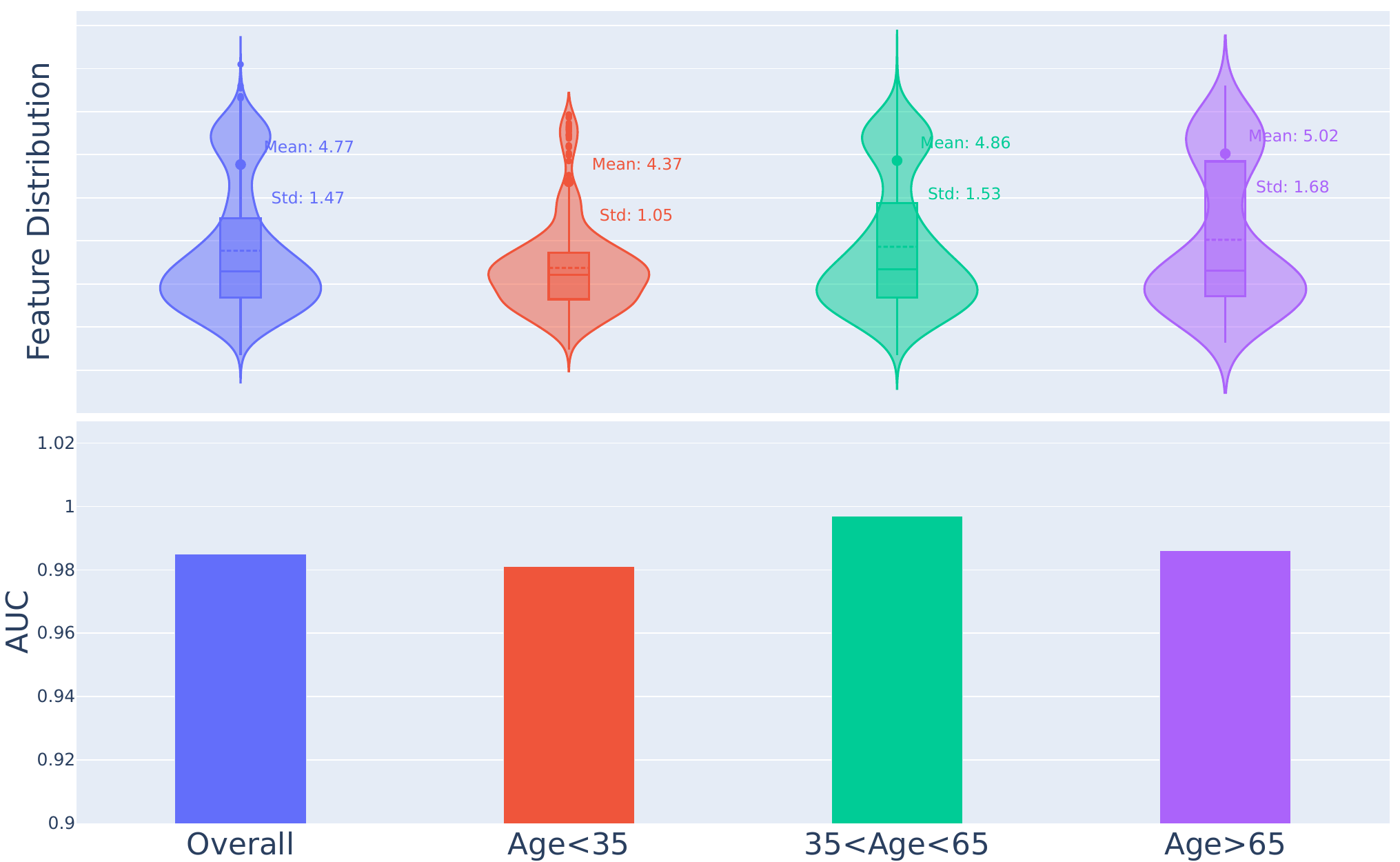}
        \caption{EffNet-FAR for Glasses Det. on Age}
        \label{fig:fairface_efffar_age}
    \end{subfigure}%
    \begin{subfigure}{0.33\textwidth}
        \centering
        \includegraphics[width=\linewidth]{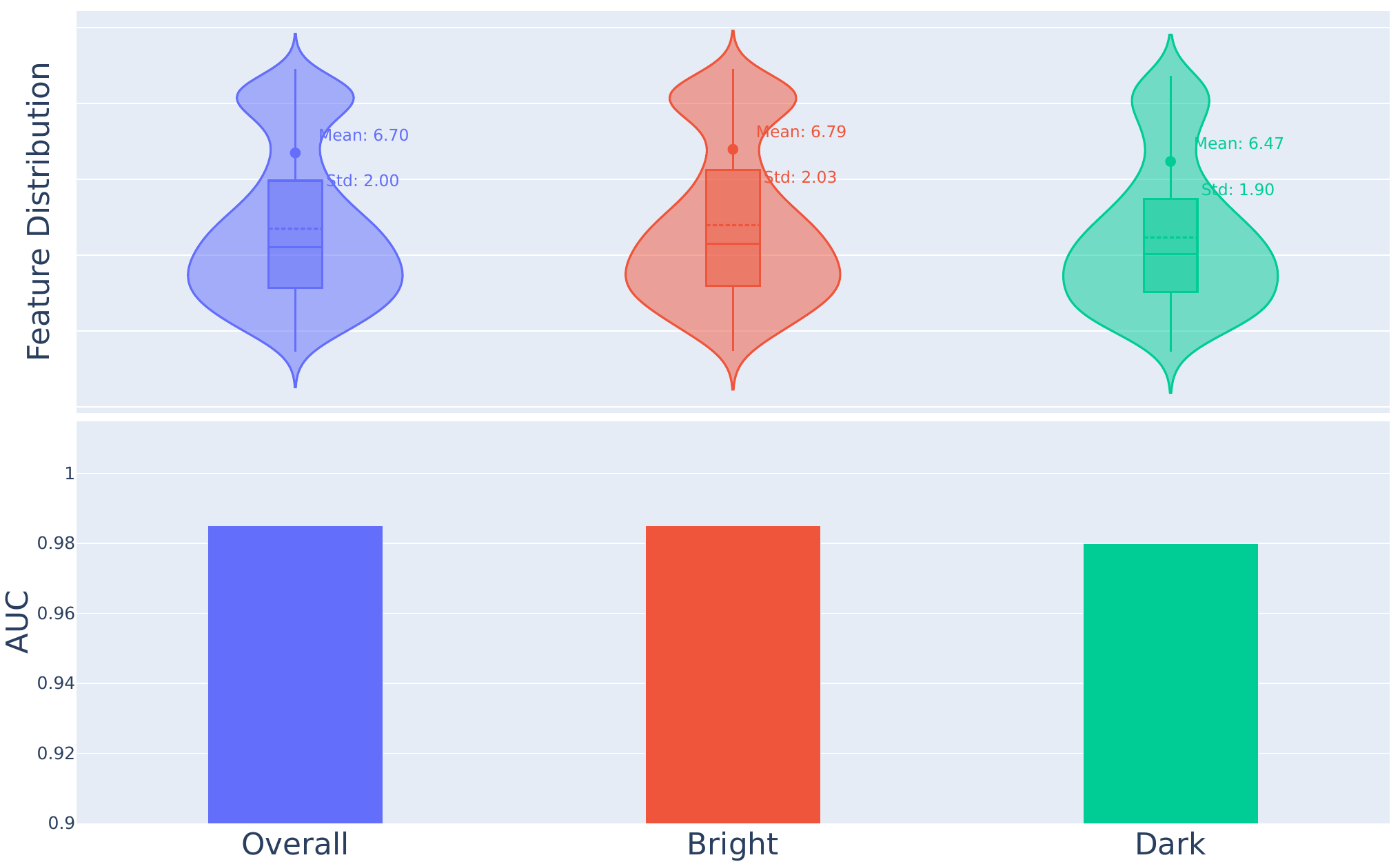}
        \caption{EffNet-FAR for Glasses Det. on ST}
        \label{fig:fairface_efffar_skin}
    \end{subfigure}%
    \begin{subfigure}{0.33\textwidth}
        \centering
        \includegraphics[width=\linewidth]{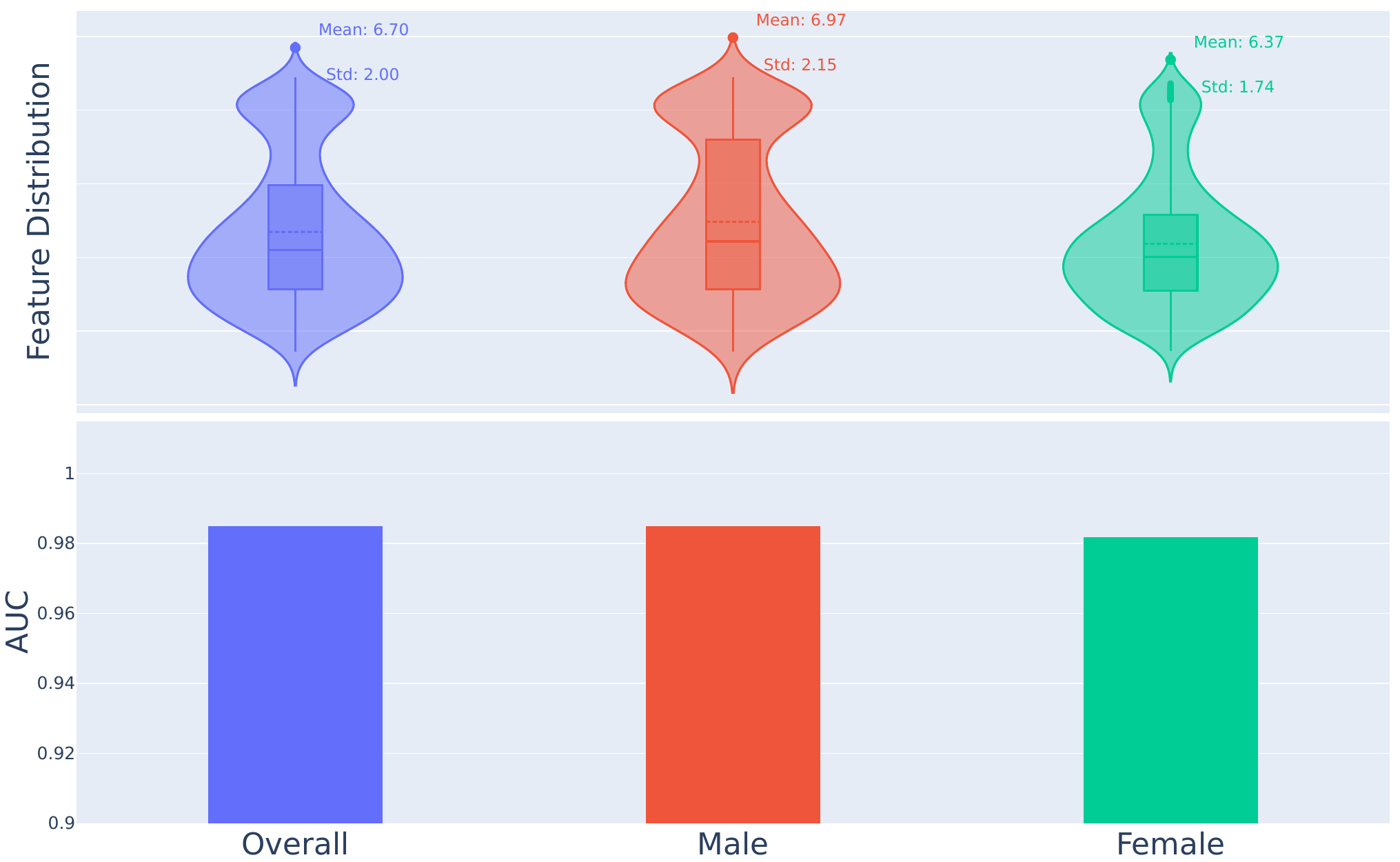}
        \caption{EffNet-FAR for Glasses Det. on Gender}
        \label{fig:fairface_efffar_gender}
    \end{subfigure}

    \begin{subfigure}{0.8\textwidth}
        \centering
        \includegraphics[width=\linewidth]{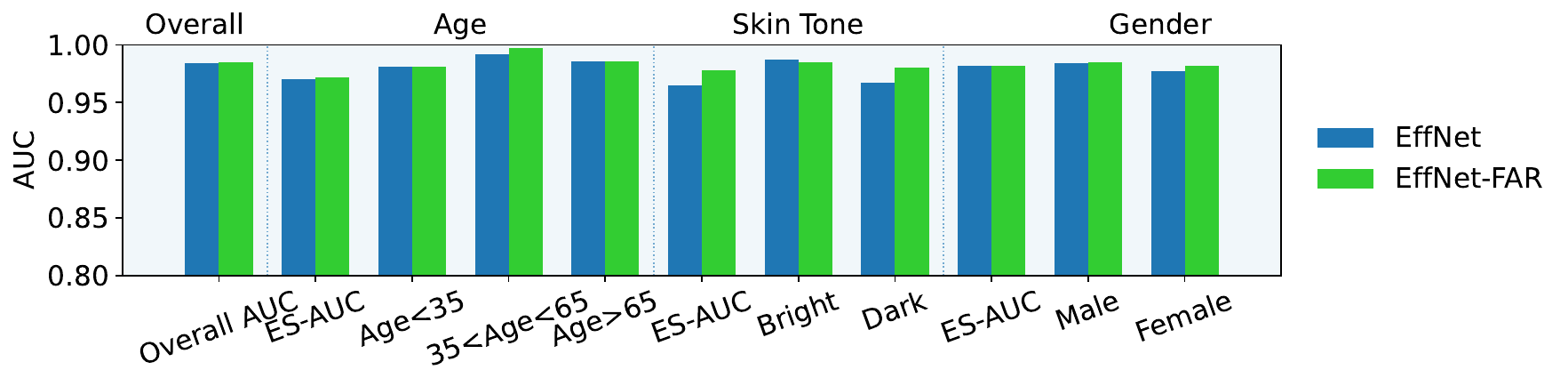}
        \caption{Comparison of EffNet and EffNet-FAR on Glasses Det.}
        \label{fig:fairface_eff_comparison}
    \end{subfigure}
\end{minipage}
\end{adjustbox}

\caption{Feature distribution, AUC performance, and AUC comparison of EfficientNet and EffNet-FAR for Glasses detection across three demographic attributes, including Age, Skin Tone (ST), and Gender, on \textbf{FairFace}.}
\label{fig:fairface}
\end{figure*}

We further extended our experiments to the CheXpert, HAM10000, and FairFace datasets to evaluate model performance on additional medical and natural image tasks. Considering the superior performance of EfficientNet on the FairVision dataset compared to ViT, we primarily focused on EfficientNet’s subgroup performance in these datasets. The results for CheXpert, HAM10000, and FairFace are presented in Fig.~\ref{fig:chexpert}, Fig.~\ref{fig:ham}, and Fig.~\ref{fig:fairface}, respectively. From these results, it is apparent that detection performance also varies significantly across demographic groups. For instance, in the CheXpert dataset, disparities in AUC are observed in pleural effusion detection across race and gender attributes. Similarly, for HAM10000, skin lesion detection shows variations across age and gender groups. In the FairFace dataset, eyeglass detection performance differs among subgroups defined by age, skin tone, and gender. Again, FAR consistently improves ES-AUC, suggesting that fairness-aware regularization mitigates disparities and yields more equitable performance across subgroups.

\begin{figure*}[h]
\centering
\begin{adjustbox}{scale=0.7}
\begin{minipage}{\linewidth}
\centering
    \begin{subfigure}{0.5\textwidth}
        \centering
        \includegraphics[width=\linewidth]{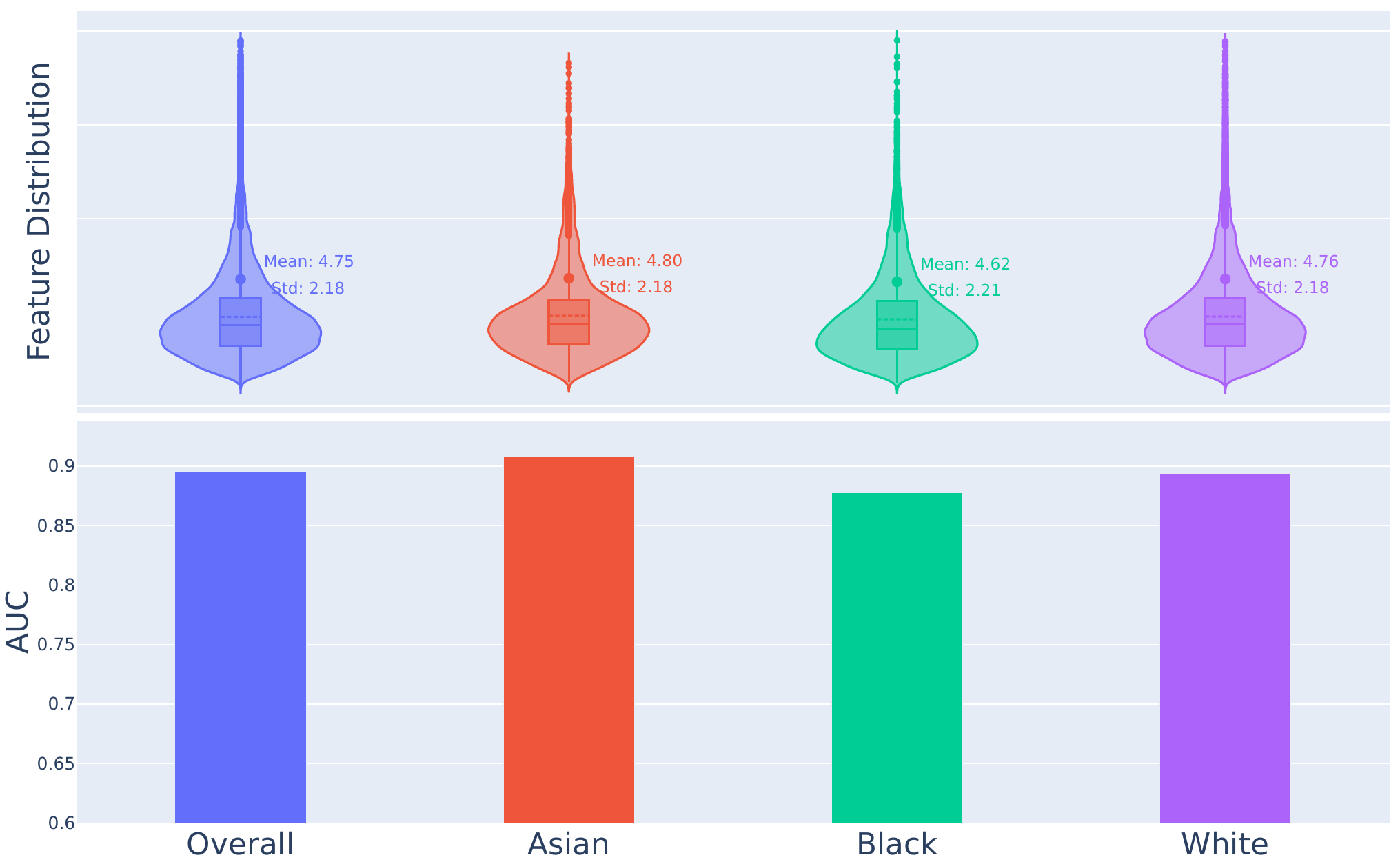}
        \caption{TabTrans for Income Prediction on Race}
        \label{fig:acs_tab_race}
    \end{subfigure}%
    \begin{subfigure}{0.5\textwidth}
        \centering
        \includegraphics[width=\linewidth]{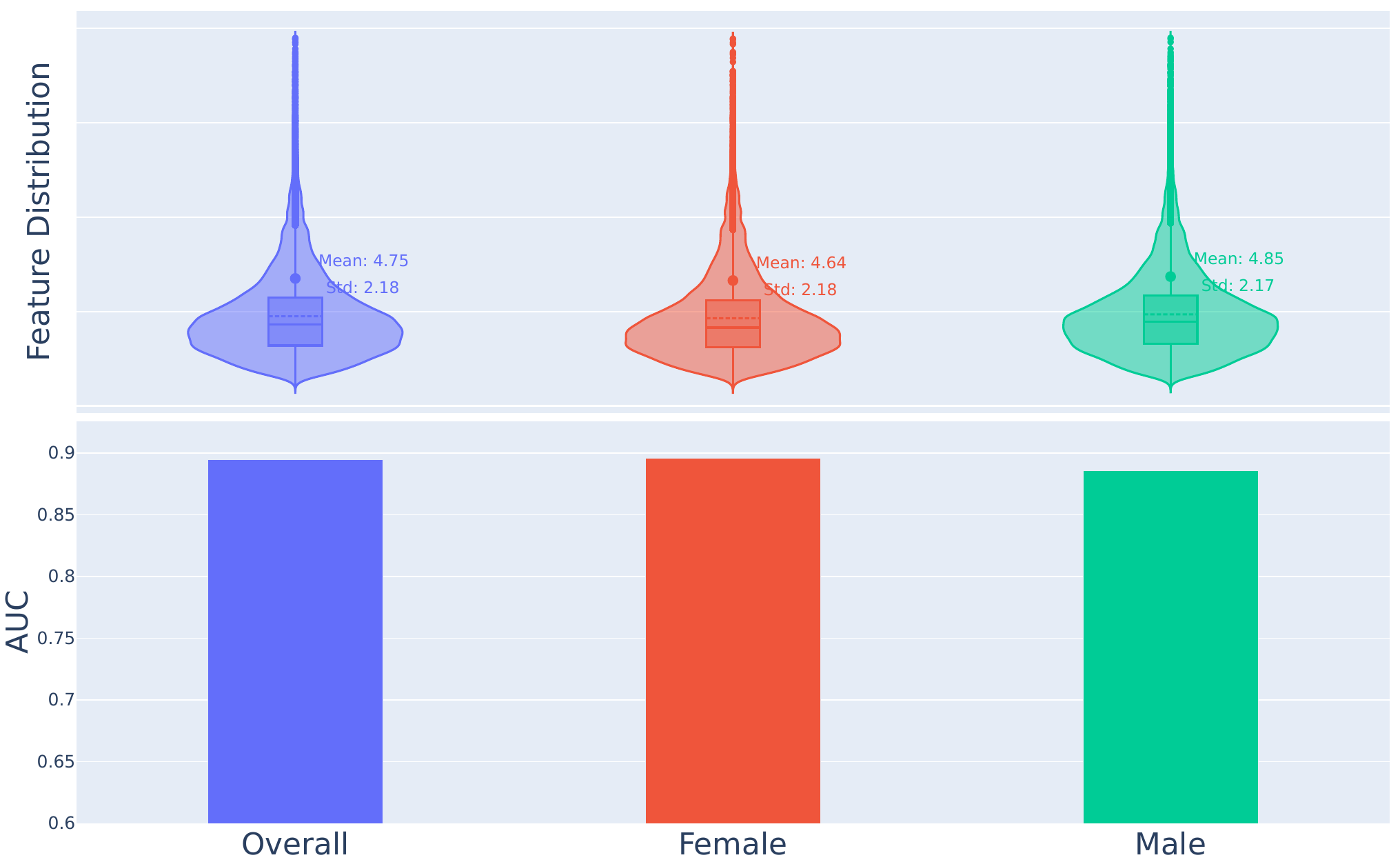}
        \caption{TabTrans for Income Prediction on Gender}
        \label{fig:acs_tab_gender}
    \end{subfigure}

    \begin{subfigure}{0.48\textwidth}
        \centering
        \includegraphics[width=\linewidth]{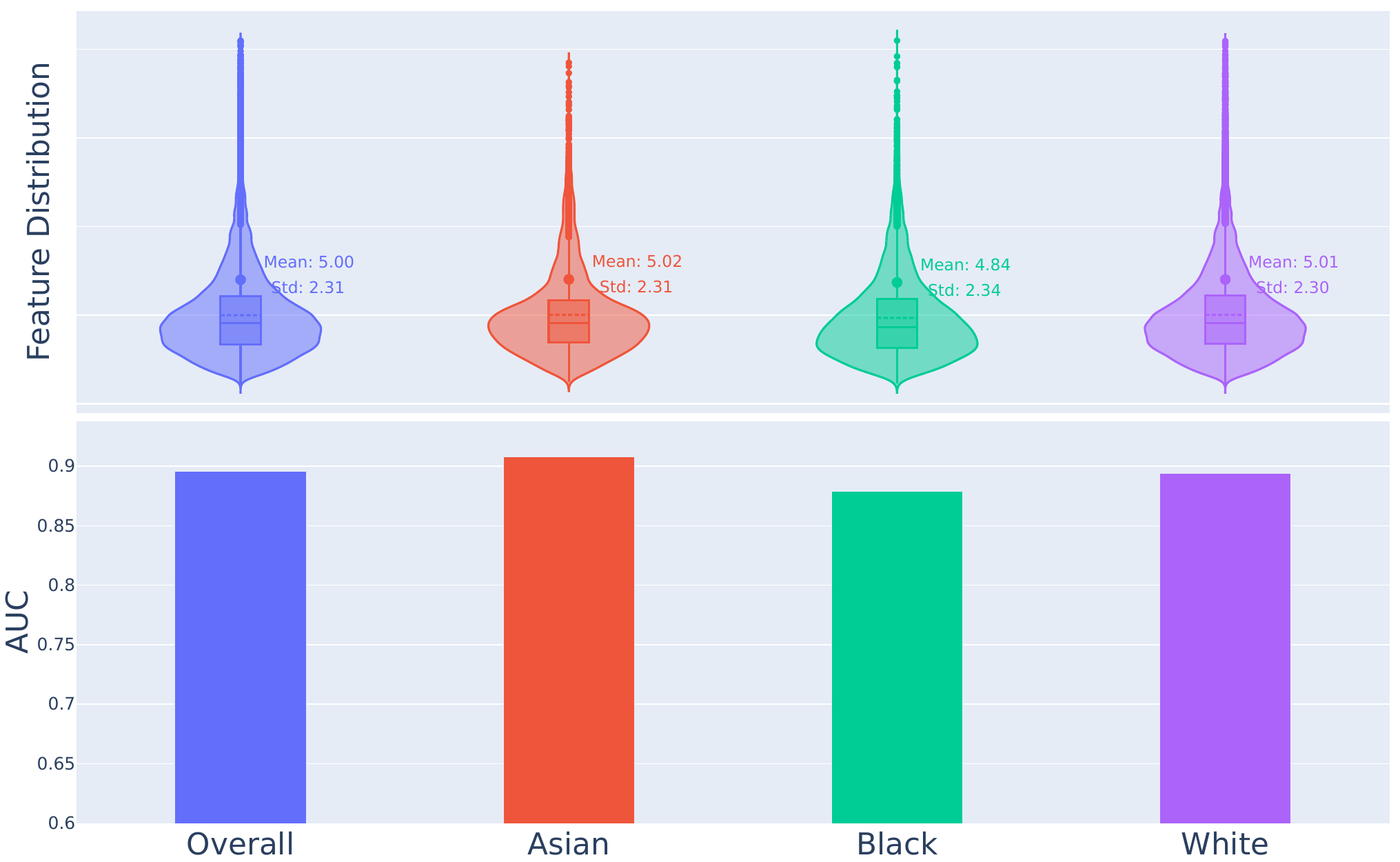}
        \caption{TabTrans-FAR for Income Prediction on Race}
        \label{fig:acs_tabfar_race}
    \end{subfigure}\hfill
    \begin{subfigure}{0.48\textwidth}
        \centering
        \includegraphics[width=\linewidth]{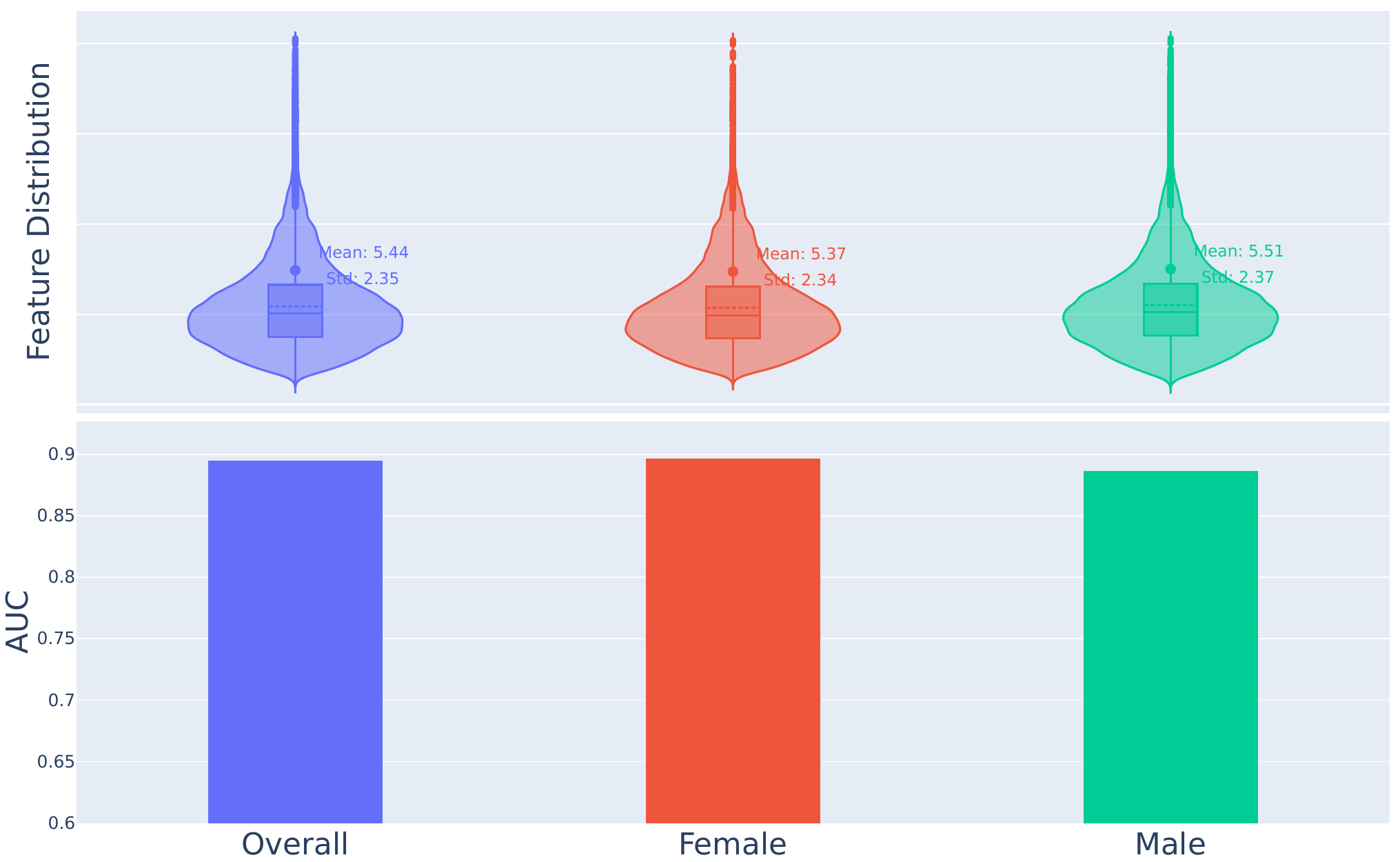}
        \caption{TabTrans-FAR for Income Prediction on Gender}
        \label{fig:acs_tabfar_gender}
    \end{subfigure}

    \begin{subfigure}{0.8\textwidth}
        \centering
        \includegraphics[width=\linewidth]{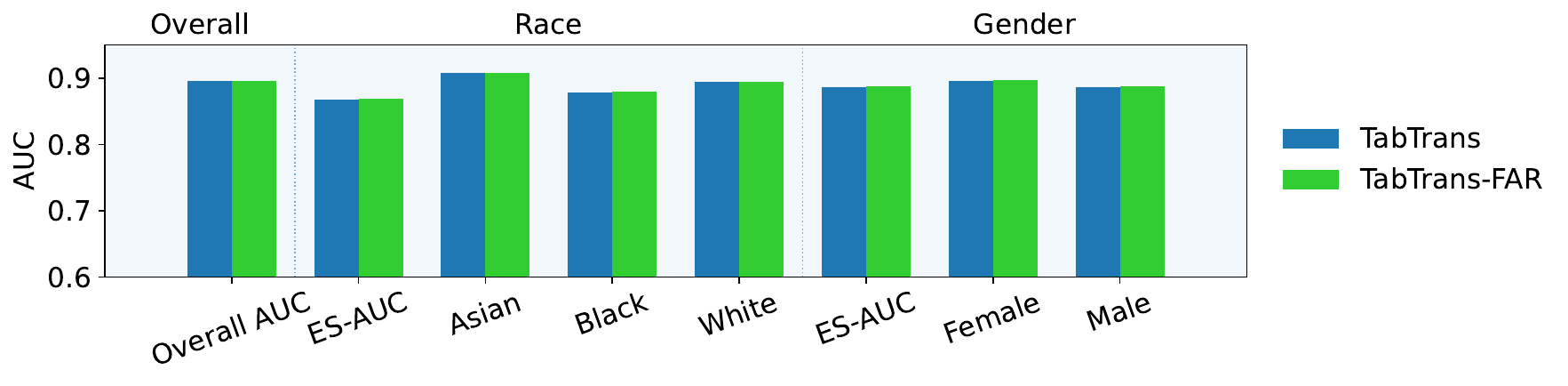}
        \caption{Comparison of TabTrans and TabTrans-FAR on Income Prediction}
        \label{fig:acs_tab_comparison}
    \end{subfigure}
\end{minipage}
\end{adjustbox}

\caption{Feature distribution, AUC performance, and AUC comparison of TabTransformer (TabTrans) and TabTransformer-FAR for Income prediction across two demographic attributes, including Race and Gender, on \textbf{ACSIncome}.}
\label{fig:acs}
\end{figure*}

\begin{figure*}[h]
\centering
\begin{adjustbox}{scale=0.7}
\begin{minipage}{\linewidth}
\centering
    \begin{subfigure}{0.5\textwidth}
        \centering
        \includegraphics[width=\linewidth]{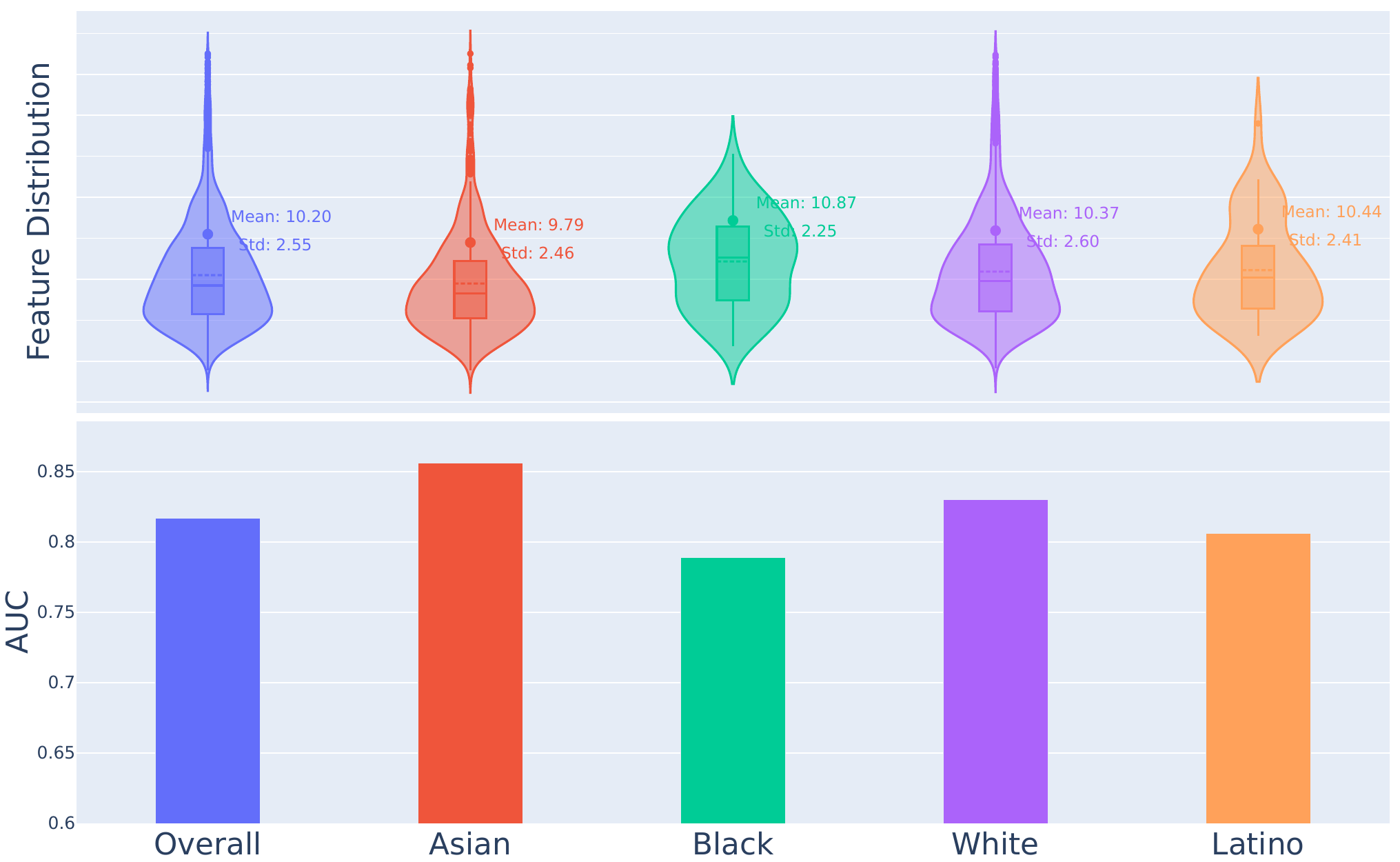}
        \caption{RoBERTa for TC Det. on Race}
        \label{fig:civil_roberta_race}
    \end{subfigure}%
    \begin{subfigure}{0.5\textwidth}
        \centering
        \includegraphics[width=\linewidth]{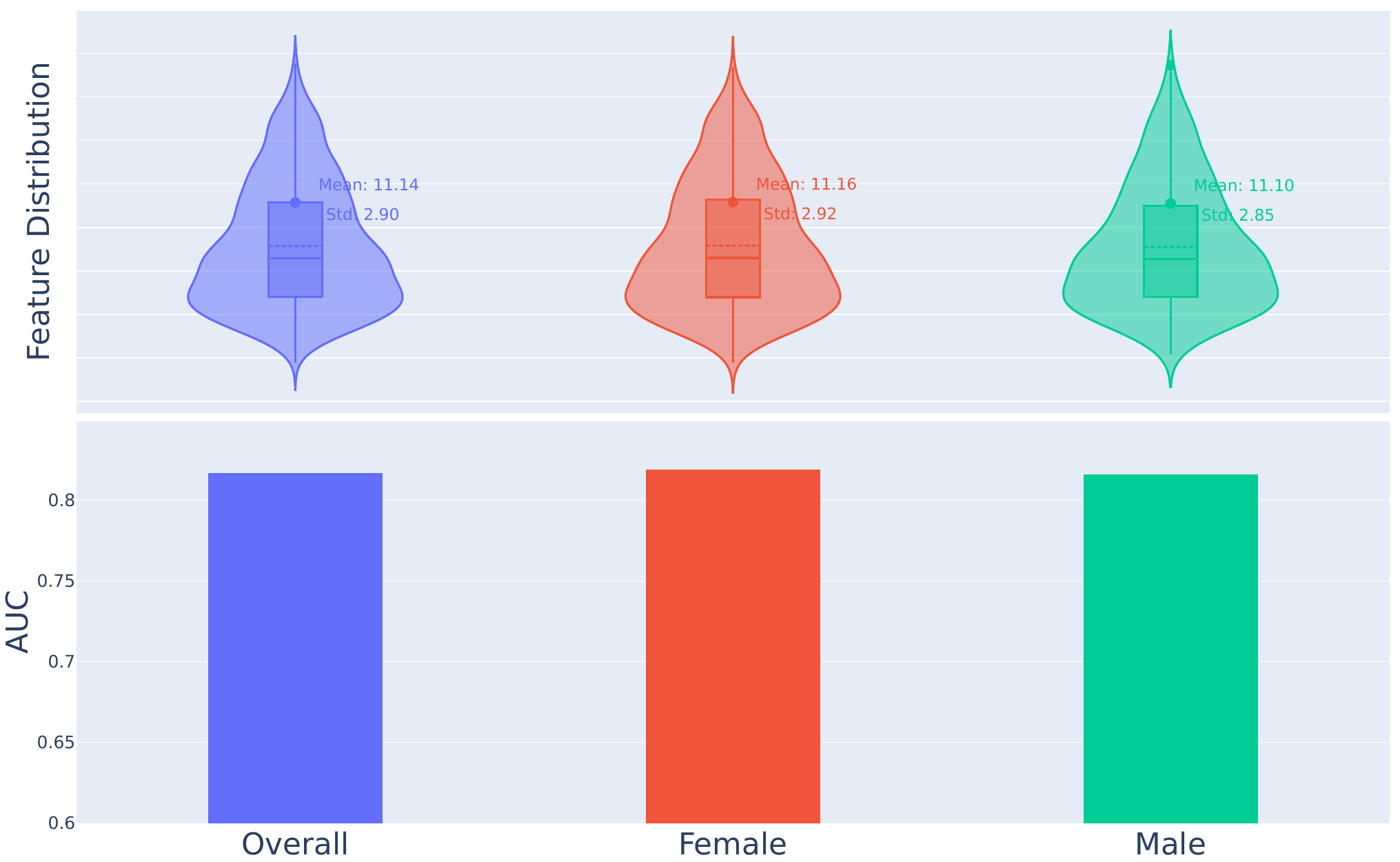}
        \caption{RoBERTa for TC Det. on Gender}
        \label{fig:civil_roberta_gender}
    \end{subfigure}

    \begin{subfigure}{0.48\textwidth}
        \centering
        \includegraphics[width=\linewidth]{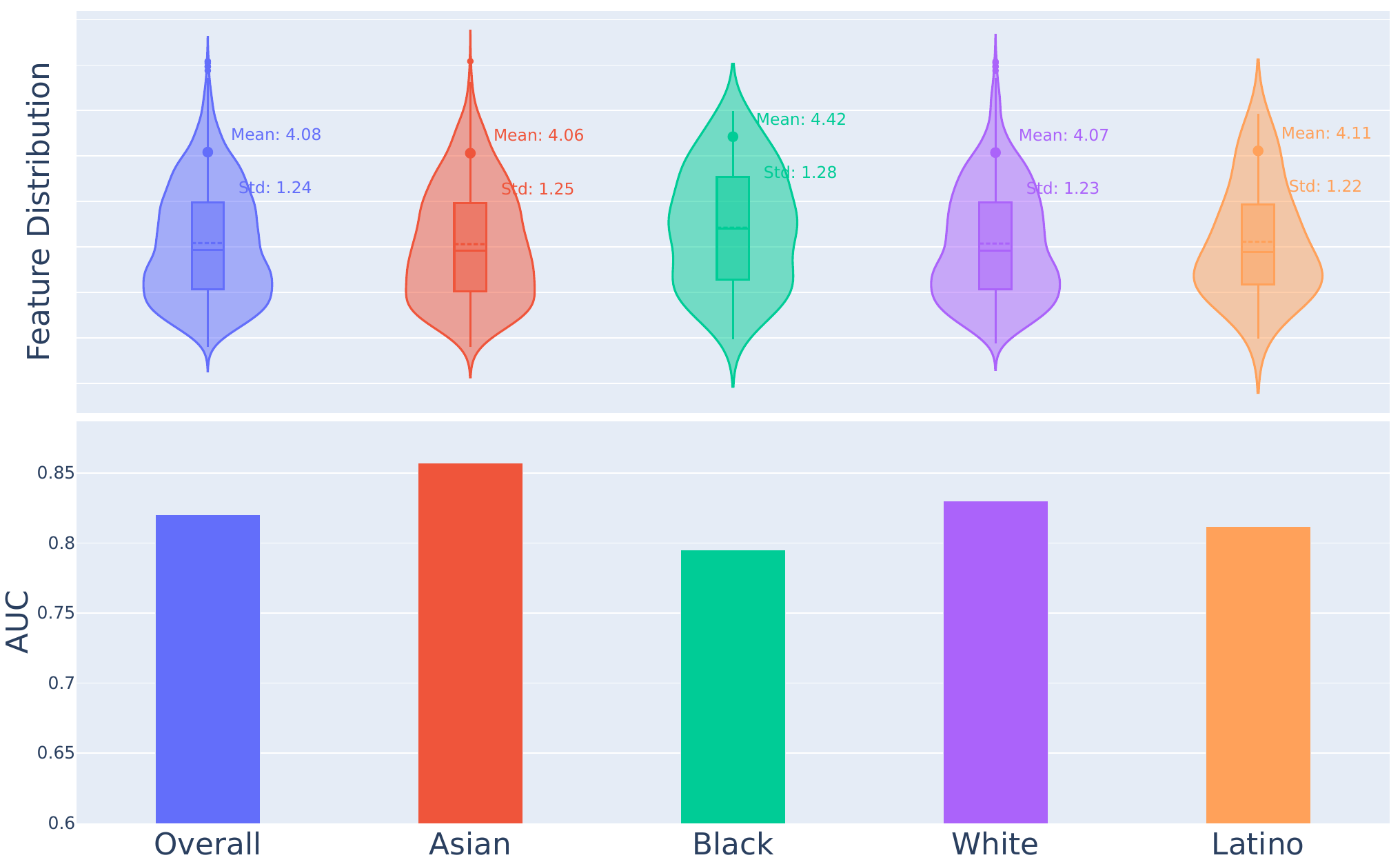}
        \caption{RoBERTa-FAR for TC Det. on Race}
        \label{fig:civil_robertafar_race}
    \end{subfigure}\hfill
    \begin{subfigure}{0.48\textwidth}
        \centering
        \includegraphics[width=\linewidth]{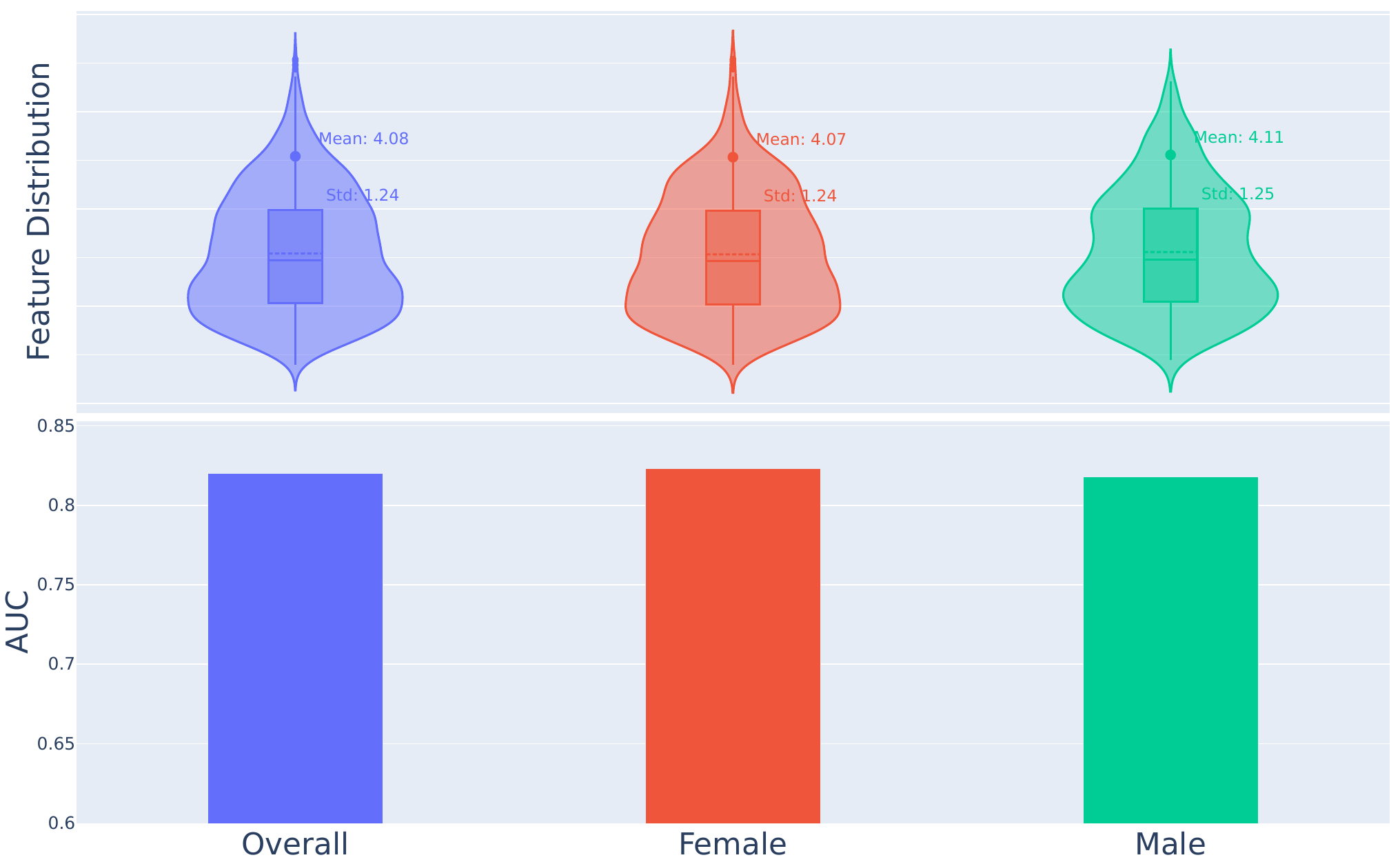}
        \caption{RoBERTa-FAR for TC Det. on Gender}
        \label{fig:civil_robertafar_gender}
    \end{subfigure}

    \begin{subfigure}{0.8\textwidth}
        \centering
        \includegraphics[width=\linewidth]{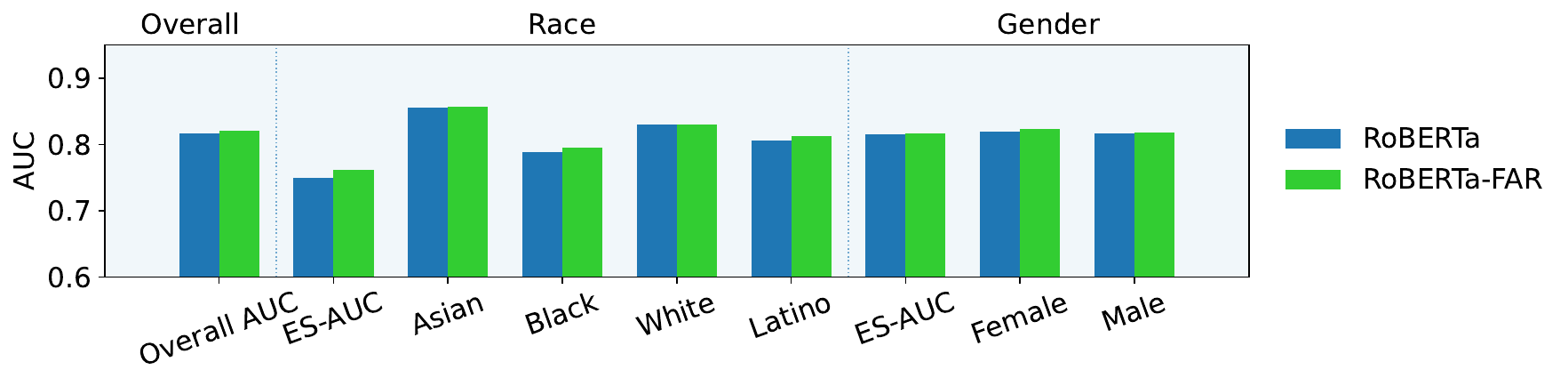}
        \caption{Comparison of RoBERTa and RoBERTa-FAR on TC Det.}
        \label{fig:civil_roberta_comparison}
    \end{subfigure}
\end{minipage}
\end{adjustbox}

\caption{Feature distribution, AUC performance, and AUC comparison of RoBERTa and RoBERTa-FAR for Toxic Comment (TC) detection across two demographic attributes, including Race and Gender, on \textbf{CivilComments-WILDS}.}
\label{fig:civilcomments}
\end{figure*}

In addition to image datasets, we further evaluated fairness on tabular and natural language benchmarks. For the \textit{ACS Income} dataset, we trained TabTransformer to predict whether an individual’s annual income exceeds \$50,000. As shown in Fig.~\ref{fig:acs}, the models achieve high overall accuracy but exhibit pronounced disparities across demographic groups, with the Black subgroup consistently underperforming compared to White and Asian counterparts. For the \textit{CivilComments-WILDS} dataset, we fine-tuned RoBERTa models for toxic comment detection. The results in Fig.~\ref{fig:civilcomments} demonstrate strong overall AUC yet reveal substantial subgroup gaps, particularly along race and gender. Importantly, incorporating fairness-aware regularization (FAR) improves ES-AUC across both datasets, indicating better performance balance among demographic groups. These observations confirm that fairness challenges are not confined to medical imaging but also arise in structured tabular data and natural language processing tasks.

    
\noindent\textbf{Discussion}.
Our experimental results across multiple datasets demonstrate the relationship between data distributions and fairness guarantees, a connection formalized by our theoretical framework, particularly Theorem \ref{thm:accuracy} and Corollary \ref{thm:correlation_lossbound_featdist}. Let's analyze the results for each dataset:
\\
\noindent\textit{FairVision Dataset} For AMD detection (Fig. \ref{fig:amd}), both ViT and EfficientNet show significant performance variations across demographic groups. With ViT, the overall feature distribution has a mean of 5.92 and standard deviation of 2.46. The Asian group (mean: 6.26, std: 2.53) and Black group (mean: 6.45, std: 2.65) show larger deviations from the overall distribution compared to the White group (mean: 5.81, std: 2.41). Our Corollary \ref{thm:correlation_lossbound_featdist} predicts that a group's excess loss is bounded by a weighted sum of the statistical distances (in both feature centroids and covariances) to \textit{all other} demographic groups. A significant deviation from the overall distribution's statistics, as observed empirically for the Asian and especially the Black groups, implies large average distances to the other constituent groups that make up the overall population. The theory, therefore, provides a direct explanation for why these groups with more distant feature distributions are expected to have a higher upper bound on their loss, which is consistent with their observed lower AUC performance.
\\
\noindent Similar patterns emerge for DR detection (Fig. \ref{fig:dr}), where the Black group shows the largest deviation (mean: 6.66, std: 1.93) from the overall distribution (mean: 6.24, std: 2.30), corresponding to lower AUC values. For Glaucoma detection (Fig. \ref{fig:gl}), the feature distribution differences are particularly pronounced between racial categories, with the Black group consistently showing the largest deviation and correspondingly lower performance. These empirical results strongly support our theoretical framework, which links such performance disparities directly to the distances between group-level feature distributions.

\noindent\textit{CheXpert Dataset} The results for pleural effusion detection (Fig. \ref{fig:chexpert}) demonstrate how distributional differences affect fairness across both race and gender. The feature distributions show notable variations, with the Black group exhibiting the largest deviation (mean: 11.50, std: 7.43) from the overall distribution (mean: 10.85, std: 6.21). This aligns with Corollary 3.20, which predicts larger expected loss bounds when feature centroids are more distant from the overall centroid.

\noindent\textit{HAM10000 Dataset} For skin cancer detection (Fig. \ref{fig:ham}), the age-based disparities are particularly notable. The feature distribution for patients aged $>$60 (mean: 12.43, std: 4.16) differs significantly from the overall distribution (mean: 12.71, std: 4.32), leading to performance disparities that align with our theoretical predictions about the relationship between distribution differences and fairness guarantees.

\noindent\textit{FairFace Dataset} The glasses detection results (Fig. \ref{fig:fairface}) show how multiple demographic attributes interact with fairness guarantees. The feature distributions vary across age groups, skin color, and gender, with the age$>$65 group showing the largest deviation (mean: 10.29, std: 2.74) from the overall distribution (mean: 10.21, std: 2.45). This multi-attribute variation demonstrates the complexity of achieving fairness when dealing with intersectional demographic characteristics.

\noindent\textit{ACS Income Dataset} The income prediction results (Fig.~\ref{fig:acs}) reveal pronounced disparities across racial subgroups. The feature distribution for the Black subgroup (mean: 4.62, std: 2.21) deviates considerably from the overall distribution (mean: 4.75, std: 2.18), leading to a lower subgroup AUC. This finding is consistent with our theoretical analysis, which suggests that greater divergence between subgroup and overall feature centroids is linked to weaker fairness guarantees and degraded predictive accuracy.

\noindent\textit{CivilComments-WILDS Dataset} For toxic comment detection (Fig.~\ref{fig:civilcomments}), disparities are evident across both race and gender. The feature distribution for comments mentioning the Black identity (mean: 2.25, std: 10.87) differs from the overall distribution (mean: 2.55, std: 10.20), aligning with observed subgroup performance gaps. These results reinforce our theoretical framework, indicating that larger shifts from the global centroid correspond to amplified fairness risks and lower reliability of subgroup performance.

    \section{Conclusion}
\label{sec:concl}
This work presents a theoretical analysis of how data distributional differences impact the fairness guarantees of deep learning models. Our results show that fairness is fundamentally constrained by inter-group shifts in feature means and covariances, as reflected in the derived bounds on fairness errors, convergence rates, and group-specific risks. Building on these insights, we proposed Fairness-Aware Regularization (FAR), a practical intervention that directly minimizes inter-group discrepancies in feature distributions. Extensive experiments across six large-scale datasets spanning medical images, natural images, tabular data, and natural language confirm our theory: demographic subgroups with larger distributional shifts consistently exhibit lower predictive performance, while incorporating FAR improves overall AUC, ES-AUC, and subgroup equity. Together, these findings bridge theoretical guarantees with algorithmic design, offering a principled and scalable pathway toward fairer deep learning models. Future research directions include extending the framework to multimodal foundation models and exploring adaptive strategies that tailor fairness interventions to specific subgroup characteristics.



	
	\ifCLASSOPTIONcompsoc
	\section*{Acknowledgments}
	\else
	\section*{Acknowledgment}
	\fi
    This work was supported by NIH R00 EY028631, NIH R21 EY035298, NIH P30 EY003790, Research To Prevent Blindness International Research Collaborators Award, and Alcon Young Investigator Grant. We also acknowledge the generous funding resources provided by NYU Abu Dhabi with code AD131.

	\bibliographystyle{IEEEtran}
	\bibliography{ref}
	

 \begin{IEEEbiography}[{\includegraphics[width=1in,height=1.25in,clip,keepaspectratio]{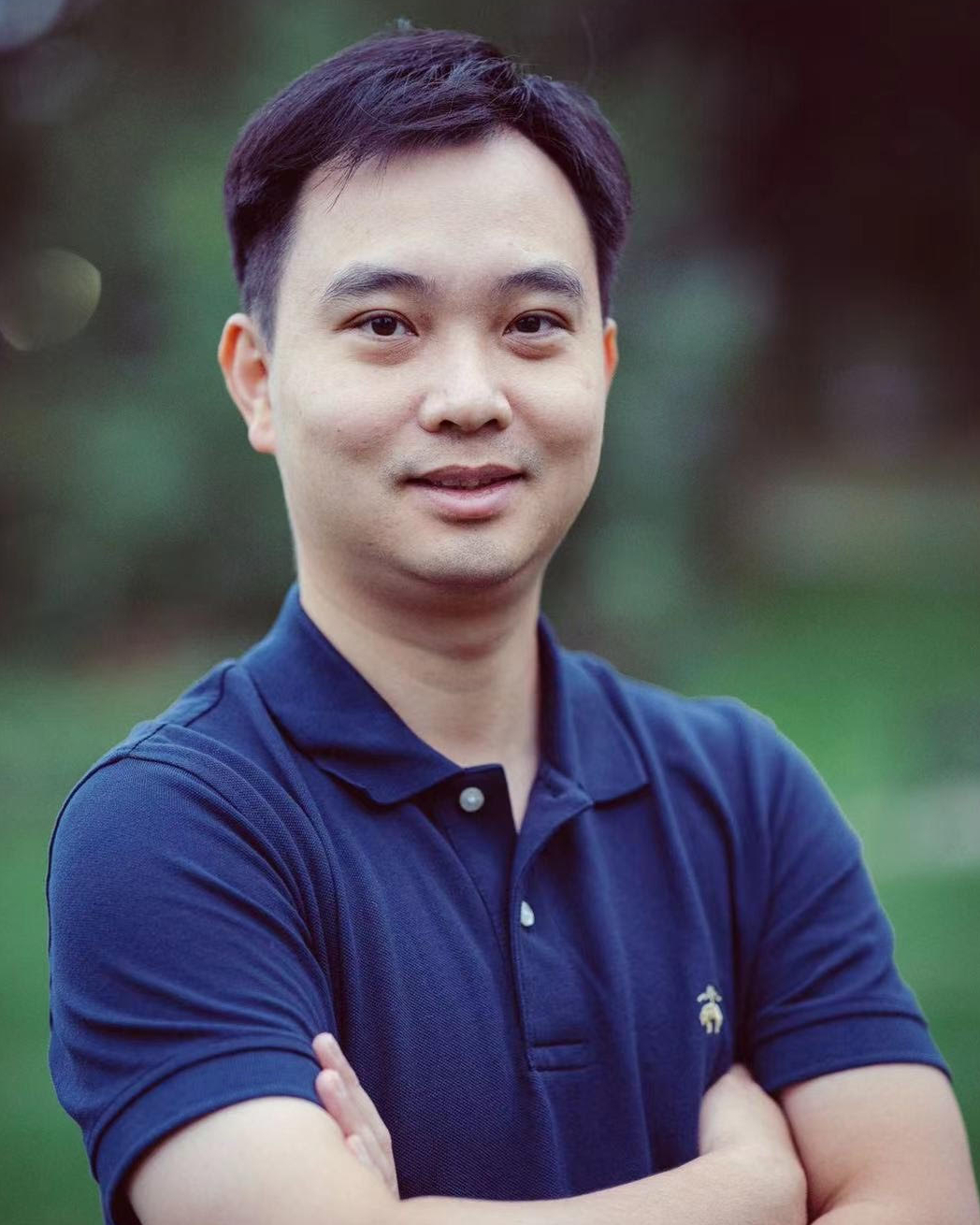}}]{Yan Luo} (Member, IEEE) is currently a postdoctoral research fellow at the Harvard AI and Robotics lab and an affiliate of the Broad Institute of MIT and Harvard. He obtained a Ph.D. degree in Computer Science from the University of Minnesota (UMN), Twin Cities. Before joining UMN, he conducted Ph.D. research at the Visual Information Processing Laboratory at the National University of Singapore (NUS) and worked as a Research Assistant at the Sensor-Enhanced Social Media (SeSaMe) Centre, part of the Interactive and Digital Media Institute at NUS. His research interests include responsible AI and equitable deep learning. He is a member of the IEEE since 2023.
\end{IEEEbiography}

\begin{IEEEbiography}[{\includegraphics[width=1in,height=1.25in,clip,keepaspectratio]{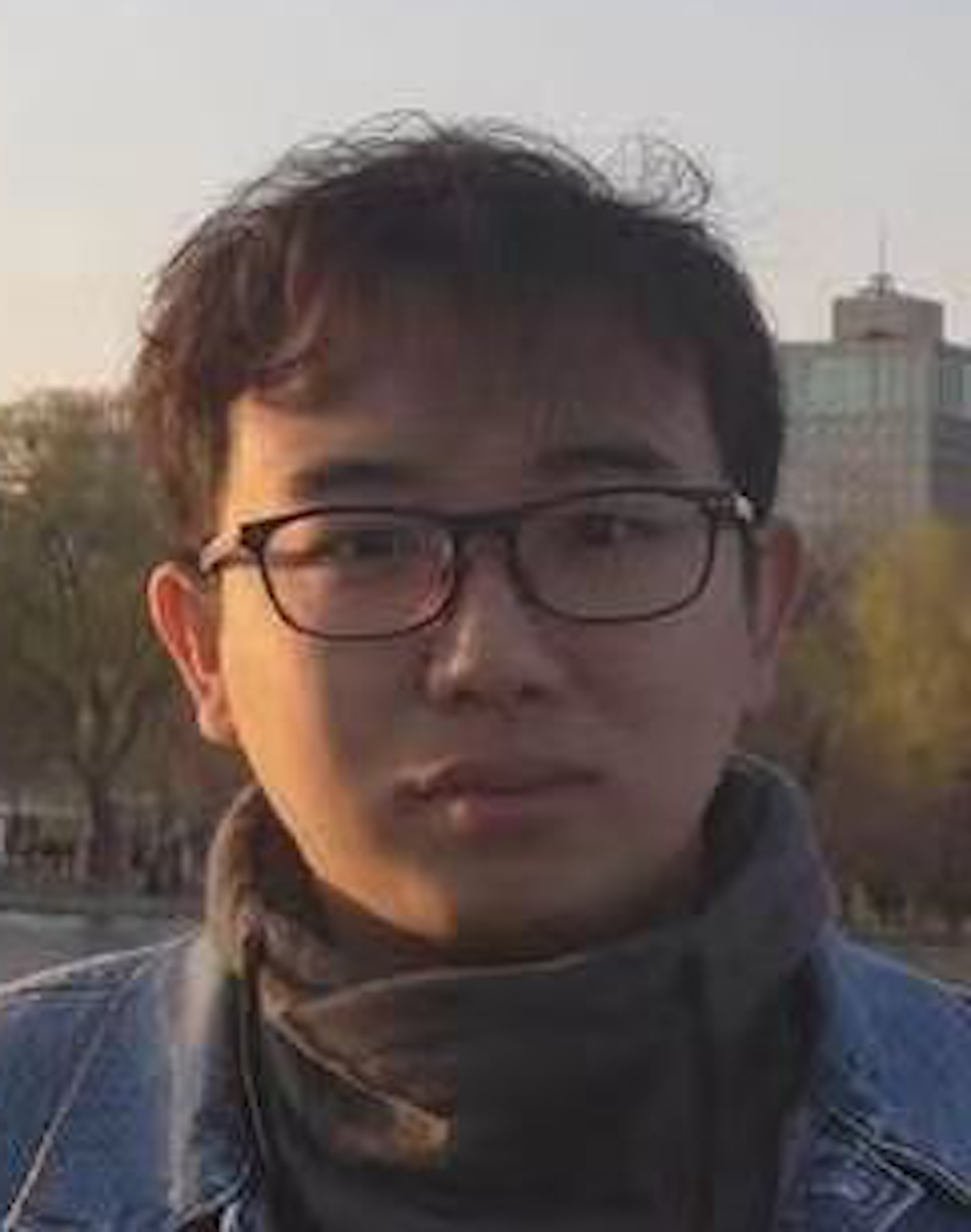}}]{Congcong Wen} (Member, IEEE) is currently a Postdoctoral Associate in the Department of Electrical and Computer Engineering at New York University and New York University Abu Dhabi, as well as a Visiting Research Fellow at Harvard University. He received his B.S. degree in Geographic Information Systems from the China University of Petroleum, China, in 2016, and his Ph.D. degree from the Aerospace Information Research Institute, Chinese Academy of Sciences, China, in 2021. His research interests include multimodal artificial intelligence, 2D/3D computer vision, and foundation models.
\end{IEEEbiography}

\begin{IEEEbiography}[{\includegraphics[width=1in,height=1.25in,clip,keepaspectratio]{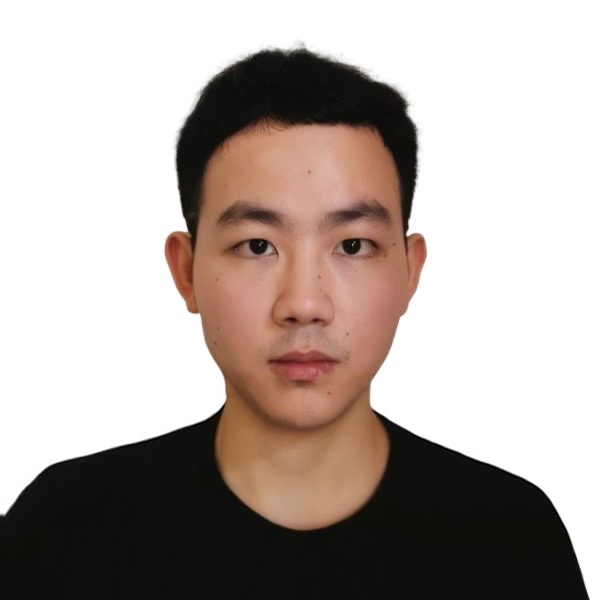}}]{Min Shi} (Member, IEEE) received the Ph.D. degree in 2020 at the College of Engineering and Computer Science, Florida Atlantic University, Boca Raton, FL, USA. He completed his postdoctoral training at Washington University in St. Louis and Harvard Medical School in 2022 and 2024, respectively. His current research includes data mining, medical image analysis, graph learning and bioinformatics.
\end{IEEEbiography}

\begin{IEEEbiography}[{\includegraphics[width=1in,height=1.25in,clip]{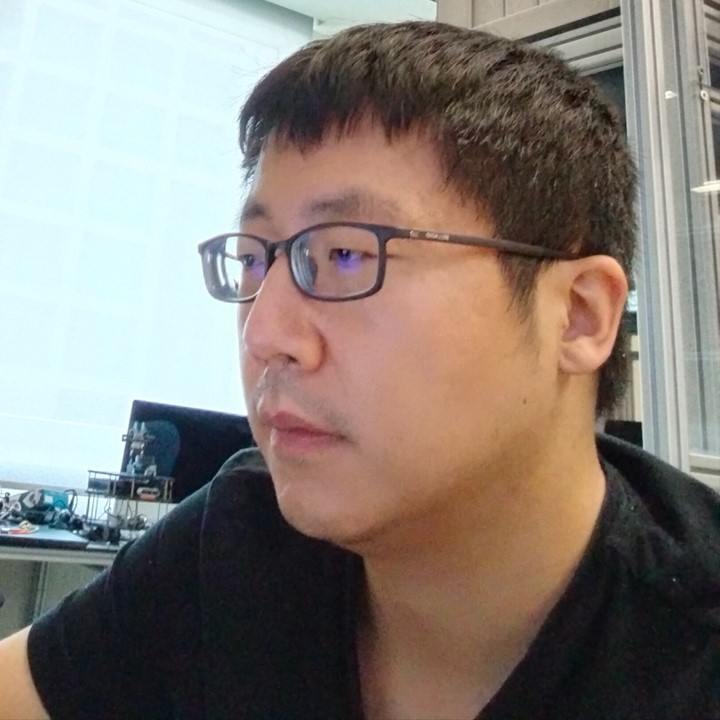}}]
{Hao Huang} received a B.S. degree in Digital Media Technology from the Beijing University of Posts and Telecommunications, China, an M.S. degree in Computer Science from the University of Rochester, and a Ph.D. degree in Computer Science from New York University. He is currently a post-doctoral associate at New York University Abu Dhabi. His research interests include three-dimensional computer vision, robotics, and bioinformatics.
\end{IEEEbiography}

\begin{IEEEbiography}[{\includegraphics[width=1in,height=1.25in,clip]{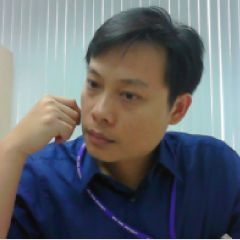}}]
{Yi Fang} received the B.S. and M.S. degrees in biomedical engineering from Xi’an Jiaotong University, Xi’an, China, in 2003 and 2006, respectively, and the Ph.D. degree in mechanical engineering from Purdue University, West Lafayette, IN, USA, in 2011.
He is currently an Associate Professor with the Department of Electrical and Computer Engineering, New York University Abu Dhabi, Abu Dhabi, United Arab Emirates. His research interests include three-dimensional computer vision and pattern recognition.
\end{IEEEbiography}    

\begin{IEEEbiography}[{\includegraphics[width=1in,height=1.25in,clip]{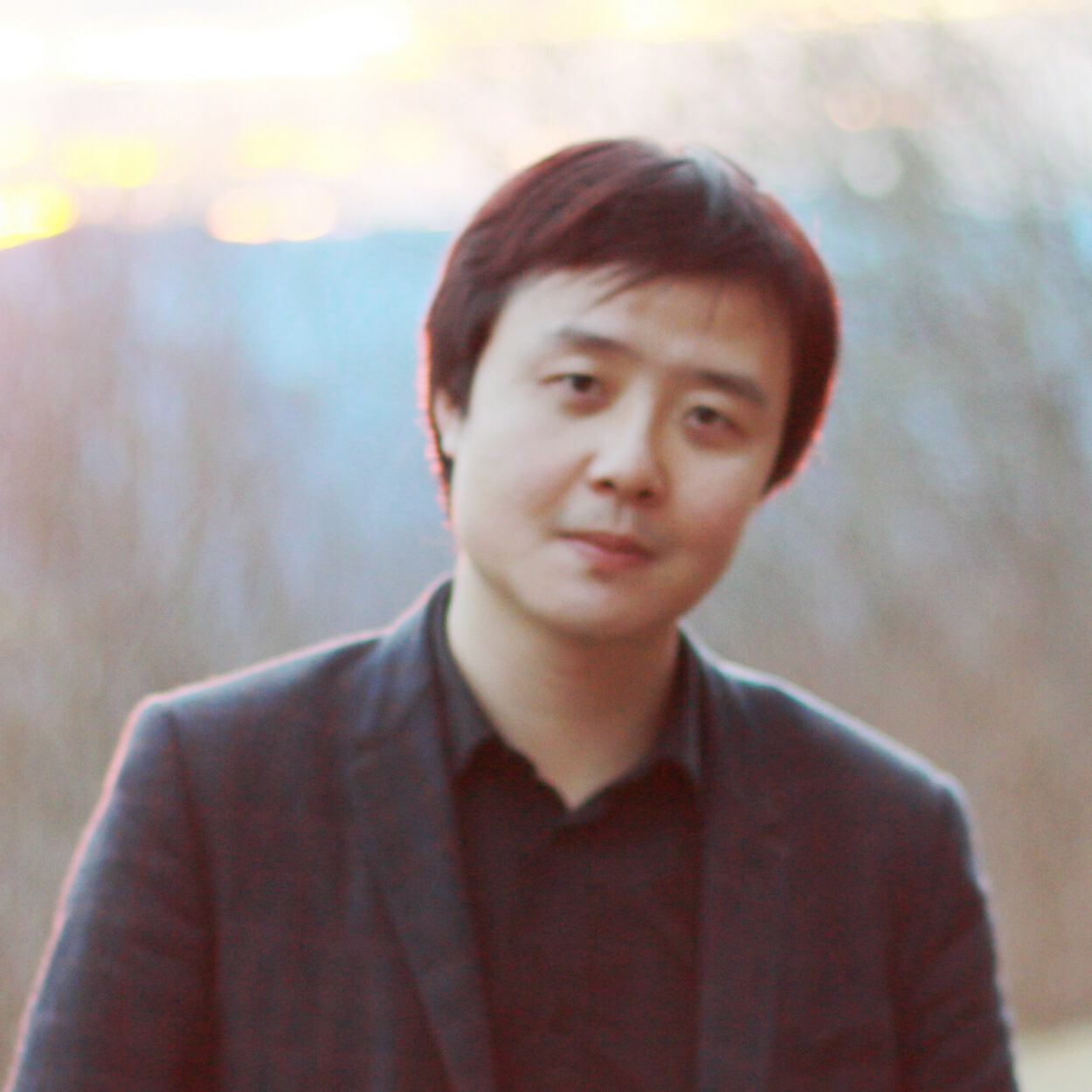}}]
{Mengyu Wang} received his PhD degree from University of Pittsburgh in 2014. He is an Assistant Professor of Ophthalmology at Harvard Medical School and Director of Harvard AI and Robotics Lab. He is an Affiliate Faculty of Kempner Institute for the Study of Natural and Artificial Intelligence at Harvard University, Broad Institute of MIT and Harvard, Harvard Data Science Initiative at Harvard University. His research has been recognized by the NIH K99/R00 Career Development Award, Alcon Young Investigator Grant, Research to Prevent Blindness International Research Collaborators Award, Alliance for Eye and Vision Research Emerging Vision Scientist Program, and Association for Research in Vision and Ophthalmology Advocacy Program.
\end{IEEEbiography}   

	
\end{document}